\documentclass[lettersize,journal]{IEEEtran}
\usepackage{amsmath,amsfonts}
\usepackage{algorithmic}
\usepackage{algorithm}
\usepackage{array}
\usepackage{subcaption}
\usepackage{textcomp}
\usepackage{stfloats}
\usepackage{url}
\usepackage{verbatim}
\usepackage{graphicx}
\usepackage{cite}
\usepackage{mathtools}
\usepackage{amsthm}
\usepackage{accents}
\usepackage{enumitem}
\usepackage{tcolorbox}
\usepackage[algo2e,algoruled,boxed,lined,norelsize]{algorithm2e} 
\usepackage{tablefootnote}
\usepackage{booktabs}
\usepackage{threeparttable}
\graphicspath{{figures/}}

\newtheorem{lemma}{Lemma}

\DeclarePairedDelimiter{\norm}{\lVert}{\rVert}

\newcommand*{\tran}{^{\mkern-1.5mu\mathsf{T}}}

\DeclareMathOperator{\proj}{Proj}

\SetKwInOut{Initial}{Initial}
\SetKwInOut{Parameter}{Parameters}
\SetKw{Continue}{continue}
\allowdisplaybreaks

\begin{document}

\title{Learning from Sparse Demonstrations}

\author{
	Wanxin Jin,
	Todd D. Murphey,
	Dana Kuli\'{c},
	Neta Ezer,
	Shaoshuai Mou
\thanks{
	Wanxin Jin is with the General Robotics, Automation, Sensing and Perception (GRASP) Laboratory, University of Pennsylvania, PA 19104, USA.  Wanxin Jin is the corresponding author. Email: wanxinjin@gmail.com.
	
	Todd D. Murphey is with  the Department of Mechanical Engineering, Northwestern University, Evanston, IL 60208,	USA. This material is partially based upon work supported by the National Science Foundation under award 1837515.  Email: t-murphey@northwestern.edu.
	
	Dana Kuli\'{c} is with the
	Monash University, Clayton, VIC 3800, Australia.	Email: dana.kulic@monash.edu.

	Neta Ezer works for the Northrop Grumman Corporation,  Linthicum Heights, MD 21090, USA. Distribution Statement A: Approved for Public Release; Distribution is Unlimited; \#20-1350; Dated 08/03/20. Email: neta.ezer@ngc.com. 
	
	 Shaoshuai Mou is with the School of Aeronautics and Astronautics, Purdue University, West Lafayette, IN 47906, USA. Dr. Mou’s research is supported in part by grants from the Research in Applications for Learning Machines (REALM) Consortium of Northrop Grumman Corporation, Rolls-Royce Corporation, and the NASA University Leadership Initiative (ULI). Email: mous@purdue.edu.}
}

\markboth{\normalsize
\textit{ T\MakeLowercase{his is a preprint}. T\MakeLowercase{he published version can be accessed at} IEEE T\MakeLowercase{ransactions on} R\MakeLowercase{obotics}.
}
}%
{Shell \MakeLowercase{\textit{et al.}}: A Sample Article Using IEEEtran.cls for IEEE Journals}


\maketitle

\begin{abstract}
This paper develops the method of Continuous Pontryagin Differentiable Programming (Continuous PDP), which enables a robot to learn an objective function from a  few sparsely demonstrated keyframes. The keyframes, labeled with some time stamps, are the desired task-space outputs, which a robot is expected to follow sequentially.  The time stamps of the keyframes can be different from the time of the robot's actual execution.   The method jointly finds an objective function and a time-warping function such that the robot's resulting trajectory sequentially follows the keyframes with minimal discrepancy loss.  The  Continuous PDP minimizes the discrepancy loss using projected gradient descent,  by efficiently solving the gradient of the robot trajectory with respect to the unknown parameters.  The method is first evaluated on a simulated robot arm and then applied to a 6-DoF  quadrotor to learn an objective function for motion planning in unmodeled environments. The results show the efficiency of the method, its ability to handle time misalignment between keyframes and robot execution, and the generalization of objective learning into unseen motion conditions.
\end{abstract}


\section{Introduction}
\IEEEPARstart{T}{he}  appeal of learning from demonstrations (LfD) lies in its capability to facilitate robot programming by simply providing demonstrations. It circumvents the need for expertise of modeling and control design, empowering non-experts to program robots as needed \cite{ravichandar2020recent}. LfD has been successfully applied   manufacturing \cite{denivsa2015learning}, assistive  robots \cite{moro2018learning}, and autonomous vehicles \cite{kuderer2015learning}.

\begin{figure}[t]
	\center	\includegraphics[width=0.95\columnwidth]{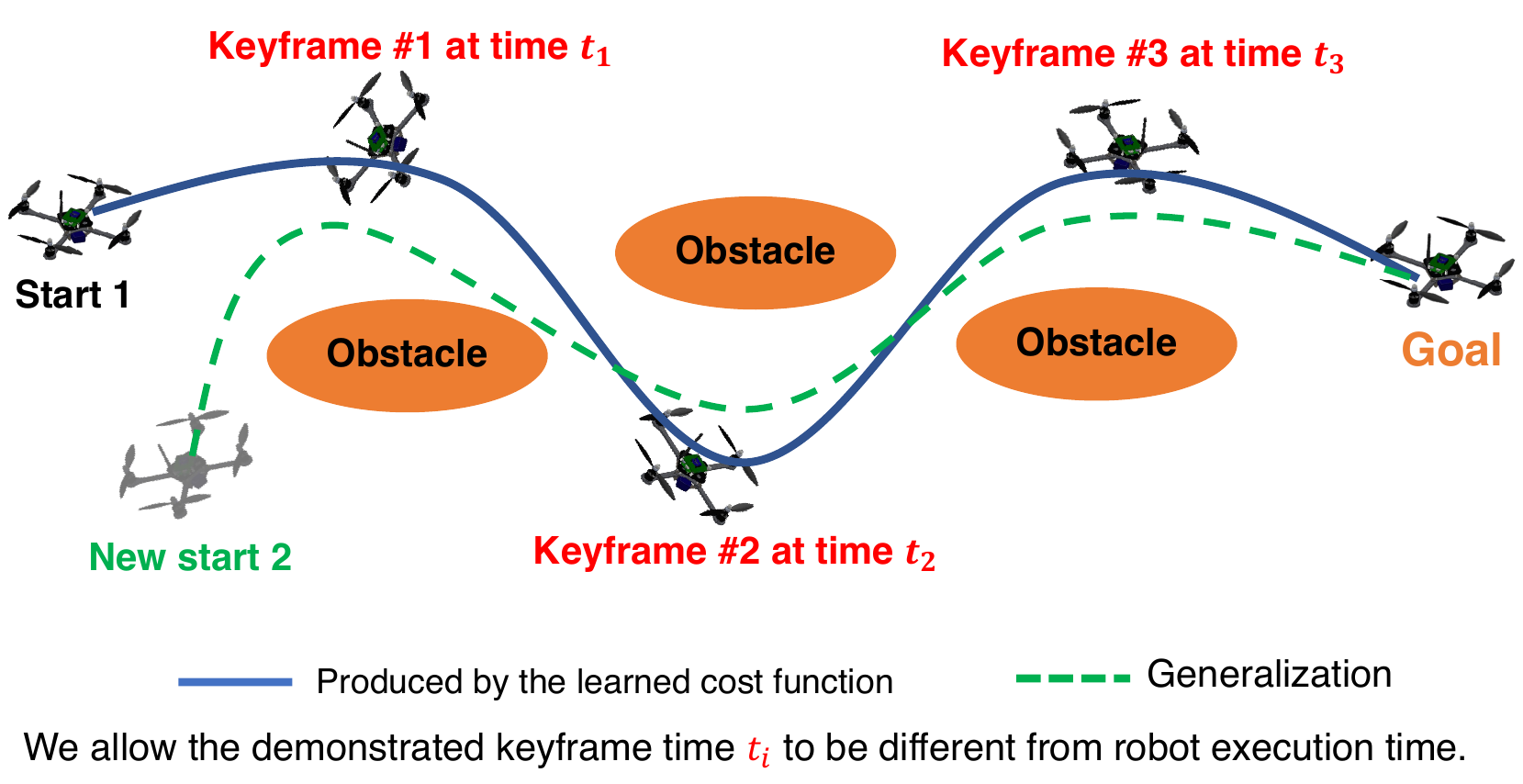}
	\caption{Illustration of learning  from sparsely demonstrated keyframes. Each keyframe is a   desired   output  with a time stamp.  We aim to  learn an  objective function from  keyframes such that the robot motion  (blue line) follows these keyframes. At  first glance, it may seem a problem  of `curve fitting' (i.e., finding a  kinematic path). However, 
		a key difference of our problem  is that learning an objective function enables a robot to generalize new motion in unseen situations, such as  given a  new initial condition (green dashed line). A key feature of the proposed method is that in addition to learning an objective   function, we jointly learn a time-warping  function to account for the  misalignment between  the  keyframe time $t_i$ and  robot actual execution time (due to dynamics constraint).  }
	\label{intro.figure}
\end{figure}

LfD can be broadly categorized into two classes based on what to learn from demonstrations. The first branch of LfD focuses on learning policies \cite{pomerleau1991efficient,englert2013probabilistic,calinon2007learning,rahmatizadeh2018vision, torabi2018behavioral}, which maps directly from robot states, environment, or raw observation data to robot actions. While effective in many situations, policy learning typically requires a considerable amount of demonstration data, and the learned policy may generalize poorly to unseen tasks \cite{ravichandar2020recent}. To alleviate this, the second line of LfD focuses on learning an objective (cost or reward) function from demonstrations \cite{abbeel2004apprenticeship},  from which the  policies or trajectories are derived. These methods assume the optimality of demonstrations and use inverse reinforcement learning (IRL) \cite{ng2000algorithms} or inverse optimal control (IOC) \cite{moylan1973nonlinear} to estimate   objective functions.  Since an objective function is a compact and high-level representation of a task and control principle, learning objective functions has shown an advantage over policy imitation in terms of better generalization  \cite{macglashan2015between} and relatively lower data complexity \cite{abbeel2004apprenticeship}. Despite appealing, objective learning based LfD inherits some limitations from existing IOC/IRL methods \footnote{The literature review here mainly focuses on  model-based IRL methods.} \cite{ratliff2006maximum,ziebart2008maximum,mombaur2010human,keshavarz2011imputing,puydupin2012convex,englert2017inverse}.

First, existing IOC/IRL methods cannot handle the time misalignment between demonstrations and actual execution of a robot \cite{kingston2011time}. For instance, the speed of demonstrations maybe not be achievable by a robot, as the robot is actuated by weak motors and cannot move as fast as the demonstrations. Second, existing methods   usually require the demonstrations of complete motion trajectories or at least a continuous segment of states-inputs,  making it challenging in data collection for high-dimensional and long-horizon tasks.  Third,  
existing IOC/IRL  may not be efficient when handling high-dimensional continuous systems/tasks or learning complex objective functions, such as deep neural network objective functions.

This paper develops the \emph{Continuous Pontryagin Differentiable Programming} method, abbreviated as the \emph{Continuous PDP}, to address the existing challenges. The method requires only a few keyframes demonstrated at sparse time instances,  and it learns both an objective function and a time-warping function, which accounts for the time misalignment between   demonstration and robot actual execution. The Continuous PDP  minimizes a  discrepancy loss between the robot reproduced motion and the given keyframes via the projected gradient descent. This is done by efficiently computing the \emph{analytical gradient} of the robot trajectory with respect to  tunable parameters in the objective and time-warping functions. The highlights of the Continuous PDP are listed as follows.

\begin{itemize}
	
	\item[(i)] It requires as input the keyframe demonstrations, defined as a small number of \emph{sparse} desired task-space outputs, which the robot is expected to follow sequentially, as in Fig.~\ref{intro.figure}.

\item[(ii)] As the time stamp of each keyframe may not correctly reflect the time of robot execution, in addition to learning an objective function, the method jointly searches for a  \emph{time-warping function}, which accounts for the time misalignment between keyframes and robot execution.

\item[(iii)] The method can efficiently handle continuous-time high-dimensional systems and accepts {any differentiable parameterization of objective functions}.  
\end{itemize}

\subsection{Related Work}\label{sectionrelatedwork}
Since the theme of this paper belongs to the category of objective learning,  in the following we mainly review IOC/IRL methods. For other types of LfD, e.g., learning policies,  please refer to the recent surveys   \cite{ravichandar2020recent,osa2018algorithmic}.

\subsubsection{Classic Strategies in IOC/IRL}

Existing IOC/IRL methods can be categorized into two classes. The first class adopts a bi-level framework, where an objective function is updated on an outer level while the corresponding reinforcement learning (or optimal control) problem is solved on an inner level. Different methods in this class use different strategies to update an objective function. Representative work includes feature-matching IRL \cite{abbeel2004apprenticeship}, where an objective function is updated to match the feature values of the reproduced trajectory with the ones of the demonstrations, max-margin IRL \cite{ratliff2006maximum,doerr2015direct}, where an objective function is updated by maximizing the margin between the objective value of the reproduced trajectory and that of demonstrations, and max-entropy IRL \cite{ziebart2008maximum}, which finds an objective function such that the trajectory distribution has maximum entropy while subject to the empirical feature values. The second class of IOC/IRL  \cite{keshavarz2011imputing,puydupin2012convex,englert2017inverse,jin2018inverse,jin2019inverse}  directly solves for objective function parameters by establishing the optimality conditions, such as KKT conditions \cite{kuhn2014nonlinear} or Pontryagin's Maximum Principle  \cite{pontryagin1962mathematical,jin2021distributed}. The key idea is that a demonstration is assumed to be optimal and thus must satisfy the optimality condition. By directly minimizing the violation of the optimality conditions by demonstration data, one can compute the objective function parameters.

\subsubsection{IOC with Trajectory Loss}
One type of bi-level IOC/IRL formulation also uses a trajectory loss as its learning criterion. A trajectory loss is to evaluate the discrepancy between the demonstrations and the robot motion reproduced by the objective function estimate. For example, \cite{mombaur2010human} and  \cite{mombaur2013forward} develop a bi-level IOC approach which learns an objective function from  human locomotion data. In their work, the trajectory loss is minimized via a derivative-free technique \cite{powell2009bobyqa}, where the key is to approximate the loss using a quadratic function. The approach requires solving optimal control problems multiple times at each update, thus is computationally expensive. Further, the derivative-free methods are known to be challenging for the problem of large size \cite{rios2013derivative}. In \cite{hatz2012estimating}, the authors convert a bi-level IOC to a \emph{plain} optimization by replacing the lower-level optimal control problem with its optimality conditions (the Pontryagin's Maximum Principle). Although the converted plain optimization can be solved by an off-the-shelf nonlinear optimization solver, the decision variables of the plain optimization include both objective parameters and system trajectory (and dual variables); thus dramatically increasing the size of the optimization. 
Besides, both lines of methods have not considered the time misalignment between demonstrations and robot execution.

Compared to the derivative-free methods in \cite{mombaur2010human, mombaur2013forward},
the proposed Continuous PDP  solves IOC/IRL by directly computing the \emph{analytical} gradient of a trajectory loss with respect to tunable parameters in an objective function and a time-warping function,  thus is capable of solving high-dimensional continuous tasks. Compared to  \cite{hatz2012estimating},   the Continuous PDP \emph{maintains} the bi-level hierarchy of the problem and solves  IOC  by \emph{differentiating through the inner-level optimal control system}.  Maintaining a bi-level structure enables us to treat the outer and inner level subproblems separately, avoiding the mixed treatment that can lead to a dramatic increase in the size of optimization. In Section \ref{comparetoplain}, we  provide the comparison  between the Continuous PDP and \cite{hatz2012estimating}.

\subsubsection{IOC/IRL via differentiable through inner-level optimization} The recent work focuses on solving bi-level IOC/IRL by differentiating through inner-level optimization. E.g.,  \cite{das2020model}  learns a cost function from visual demonstrations by differentiating through the inner-level MPC. Specifically, those methods treat the inner-level optimization as an \emph{unrolling computational graph} of repetitively applying gradient descent, such that the automatic differentiation
\cite{abadi2016tensorflow} can be applied. However, as shown in   \cite{jin2020pontryagin} and \cite{amos2018differentiable},  auto-differentiating an  ‘unrolling’ graph has the following drawback: (i) it
needs to store all intermediate results along with the  graph, thus is memory-expensive;
and (ii)  the accuracy of the  unrolling differentiation  depends on the length of the ‘unrolled’ graph, thus facing a trade-off
between complexity and accuracy. 
In contrast, the Continuous PDP  computes the gradient directly on the optimal trajectory produced from the inner level, without memorizing how this inner-level solution is obtained. Thus, there are no above challenges for the proposed method.

\subsubsection{Time-warping}
Using time-warping functions to model the time misalignment between two temporal sequences has been extensively studied in signal processing \cite{sakoe1978dynamic} and pattern recognition \cite{chang2019d3tw}. In \cite{vakanski2012trajectory,vukovic2015trajectory},  time-warping is used in LfD for learning and producing robot trajectories. In \cite{kingston2011time},  a time-warping function between robot and demonstrator is learned for optimal tracking. 	
All the above methods focus on learning policy or trajectory models instead of objective functions. For time-misalignment in IOC/IRL, a main technical challenge is how to incorporate the search of a time-warping function into the objective learning process. 
The Continuous PDP  addresses this challenge by finding an objective function and a time-warping function simultaneously using gradient descent.

\subsubsection{Incomplete Trajectory or Sparse Waypoints}
Some methods focus on learning from incomplete trajectories. In  \cite{jin2018inverse,jin2020inverse}, the authors develop a method to solve IOC with trajectory segments. It requires the length of a  segment to satisfy a recovery condition and cannot directly learn from sparse points.   \cite{akgun2012keyframe,akgun2012trajectories} consider learning from a set of sparse waypoints,   but they learn a kinematic model instead of an objective function.  Compared to those methods, the proposed method learns an objective function and a time-warping function from a  small set of time-stamped  \emph{sparse keyframes}, i.e., a few desired task-space outputs. In Section \ref{exp.compare.keyframes}, we will provide a comparison  with \cite{akgun2012keyframe}.

\subsubsection{Sensitivity Analysis and Continuous PDP}

The idea of the  Continuous PDP  is similar to the well-known sensitivity analysis \cite{fiacco1976sensitivity,levysensitivity} in nonlinear optimization, where the KKT conditions are differentiated to obtain the gradient of a  solution with respect to the objective function parameters. In sensitivity analysis, it requires to compute the inverse of the Hessian matrix in order to apply the well-known implicit function theorem \cite{krantz2012implicit}.  If trying to apply the sensitivity analysis to a  \emph{continuous-time} optimal control problem in our formulation, we may face the following challenge. 
Since the optimality condition of a continuous-time optimal control problem is Pontryagin's Maximum Principle \cite{pontryagin1962mathematical}, which is a set of \emph{ODE equations}. To apply the sensitivity analysis, one would need to first discretize the continuous-time system, and this will lead to a Hessian matrix of the size at least $\frac{T}{\Delta t}\times \frac{T}{\Delta t}$ ($T$ is the time horizon, and $\Delta t$ is the discretization interval); this will cause huge computation cost when taking its inverse (the complexity is at least  $\mathcal{O}((\frac{T}{\Delta t})^2)$). The reason why we do not formulate the problem in discrete-time in the first place is that otherwise, learning a \emph{discrete time-warping} function will lead the problem to a mixed-integer optimization, which becomes more challenging to attack.

Compared to sensitivity analysis,  the Continuous  PDP has the following new technical aspects. First, it directly differentiates  the  ODE  equations in Pontryagin's  Maximum Principle \cite{pontryagin1962mathematical}, producing  \emph{Differential Pontryagin's Maximum Principle}; and second, importantly, it develops  \emph{Riccati-type equations} to solve the Differential Pontryagin's Maximum Principle to obtain the	
trajectory gradient (Lemma \ref{theorem1}). The complexity of this process is only $\mathcal{O}(T)$.
The  Continuous PDP  is an extension of our previous work Pontryagin Differentiable Programming (PDP) \cite{jin2020pontryagin,jin2021safe} into the continuous-time systems. For a more detailed comparison between PDP and the sensitivity analysis, we refer the reader to \cite{jin2020pontryagin,jin2021safe}.

The following paper  is organized as follows: Section II sets up the problem. Section III reformulates the problem using  time-warping techniques. Section IV proposes the Continuous  PDP method. Experiments are given in Sections V and  VI. Section VII presents discussion, and  Section VIII draws conclusions.

\section{Problem Formulation}
Consider a robot with the following continuous dynamics:
\begin{equation}\label{dyn}
\boldsymbol{\dot{x}}(t)=\boldsymbol{f}(\boldsymbol{x}(t),\boldsymbol{u}(t)) \quad \text{with} \quad \boldsymbol{x}(0),
\end{equation}
where $\boldsymbol{x}(t)\in\mathbb{R}^n$ is the robot state; $\boldsymbol{u}(t)\in\mathbb{R}^m$ is the control input;  vector function $\boldsymbol{f}:\mathbb{R}^n\times\mathbb{R}^m \mapsto \mathbb{R}^n$ is assumed to be twice-differentiable,  and $t\in[0,\infty)$ is time. Suppose  the robot motion over a time horizon $t_f>0$ is controlled by minimizing the following parameterized  cost function:
\begin{equation}\label{costfun}
J(\boldsymbol{p})=\int_{0}^{t_f}c(\boldsymbol{x}(t),\boldsymbol{u}(t),{\boldsymbol{p}})dt+h(\boldsymbol{x}(t_f),{\boldsymbol{p}}),
\end{equation}
where  $c(\boldsymbol{x},\boldsymbol{u},{\boldsymbol{p}})$ and $h{(\boldsymbol{x},\boldsymbol{p}})$ are the running  and final  costs, respectively, both of which are  assumed twice-differentiable; and $\boldsymbol{p}\in\mathbb{R}^{r}$ is a tunable parameter vector. For a fixed choice of $\boldsymbol{p}$, the robot  produces a trajectory of states and inputs
\begin{equation}\label{traj}
\boldsymbol{\xi}_{\boldsymbol{p}}=\{\boldsymbol{\xi}_{\boldsymbol{p}}(t)\,| \,0\leq t\leq t_f\}\,\, \text{with}\,\, \boldsymbol{\xi}_{\boldsymbol{p}}(t)=\{\boldsymbol{x}_{\boldsymbol{p}}(t),\boldsymbol{u}_{\boldsymbol{p}}(t)\}.
\end{equation} 
which minimizes  (\ref{costfun}) subject to (\ref{dyn}). The subscript in $\boldsymbol{\xi}_{\boldsymbol{p}}$ means that the  trajectory  implicitly depends on $\boldsymbol{p}$.

The goal of learning from demonstrations is to estimate the cost function parameter  $\boldsymbol{p}$ from the given demonstrations by a user (usually a human). 
Suppose that a user provides demonstrations in a task space (e.g., Cartesian space or  vision measurement), which is a known differentiable mapping of the robot state-input pair:
\begin{equation}\label{interface}
\boldsymbol{y}=\boldsymbol{g}(\boldsymbol{x},\boldsymbol{u}),
\end{equation}
where  $\boldsymbol{g}:\mathbb{R}^n\times\mathbb{R}^m\rightarrow\mathbb{R}^o$ defines a mapping  from the robot state-input to a task output   $\boldsymbol{y}\in \mathbb{R}^{o}$. The user's  demonstrations include  (i) an \emph{expected} time horizon $T$, and (ii) a number of $N$  keyframes, each of which is a desired   output   labeled with an \emph{expected} time stamp $\tau_i$, denoted as
\begin{equation}\label{corrections}
\mathcal{D}=\{\boldsymbol{y}^*(\tau_i)\,|\,
\tau_i\in[0,T],\, i=1,2, \cdots, N
\}.
\end{equation}
Here, $\boldsymbol{y}^*(\tau_i)$ is the $i$th   keyframe demonstrated by the user, and ${\tau_i}$ is the expected time stamp at which the user wants the robot to reach $\boldsymbol{y}^*(\tau_i)$. The keyframe time $\{\tau_1, \tau_2, ..., \tau_N\}$ can be sparsely located within range $[0, T]$. As the user can freely choose   $N$  and  $\tau_i$ relative to the expected horizon $T$, we call  $\mathcal{D}$ as \emph{keyframes}. As shown later in experiments, $N$  can be small.

Note that both the expected  horizon $T$ and the expected time stamps $\tau_i$ are in the  time axis of the user's demonstration. This demonstration time axis may not be identical to the  actual time axis of robot execution; in other words,    $T$ and $\tau_i$ may not be achievable by the robot. For example, when the robot is actuated by a weak servo motor, its motion inherently cannot meet $\tau_i$. To accommodate the time misalignment between  the robot execution and keyframes, we introduce a   \emph{time warping function}:
\begin{equation}\label{warping_fun}
t=w(\tau),
\end{equation}
which   maps from  keyframe time  $\tau$ to  robot time  $t$. We make the following reasonable assumption:  $w$ is strictly increasing in the range  $[0,T]$, continuously differentiable, and $w(0)=0$.

Given the keyframes $\mathcal{D}$, \textbf{the problem of interest} is  to find   cost function parameter $\boldsymbol{p}$ and a time-warping function $w(\cdot)$ such that the  \emph{task discrepancy loss} is minimized:
\begin{equation}\label{loss}
\begin{aligned}
\min\limits_{\boldsymbol{p},w}\sum\nolimits_{i=1}^{N}l\Big(\boldsymbol{y}^*(\tau_i),
\boldsymbol{g}\big(\boldsymbol{\xi}_{\boldsymbol{p}}(w(\tau_{i}))   \big)    \Big),
\end{aligned}
\end{equation}
where ${l}(\boldsymbol{a},\boldsymbol{b})$ is  a given differentiable scalar function defined in the task space which quantifies a  distance metric between vectors $\boldsymbol{a}$ and $\boldsymbol{b}$, e.g., $l(\boldsymbol{a},\boldsymbol{b})=\norm{\boldsymbol{a}-\boldsymbol{b}}^2$. 
Minimizing (\ref{loss}) means that we want  the robot to find  cost function parameter $\boldsymbol{p}$ and a time-warping function $w(\cdot)$, such that its reproduced trajectory gets as close to the given keyframes  as possible.

\section{Problem Reformulation based on Time-Warping Techniques}\label{section_problem_formulation}
In this section, we re-formulate the problem of interest using  the time-warping techniques.

\subsection{Parametric Time Warping Function}
To facilitate learning of an unknown time-warping function, we first parameterize the time-warping function. Recall  that a differentiable time-warping function ${w}(\tau)$ satisfies  $w(0)=0$ and is strictly increasing in the range $[0, T]$, i.e.,  
\begin{equation}\label{derivativetime-warping}
v(\tau)=\frac{d w(\tau)}{d \tau}>0 
\end{equation}
for all $\tau\in[0, T]$. We   use a polynomial time-warping function: 
\begin{equation}\label{para_warp}
t=w_{\boldsymbol{\beta}}(\tau)=\sum\nolimits_{i=1}^{s}\beta_i\tau^i,
\end{equation}
where $\boldsymbol{\beta}=[\beta_1,\beta_2,\cdots,\beta_s]\tran\in\mathbb{R}^s$ is the coefficient vector. Since $w_{\boldsymbol{\beta}}(0)=0$, there is no constant (zero-order) term in (\ref{para_warp}) (i.e., $\beta_0=0$). Due to the requirement of $ {d w_{\boldsymbol{\beta}}}/{d\tau}=v_{\boldsymbol{\beta}}(\tau)>0$ for all $\tau \in [0, T]$, one can  obtain a feasible set, denoted as $\Omega_{\boldsymbol{\beta}}$, such that $ \frac{d w_{\boldsymbol{\beta}}(\tau)}{d \tau}>0$ for all $\tau\in[0,T]$ if $\boldsymbol{\beta}\in\Omega_{\boldsymbol{\beta}}$. The choice of  polynomial degree $s$ will decide the representation power of (\ref{derivativetime-warping}): larger $s$ means that  $w_{\boldsymbol{\beta}}(\tau)$ can represent more complex time warping curves. Note that although  we   use  a polynomial  time-warping function, the   method in this paper  allows for  more general parameterization of a time-warping function, as long as it is differentiable. This paper uses  polynomial time-warping functions due to the simplicity for implementation.

\subsection{Equivalent Formulation by Time Warping}
Substituting the parametric time-warping function ${w}_{\boldsymbol{\beta}}$ in (\ref{para_warp}) into both the robot dynamics (\ref{dyn}) and   cost function  (\ref{costfun}), we obtain the following time-warped dynamics
\begin{equation}\label{dyn_wrap}
\frac{d \boldsymbol{x}}{d{\tau}}=\frac{d {w}_{\boldsymbol{\beta}}}{d\tau} \boldsymbol{f}\Big( \boldsymbol{x}(w_{\boldsymbol{\beta}}(\tau)),\boldsymbol{u}(w_{\boldsymbol{\beta}}(\tau))\Big) \quad \text{with} \,\, \boldsymbol{x}(0),
\end{equation}
and the  time-warped  cost function
\begin{equation}\label{costfun_wrap}
\begin{aligned}
J(\boldsymbol{p},\boldsymbol{\beta})=&\int_{0}^{T}\frac{d w_{\boldsymbol{\beta}}}{d \tau}c_{\boldsymbol{p}}(\boldsymbol{x}(w_{\boldsymbol{\beta}}(\tau)),\boldsymbol{u}(w_{\boldsymbol{\beta}}(\tau)))d\tau\\
&+h_{\boldsymbol{p}}(\boldsymbol{x}(w_{\boldsymbol{\beta}}(T))).
\end{aligned}
\end{equation}
Here, the left side of (\ref{dyn_wrap}) is due to  chain rule: $\frac{d \boldsymbol{x}}{d\tau}=\dot{\boldsymbol{x}}\frac{d t}{d \tau}$, and the   time horizon satisfies $t_f=w_{\boldsymbol{\beta}}(T)$ (note that $T$ is specified by the demonstrator). 
For notation simplicity, we write $\frac{d {w}_{\boldsymbol{\beta}}}{d\tau}=v_{\boldsymbol{\beta}}(\tau)$, $\boldsymbol{x}(w(\tau))=\boldsymbol{x}(\tau)$, $\boldsymbol{u}(w(\tau))=\boldsymbol{u}(\tau)$, and $\frac{d \boldsymbol{x}}{d{\tau}}=\dot{\boldsymbol{x}}(\tau)$. Then, the above time-warped dynamics  (\ref{dyn_wrap}) and time-warped  cost function  (\ref{costfun_wrap})  are rewritten as:
\begin{subequations}\label{ocsys}
	\begin{equation}\label{dyn_wrap2}
	\dot{\boldsymbol{x}}(\tau)=v_{\boldsymbol{\beta}}(\tau)\boldsymbol{f}(\boldsymbol{x}(\tau),\boldsymbol{u}(\tau)) \quad \text{with} \,\, \boldsymbol{x}(0)
	\end{equation}
	and
	\begin{equation}\label{costfun_wrap2}
	J(\boldsymbol{p},\boldsymbol{\beta}){=}\int_{0}^{T}v_{\boldsymbol{\beta}}(\tau)c(\boldsymbol{x}(\tau),\boldsymbol{u}(\tau), {\boldsymbol{p}})d\tau+h(\boldsymbol{x}(T),{\boldsymbol{p}}),
	\end{equation}
\end{subequations}
respectively. We pack the tunable cost  parameter  $\boldsymbol{p}$ and  time-warping  parameter  $\boldsymbol{\beta}$   together as
\begin{equation}\label{parameter}
\boldsymbol{{\theta}}=[\boldsymbol{p}\tran,\boldsymbol{\beta}\tran]\tran\in\mathbb{R}^{r+s}.
\end{equation}
For a fixed $\boldsymbol{{\theta}}$, the  optimal trajectory  from solving the above time-warped optimal control system (\ref{ocsys}) is rewritten as
\begin{equation}\label{traj_warped}
\boldsymbol{\xi}_{\boldsymbol{\theta}}=\{\boldsymbol{\xi}_{\boldsymbol{\theta}}(\tau)\,\,| \,\,0\leq \tau\leq T\}, \,\,
\end{equation}
with $\boldsymbol{\xi}_{\boldsymbol{\theta}}(\tau)=\{\boldsymbol{x}_{\boldsymbol{\theta}}(\tau),\boldsymbol{u}_{\boldsymbol{\theta}}(\tau)\}$.  The discrepancy loss  (\ref{loss})  can now be defined as
\begin{equation}\label{loss_warped}
L (\boldsymbol{\xi}_{\boldsymbol{\theta}},\mathcal{D})=\sum_{i=1}^{N}l\Big(\boldsymbol{y}^*(\tau_i),\,\,
\boldsymbol{g}\big(\boldsymbol{\xi}_{\boldsymbol{\theta}}(\tau_i)\big)\Big).
\end{equation}
Minimizing   (\ref{loss_warped}) over   $\boldsymbol{\theta}$  is a process of simultaneously searching for  a  cost function $J(\boldsymbol{p})$ and   time-warping function $w_{\boldsymbol{\beta}}(\tau)$. 
In sum, the problem of interest is now reformulated as the following optimization 
\begin{equation}\label{problem}
\begin{aligned}
&\min_{\boldsymbol{\theta}\in\boldsymbol{\Theta}}\,\,\,  L (\boldsymbol{\xi}_{\boldsymbol{\theta}},\mathcal{D})\\
&\quad\text{s.t.}\quad \boldsymbol{\xi}_{\boldsymbol{\theta}}  \text{ is from the optimal control system (\ref{ocsys})}.
\end{aligned}
\end{equation}
Here $\boldsymbol{\Theta}$ defines a feasible set of  $\boldsymbol{\theta}$,   $\boldsymbol{\Theta}=\mathbb{R}^{r}\times\Omega_{\boldsymbol{\beta}}$.   (\ref{problem}) is a bi-level optimization, where the upper level is to minimize a discrepancy loss between the keyframes $\mathcal{D}$ and the reproduced time-warped trajectory $\boldsymbol{\xi}_{\boldsymbol{\theta}}$, and  the inner level  is to generate such $\boldsymbol{\xi}_{\boldsymbol{\theta}}$ by  solving the optimal control problem  (\ref{ocsys}). In the next section, we will develop the  \emph{Continuous Pontryagin Differentiable Programming}  to efficiently solve   (\ref{problem}).

\section{Continuous Pontryagin Differentiable Programming}

\subsection{Algorithm Overview}
\begin{figure}[h] 
	\centering	\includegraphics[width=1\columnwidth]{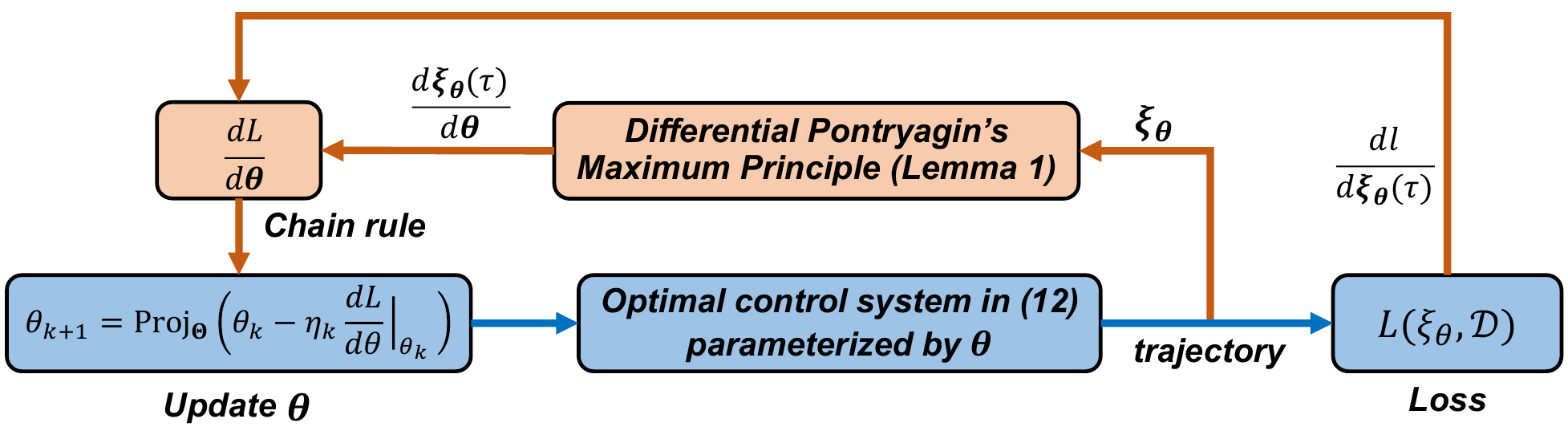}
	\caption{Computational diagram of the Continuous Pontryagin Differentiable Programming.}
	\label{pdp.diagm}
\end{figure}

To solve the optimization (\ref{problem}), we start with an arbitrary initial guess $\boldsymbol{\theta}_0\in\boldsymbol{\Theta}$, and apply the gradient descent
\begin{equation}\label{update}
\boldsymbol{\theta}_{k+1}=\proj_{\boldsymbol{\Theta}}\left(\boldsymbol{\theta}_k-\eta_k\frac{d L}{d \boldsymbol{\theta}}\Bigr|_{\boldsymbol{\theta}_k}\right),
\end{equation}
where $k$ is the iteration  index; $\eta_k$ is the  step size (or learning rate); $\proj_{\boldsymbol{\Theta}}$ is a projection operator to enforce the feasibility of $\boldsymbol{\theta}_k$ in $\boldsymbol{\Theta}$, e.g.,  $\proj_{\boldsymbol{\Theta}}(\boldsymbol{\theta})=\arg\min_{\boldsymbol{z}\in\boldsymbol{\Theta}}\norm{\boldsymbol{\theta}-\boldsymbol{z}}$; and $\frac{d L}{d \boldsymbol{\theta}}\bigr|_{\boldsymbol{\theta}_k}$ denotes the gradient of the  loss  (\ref{loss_warped}) directly with respect to $\boldsymbol{\theta}$ evaluated at $\boldsymbol{\theta}_k$. Applying the chain rule, we have
\begin{equation}\label{chainrule}
\frac{d L}{d \boldsymbol{\theta}}\Bigr|_{\boldsymbol{\theta}_k}=\sum_{i=1}^{N}\frac{\partial l}{\partial \boldsymbol{\xi}_{\boldsymbol{\theta}}(\tau_i)}\Bigr|_{\boldsymbol{\xi}_{\boldsymbol{\theta}_k}(\tau_i)}\frac{{\partial \boldsymbol{\xi}_{\boldsymbol{\theta}}{(\tau_i)}}}{\partial \boldsymbol{\theta}}\Big\rvert_{{\boldsymbol{\theta}}_k},
\end{equation}
where  $\frac{\partial l}{\partial \boldsymbol{\xi}_{\boldsymbol{\theta}}{(\tau_i)}}\big\rvert_{\boldsymbol{\xi}_{\boldsymbol{\theta}_k}(\tau_i)}$ is the gradient of the single keyframe  loss  $l=\Big(\boldsymbol{y}^*(\tau_i),\,\,
\boldsymbol{g}\big(\boldsymbol{\xi}_{\boldsymbol{\theta}}(\tau_i)\big)\Big)$ in (\ref{loss_warped})  with respect to  the time-$\tau_{i}$ trajectory point $\boldsymbol{\xi}_{\boldsymbol{\theta}}{(\tau_{i})}$, evaluated at value $\boldsymbol{\xi}_{\boldsymbol{\theta}_k}{(\tau_{i})}$, and $\frac{{\partial \boldsymbol{\xi}_{\boldsymbol{\theta}}{(\tau_i)}}}{\partial \boldsymbol{\theta}}\big\rvert_{{\boldsymbol{\theta}}_k}$ is the gradient of the time-$\tau_{i}$ trajectory point  $\boldsymbol{\xi}_{\boldsymbol{\theta}}{(\tau_i)}$, with respect to  $\boldsymbol{\theta}$, evaluated at value $\boldsymbol{\theta}_k$.  From  (\ref{update}) and (\ref{chainrule}), we can draw the computational  diagram in Fig. \ref{pdp.diagm}. Fig. \ref{pdp.diagm} shows that at each iteration $k$,  the update of   $\boldsymbol{\theta}_k$ includes the following three steps:

\begin{itemize}[leftmargin=35pt,font=\itshape]
	\setlength\itemsep{0.6em}
	\item[\textbf{Step 1}:] \emph{	Obtain the optimal  trajectory $\boldsymbol{{\xi}}_{\boldsymbol{\theta}_{k}}$ by solving the optimal control (trajectory optimization) problem  (\ref{ocsys})  with current  $\boldsymbol{\theta}_k$;}
	
	\item[\textbf{Step 2}:] \emph{Compute the gradient    $\frac{\partial l}{\partial \boldsymbol{\xi}_{\boldsymbol{\theta}}{(\tau_i)}}\big\rvert_{\boldsymbol{\xi}_{\boldsymbol{\theta}_k}(\tau_i)}$;}
	
	\item[\textbf{Step 3}:] \emph{Compute the gradient  $\frac{{\partial \boldsymbol{\xi}_{\boldsymbol{\theta}}{(\tau_i)}}}{\partial \boldsymbol{\theta}}\big\rvert_{{\boldsymbol{\theta}}_k}$; }
	
	\item[\textbf{Step 4}:] \emph{Apply  chain rule (\ref{chainrule}) to compute  $\frac{d L}{d \boldsymbol{\theta}}\big\rvert_{\boldsymbol{\theta}_k}$, and update  $\boldsymbol{\theta}_k$ using (\ref{update}) to $\boldsymbol{\theta}_{k+1}$.}
\end{itemize}

The interpretation of the above procedure is straightforward. At each update $k$, the first step is to use the current parameter $\boldsymbol{\theta}_k$ to compute the current optimal trajectory $\boldsymbol{\xi}_{\boldsymbol{\theta}_{k}}$ by solving the optimal control problem    (\ref{ocsys}). In Step 2 and Step 3, the gradient of  the loss with respect to the trajectory point,  $\frac{\partial l}{\partial \boldsymbol{\xi}_{\boldsymbol{\theta}}{(\tau_i)}}\big\rvert_{\boldsymbol{\xi}_{\boldsymbol{\theta}_k}(\tau_i)}$, and the gradient of the trajectory point with respect to  parameters, $\frac{{\partial \boldsymbol{\xi}_{\boldsymbol{\theta}}{(\tau_i)}}}{\partial \boldsymbol{\theta}}\big\rvert_{{\boldsymbol{\theta}}_k}$, are computed, respectively.  In Step 4,  the total gradient of the loss with respect to  the parameter, $\frac{d L}{d \boldsymbol{\theta}}\big\rvert_{\boldsymbol{\theta}_k}$, is assembled via  chain rule  (\ref{chainrule}), and then used to update   $ {\boldsymbol{\theta}_{k}}$ by the projected gradient descent  (\ref{update}).

In Step 1,  the  optimal trajectory $\boldsymbol{\xi}_{\boldsymbol{\theta}_k}$  can be solved  by  available optimal control (trajectory optimization) solvers such as iLQR \cite{li2004iterative}, DDP \cite{jacobson1970differential}, Casadi \cite{Andersson2019}, GPOPS \cite{patterson2014gpops}, etc. In Step 2, the gradient  $\frac{\partial L}{\partial \boldsymbol{\xi}_{\boldsymbol{\theta}}{(\tau_i)}}$  can be readily computed by directly differentiating the given  loss   (\ref{loss_warped}).  The main challenge, however, lies in Step 3, i.e.,  computing  $\frac{{\partial \boldsymbol{\xi}_{\boldsymbol{\theta}}}}{\partial \boldsymbol{\theta}}\big\rvert_{{\boldsymbol{\theta}}_k}$, the gradient of the  optimal trajectory $\boldsymbol{\xi}_{\boldsymbol{\theta}}$ with respect to  the parameter  $\boldsymbol{\theta}$ of  the optimal control system (\ref{ocsys}).  In what follows, we will efficiently solve it  by proposing the technique  of Differential  Pontryagin's Maximum Principle.  For notation simplicity,  we suppress the iteration index $k$ below.

\subsection{Differential Pontryagin's Maximum Principle}
\label{section.dpmp}

In this section, we focus on efficiently solving the analytical gradient of a trajectory of a continuous-time optimal control system with respect to the system parameter. We assume that the resulting optimal trajectory  $\boldsymbol{\xi}_{\boldsymbol{\theta}}$ in  (\ref{traj_warped})  is differentiable with respect to the system parameter $\boldsymbol{\theta}$. This assumption is satisfied if  $\boldsymbol{\xi}_{\boldsymbol{\theta}}$  satisfies the second-order sufficient condition, that is, $\boldsymbol{\theta}$ is a \emph{locally unique optimal} trajectory  (see Lemma 1 in \cite{jin2021safe}). Both our later experiments and previous empirical results \cite{jin2020pontryagin,kolstad1990derivative} show that the differentiability condition is very mild. For more  detailed results about the differentiability for a general optimal control system with respect to system parameters, we refer the reader to   \cite{jin2021safe}.

Consider an optimal trajectory  $\boldsymbol{\xi}_{\boldsymbol{\theta}}$  in (\ref{traj_warped}) produced by an optimal control system  (\ref{ocsys}) with a fixed $\boldsymbol{\theta}$. The Pontryagin's Maximum Principle \cite{pontryagin1962mathematical} states  a set of ODE conditions  that  $\boldsymbol{\xi}_{\boldsymbol{\theta}}$ must  satisfy. To present the  Pontryagin's Maximum Principle,   define the   Hamiltonian \cite{lewis2012optimal}:
\begin{multline}\label{Hamil}
H(\tau)=v_{\boldsymbol{\beta}}(\tau)c_{\boldsymbol{p}}(\boldsymbol{x}(\tau),\boldsymbol{u}(\tau))\\+\boldsymbol{\lambda}(\tau)\tran v_{\boldsymbol{\beta}}(\tau)\boldsymbol f(\boldsymbol{x}(\tau),\boldsymbol{u}(\tau)),
\end{multline} 
where $\boldsymbol{\lambda}(\tau)\in\mathbb{R}^{n}$ is called the costate, $0\leq \tau\leq  T$. According to the Pontryagin's Maximum Principle \cite{pontryagin1962mathematical}, there exists
\begin{equation}\label{costate}
\{\boldsymbol{\lambda}_{\boldsymbol{\theta}}(\tau)\,|\,0\leq \tau\leq T\},
\end{equation}
associated with the optimal trajectory $\boldsymbol{\xi}_{\boldsymbol{\theta}}$ in (\ref{traj_warped}), such that the following ODE equations  hold \cite{pontryagin1962mathematical}:
\begin{subequations}\label{pmp}
	\begin{align}
	\dot{\boldsymbol{x}}_{\boldsymbol{\theta}}(\tau)&=\frac{\partial H}{\partial \boldsymbol{\lambda}_{\boldsymbol{\theta}}}(\boldsymbol{x}_{\boldsymbol{\theta}}(\tau),\boldsymbol{u}_{\boldsymbol{\theta}}(\tau),\boldsymbol{\lambda}_{\boldsymbol{\theta}}(\tau)), \label{pmp.1}\\
	-\dot{\boldsymbol{\lambda}}_{\boldsymbol{\theta}}(\tau)& =\frac{\partial H}{\partial \boldsymbol{x}}(\boldsymbol{x}_{\boldsymbol{\theta}}(\tau),\boldsymbol{u}_{\boldsymbol{\theta}}(\tau),\boldsymbol{\lambda}_{\boldsymbol{\theta}}(\tau)) \label{pmp.2}, \\
	\boldsymbol{0}&=\frac{\partial H}{\partial \boldsymbol{u}}(\boldsymbol{x}_{\boldsymbol{\theta}}(\tau),\boldsymbol{u}_{\boldsymbol{\theta}}(\tau),\boldsymbol{\lambda}_{\boldsymbol{\theta}}(\tau)),\label{pmp.3}\\
	\boldsymbol{\lambda}_{\boldsymbol{{\theta}}}(T)&=\frac{\partial h_{\boldsymbol{p}}}{\partial \boldsymbol{x}}(\boldsymbol{x}_{\boldsymbol{\theta}}(T)) \quad\text{and} \quad \boldsymbol{x}_{\boldsymbol{\theta}}(0)=\boldsymbol{x}(0) \label{pmp.4}.
	\end{align}
\end{subequations}
Here,  (\ref{pmp.1}) is the dynamics; (\ref{pmp.2}) is the costate  ODE; (\ref{pmp.3}) is the input ODE, and (\ref{pmp.4}) is the boundary conditions.
Given $\boldsymbol{\xi}_{\boldsymbol{{\theta}}}$, one can always solve the corresponding   $\{\boldsymbol{\lambda}_{\boldsymbol{\theta}}(\tau)\,|\,0\leq \tau\leq T\}$ by integrating the costate ODE in  (\ref{pmp.2}) backward in time with the boundary condition in (\ref{pmp.4}).

Recall that our technical challenge  is to obtain the gradient $\frac{{\partial \boldsymbol{\xi}_{\boldsymbol{\theta}}}}{\partial \boldsymbol{\theta}}$. Towards this goal, we differentiate the above Pontryagin's Maximum Principle  in (\ref{pmp}) on both sides with respect to the system parameter $\boldsymbol{\theta}$,  yielding the following \emph{Differential Pontryagin's Maximum Principle}:
	\begin{subequations}\label{dpmp}
		\begin{align}
		\frac{d}{d\tau}(\frac{\partial \boldsymbol{x}_{\boldsymbol{\theta}}}{\partial\boldsymbol{\theta}})&=F(\tau)\frac{\partial \boldsymbol{x}_{\boldsymbol{\theta}}}{\partial\boldsymbol{\theta}}+G(\tau)\frac{\partial \boldsymbol{u}_{\boldsymbol{\theta}}}{\partial\boldsymbol{\theta}}+E(\tau),\label{dpmp.1}\\[4pt]
		{-}\frac{d}{d\tau}(\frac{\partial \boldsymbol{\lambda}_{\boldsymbol{\theta}}}{\partial\boldsymbol{\theta}})&=H_{xx}(\tau)\frac{\partial \boldsymbol{x}_{\boldsymbol{\theta}}}{\partial\boldsymbol{\theta}}+H_{xu}(\tau)\frac{\partial \boldsymbol{u}_{\boldsymbol{\theta}}}{\partial\boldsymbol{\theta}} \notag\\
		&\qquad\qquad\qquad
		+{F}(\tau)\tran \frac{\partial \boldsymbol{\lambda}_{\boldsymbol{\theta}}}{\partial\boldsymbol{\theta}} + H_{xe}(\tau),\label{dpmp.2}\\[4pt]
		\boldsymbol{0}&=H_{ux}(\tau)\frac{\partial \boldsymbol{x}_{\boldsymbol{\theta}}}{\partial\boldsymbol{\theta}}+H_{uu}(\tau)\frac{\partial \boldsymbol{u}_{\boldsymbol{\theta}}}{\partial\boldsymbol{\theta}}\notag\\
		&\qquad\qquad\qquad
		+{G}(\tau)\tran \frac{\partial \boldsymbol{\lambda}_{\boldsymbol{\theta}}}{\partial\boldsymbol{\theta}}+H_{ue}(\tau), \label{dpmp.3}\\[4pt]
		\frac{\partial \boldsymbol{\lambda}_{\boldsymbol{\theta}}}{\partial\boldsymbol{\theta}}(T)&=H_{xx}(T)\frac{\partial \boldsymbol{x}_{\boldsymbol{\theta}}}{\partial\boldsymbol{\theta}} + H_{xe}(T) \notag\\
		&\qquad\qquad\qquad
		\quad\text{and} \quad \frac{\partial \boldsymbol{x}_{\boldsymbol{\theta}}}{\partial\boldsymbol{\theta}}(0)=\boldsymbol{0}. \label{dpmp.4}
		\end{align}
	\end{subequations}
\noindent
The coefficient matrices in the above (\ref{dpmp})   are defined as
\begin{small}
	\begin{subequations}\label{hmaltionmatrix}
		\begin{align}
		&F(\tau) {=}\dfrac{\partial^2 {H}}{\partial \boldsymbol{\lambda}_{\boldsymbol{\theta}}\partial \boldsymbol{{x}}_{\boldsymbol{\theta}}},\quad G(\tau){=} \dfrac{\partial^2 {H}}{\partial \boldsymbol{\lambda}_{\boldsymbol{\theta}}\partial \boldsymbol{{u}}_{\boldsymbol{\theta}}},\quad  E(\tau){=}
		\dfrac{\partial^2 {H}}{\partial \boldsymbol{\lambda}_{\boldsymbol{\theta}}\partial \boldsymbol{\theta}}, \label{matFGE}\\[4pt]
		&H_{xx}(\tau){=}\frac{\partial^2 H}{(\partial \boldsymbol{{x}}_{\boldsymbol{\theta}})^2},  
		\,\, H_{xu}(\tau){=}\frac{\partial^2 H}{\partial  \boldsymbol{{x}}_{\boldsymbol{\theta}} \partial\boldsymbol{{u}}_{\boldsymbol{\theta}}},\,  H_{xe}(\tau) {=}\frac{\partial^2 H}{\partial \boldsymbol{{x}}_{\boldsymbol{\theta}}\partial \boldsymbol{\theta}},
		\label{matHx}\\[4pt]
		&H_{ux}(\tau)   {=}H_{xu}\tran(\tau),  \,\,\,\,  H_{uu}(\tau) {=}\dfrac{\partial^2 H}{(\partial \boldsymbol{{u}}_{\boldsymbol{\theta}})^2},  \quad  H_{ue} (\tau) {=}\dfrac{\partial^2 H}{\partial \boldsymbol{{u}}_{\boldsymbol{\theta}}\partial \boldsymbol{\theta}}\label{matHu},\\[4pt]
		&H_{xx}(T) {=}\dfrac{\partial^2 h_{\boldsymbol{p}}}{(\partial \boldsymbol{{x}}_{\boldsymbol{\theta}})^2}, \,\,   H_{xe}(T) {=}\dfrac{\partial^2 h_{\boldsymbol{p}}}{\partial \boldsymbol{{x}}_{\boldsymbol{\theta}}\partial \boldsymbol{\theta}}\label{matHT}.
		\end{align}
	\end{subequations}
\end{small}%
Once we obtain the optimal trajectory $\boldsymbol{\xi}_{\boldsymbol{\theta}}$  and the associated costate trajectory $\{\boldsymbol{\lambda}_{\boldsymbol{\theta}}(\tau)\,|\,0\leq \tau\leq T\}$ in (\ref{costate}),  all the above coefficient matrices in (\ref{hmaltionmatrix}) are known and their computation   is straightforward. 
Given the {Differential Pontryagin's Maximum Principle}  in (\ref{dpmp}), one can observe that these ODEs have a similar form to the original Pontryagin's Maximum Principle in (\ref{pmp}). Thus, if one thinks of $\frac{{\partial \boldsymbol{x}_{\boldsymbol{\theta}}}}{\partial \boldsymbol{\theta}}$ as a new state variable, $\frac{{\partial \boldsymbol{u}_{\boldsymbol{\theta}}}}{\partial \boldsymbol{\theta}}$ as a new control variable, and $\frac{{\partial \boldsymbol{\lambda}_{\boldsymbol{\theta}}}}{\partial \boldsymbol{\theta}}$ as a new costate variable, then the  {Differential Pontryagin's Maximum Principle} in (\ref{dpmp}) can be thought of as the  Pontryagin's Maximum Principle of a new LQR system, as investigated in \cite{jin2020pontryagin,jin2021safe}. By deriving the equivalent \emph{Raccati-type equations}, the  lemma below gives an efficient way to compute the trajectory gradient $\frac{{\partial \boldsymbol{\xi}_{\boldsymbol{\theta}}(\tau)}}{\partial \boldsymbol{\theta}}$, $0\leq \tau\leq T$, from  the above (\ref{dpmp}).

\begin{lemma}\label{theorem1}
	If ${H_{uu}(\tau)}$  in (\ref{matHu}) is invertible for all $0\leq \tau\leq T$,  define the following  differential equations for matrix variables $P(\tau)\in\mathbb{R}^{n\times n}$ and $W(\tau)\in\mathbb{R}^{n\times {(r+s)}}$:
	\begin{subequations}\label{cricc}
		\begin{align}
		-{\dot{{P}}}&={{Q}}(\tau)+{{A}}(\tau)\tran{{P}}+{{P}}{{A}}(\tau)-{{P}}{{R}}(\tau){{P}},\label{cricc.1}\\
		{\dot{{W}}}&={{P}}{{R}}(\tau){{W}}-{{A}}(\tau)\tran{{W}}-{{P}}{{M}}(\tau)-{{N}}(\tau)\label{cricc.2},
		\end{align}
	\end{subequations}
	with   ${{P}}(T)=H_{xx}(T)$ and ${{W}}(T)=H_{xe}(T).$
	Here, 
	\begin{subequations}\label{raccaticoefficient}
		\begin{align}
		{{A}} (\tau)&=F  -G  (H_{uu} )^{-1}H_{ux} ,\label{matA}\\[2pt]
		{{R}} (\tau)&=G  (H_{uu} )^{-1}G  \tran,\label{matR}\\[2pt]
		{{M}} (\tau)&=E  -G  (H_{uu} )^{-1}H_{ue} ,\label{matr}\\[2pt]
		{{Q}} (\tau)&=H_{xx}-H_{xu} (H_{uu} )^{-1}H_{ux} ,\label{matQ}\\[2pt]
		{{N}} (\tau)&=H_{xe} -H_{xu} (H_{uu} )^{-1}H_{ue} ,\label{matq}
		\end{align}
	\end{subequations}
	are all known given  (\ref{hmaltionmatrix}). The gradient of the optimal trajectory $\boldsymbol{\xi}_{\boldsymbol{\theta}}$, denoted as
	\begin{equation}
	\small
	\frac{{\partial \boldsymbol{\xi}_{\boldsymbol{\theta}}(\tau) }}{\partial \boldsymbol{\theta}}=\left(\frac{{\partial \boldsymbol{x}_{\boldsymbol{\theta}}}}{\partial \boldsymbol{\theta}}(\tau), \frac{{\partial \boldsymbol{u}_{\boldsymbol{\theta}}}}{\partial \boldsymbol{\theta}}(\tau) \right), \quad 0\leq \tau\leq T
	\end{equation}  is  obtained by integrating the following ODEs up to $\tau$:
		\begin{subequations}\label{citer}
			\begin{align}
			{\frac{{\partial \boldsymbol{u}_{\boldsymbol{\theta}}}}{\partial \boldsymbol{\theta}}}&{=}-(H_{uu}(\tau))^{-1}
			\Big(H_{ux}(\tau)\frac{\partial \boldsymbol{x}_{\boldsymbol{\theta}}}{\partial\boldsymbol{\theta}}(\tau)+H_{ue}(\tau)
			\nonumber\\
			&+{G(\tau)}\tran{{W}(\tau)}+{G(\tau)}\tran{{P}}(\tau)\frac{{\partial \boldsymbol{x}_{\boldsymbol{\theta}}}}{\partial \boldsymbol{\theta}}(\tau) \Big)
			\label{citer.1},\\[3pt]
			\frac{d}{d\tau}\left({\frac{{\partial \boldsymbol{x}_{\boldsymbol{\theta}}}}{\partial \boldsymbol{\theta}}}\right)&{=}
			F(\tau)\frac{{\partial \boldsymbol{x}_{\boldsymbol{\theta}}}}{\partial \boldsymbol{\theta}}(\tau)+G(\tau)\frac{{\partial \boldsymbol{u}_{\boldsymbol{\theta}}}}{\partial \boldsymbol{\theta}}(\tau)+E(\tau)\label{citer.2},
			\end{align}
		\end{subequations}
	with  $ \frac{\partial \boldsymbol{x}_{\boldsymbol{\theta}}}{\partial \boldsymbol{\theta}}{(0)}=\boldsymbol{0}$ in (\ref{dpmp.4}). Here, the matrices $P(\tau)$ and $W(\tau)$ are  solutions to  (\ref{cricc.1}) and (\ref{cricc.2}), respectively.
	
\end{lemma}
\noindent
The proof of Lemma \ref{theorem1} is given in  Appendix.
Lemma \ref{theorem1} states that for the optimal control system (\ref{ocsys}), the gradient of its  optimal trajectory $\boldsymbol{\xi}_{\boldsymbol{\theta}}$  with respect to the system parameter $\boldsymbol{\theta}$  can be obtained in two steps: first, integrate \eqref{cricc} backward in time to obtain  $P(\tau)$ and $W(\tau)$ for $0\leq\tau\leq T$; and second, obtain $\frac{{\partial \boldsymbol{\xi}_{\boldsymbol{\theta}}}}{\partial \boldsymbol{\theta}}({\tau})$ by integrating (\ref{citer}) forward in time.  Based on the  Differential  Pontryagin's Maximum Principle, Lemma \ref{theorem1} gives an efficient way to compute the gradient of an  optimal trajectory with respect to the  parameter in an optimal control system. By  Lemma \ref{theorem1}, one can obtain the derivative  of the trajectory point $\boldsymbol{\xi}_{\boldsymbol{\theta}}(\tau)$,  at any time   $0\leq \tau\leq T$, with respect to the system parameter $\boldsymbol{\theta}$, i.e.,  $\frac{{\partial \boldsymbol{\xi}_{\boldsymbol{\theta}  }}}{\partial \boldsymbol{\theta}}(\tau) $.

Additionally, we have the following comments on Lemma~\ref{theorem1}. First,   (\ref{cricc}) are Riccati-type equations, which are derived from  Differential Pontryagin's Maximum Principle in (\ref{dpmp}). Second, Lemma \ref{theorem1} requires the matrix $H_{uu}(\tau){=}\small\frac{\partial^2 H}{\partial \boldsymbol{{u}}_{\boldsymbol{\theta}}\partial\boldsymbol{{u}}_{\boldsymbol{\theta}}}$ in (\ref{matHu}) to be invertible, this is in fact a necessary condition \cite{jin2021safe} for the   differentiability of $\boldsymbol{\xi}_{\boldsymbol{\theta}}$. As we have mentioned at the beginning of this subsection, if $\boldsymbol{\xi}_{\boldsymbol{\theta}}$ satisfies  the second-order sufficient condition (i.e., is a locally unique optimal trajectory) for the optimal control problem (\ref{ocsys}), then $\boldsymbol{\xi}_{\boldsymbol{\theta}}$ is differentiable in $\boldsymbol{\theta}$ and   $H_{uu}(\tau)$ is automatically invertible (see \cite{jin2021safe} for the details and proofs).  A similar invertiblility requirement is common in  sensitivity analysis methods \cite{fiacco1976sensitivity,levysensitivity}, where  they analogously requires the Hessian matrix to be invertible in order to apply the  implicit function theorem \cite{krantz2012implicit}. Both our later experiments and other related existing work \cite{jin2020pontryagin,jin2021safe,amos2018differentiable} have empirically shown that the invertibility of $H_{uu}(\tau)$ is a mild condition and could be easily satisfied. 
With Lemma \ref{theorem1}, we summarize the overall algorithm of the Continuous PDP in Algorithm~\ref{algorithm1}.

\begin{algorithm2e}[h]
	\small 
	\SetKwInput{Initialization}{Initialization}
	\KwIn{keyframes $\mathcal{D}$ in (\ref{corrections}) and learning rate $\{\eta_{k}\}$. }
	\Initialization{initial parameter guess $\boldsymbol{\theta}_0$,}
	\For{$k=0,1,2,\cdots$}{
		
		\smallskip	
		Obtain the  optimal trajectory $\boldsymbol{{\xi}}_{\boldsymbol{\theta}_{k}}$ by solving the optimal control problem in (\ref{ocsys})  with current  $\boldsymbol{\theta}_k$\;
		\smallskip
		
		Obtain the costate trajectory $\{\boldsymbol{\lambda}_{\boldsymbol{\theta}_k}(\tau)\}$  by integrating   (\ref{pmp.2}) given (\ref{pmp.4})\;
		
		\smallskip
		
		Compute $\frac{{\partial \boldsymbol{\xi}_{\boldsymbol{\theta}}{(\tau_i)}}}{\partial \boldsymbol{\theta}}\big\rvert_{{\boldsymbol{\theta}}_k}$ using Lemma \ref{theorem1} for $i=1,2,\cdots N$\;
		\smallskip

		Compute   $\frac{\partial l}{\partial \boldsymbol{\xi}_{\boldsymbol{\theta}}{(\tau_i)}}\big\rvert_{\boldsymbol{\xi}_{\boldsymbol{\theta}_k}(\tau_i)}$ from (\ref{loss_warped})\;
		\smallskip
		
		Obtain $\frac{d L}{d \boldsymbol{\theta}}\rvert_{\boldsymbol{\theta}_{k}}$ using the chain rule (\ref{chainrule})\;
		\medskip

		Update $\boldsymbol{\theta}_{k+1}\leftarrow\proj_{\boldsymbol{\Theta}}\left(\boldsymbol{\theta}_k-\eta_k\frac{d L}{d \boldsymbol{\theta}}\bigr|_{\boldsymbol{\theta}_k}\right)$\;
	}
	\caption{Learning from  sparse demonstrations.} \label{algorithm1}
\end{algorithm2e}

\section{Numerical Experiments} \label{section_numericalexamples}
In this section, we  evaluate  different aspects of the proposed method using a two-link robot arm performing reaching tasks. The dynamics of a  robot arm  (moving horizontally) is \cite{spong2008robot}
\begin{equation}
M(\boldsymbol{q})\ddot{\boldsymbol{q}}+c(\boldsymbol{q},\boldsymbol{\dot{q}})=\boldsymbol{\tau},
\end{equation}
where $M(\boldsymbol{q})\in\mathbb{R}^{2\times2}$ is the inertia matrix, $c(\boldsymbol{q},\boldsymbol{\dot{q}})\in\mathbb{R}^{2}$ is the Coriolis term; $\boldsymbol{q}=[q_1,q_2]\tran\in\mathbb{R}^{2}$ is the joint angle vector, and $\boldsymbol{\tau}=[\tau_1, \tau_2]\tran\in\mathbb{R}^{2}$ is the joint toque vector. The  physical parameters for the dynamics are:  $m_1{=}2\text{kg}$ and $m_2{=}1\text{kg}$ for the mass of each link;   $l_1{=}1\text{m}$ and  $l_2{=}1\text{m}$ for the length of each link (assume mass is evenly distributed).  The state  and control vectors  are  $\boldsymbol{x}=[\boldsymbol{q},\dot{\boldsymbol{q}}]\tran\in\mathbb{R}^4$ and $\boldsymbol{u}=\boldsymbol{\tau}\in\mathbb{R}^2$, respectively.

For the task of  reaching  to a goal state $\boldsymbol{x}^{\text{g}}=[q_1^\text{g}, q_2^\text{g}, 0, 0]\tran\in\mathbb{R}^4$, we set the  cost function  (\ref{costfun})  as 
\begin{subequations}\label{exp.robotarm.cost}
	\begin{align}
	c(\boldsymbol{x},\boldsymbol{u},\boldsymbol{p})=&
	p_1(q_1-q_1^\text{g})^2+p_2(q_2-q_2^\text{g})^2+\nonumber\\
	&
	p_3\,\dot q_1^2+p_4\,\dot q_2^2+0.5\norm{\boldsymbol{u}}^2,\\
	h(\boldsymbol{x},\boldsymbol{p})=&
	p_1(q_1{-}q_1^\text{g})^2{+}p_2(q_2{-}q_2^\text{g})^2+
	p_3\,\dot q_1^2{+}p_4\,\dot q_2^2.
	\end{align}
\end{subequations}
with the tunable parameter $\boldsymbol{p}=[p_1, p_2, p_3, p_4]\tran\in\mathbb{R}^4$. Note that (\ref{exp.robotarm.cost}) is a weighted distance-to-goal function with a fixed weight to  $\norm{\boldsymbol{u}}^2$, because otherwise, learning all  weights will lead to scaling ambiguity  \cite{jin2018inverse}. We set the goal state $\boldsymbol{x}^{\text{g}}=[\frac{\pi}{2}, 0, 0, 0]\tran$, and the initial  state $\boldsymbol{x}(0)=[-\frac{\pi}{2}, \frac{3\pi}{4}, -5, 3]\tran$.

For  parametric time-warping function  (\ref{para_warp}), we simply use
\begin{equation}\label{beta}
t=w_{\boldsymbol{\beta}}(\tau)=\beta\tau,
\end{equation}
with $\Omega_{\boldsymbol{\beta}}=\{\beta \,| \,\beta>0\}$ (more complex time-warping functions will be used  later). The overall parameter  to be tuned is  $\boldsymbol{\theta}=[\boldsymbol{p}\tran,\beta]\tran\in\mathbb{R}^5$.
The task-space mapping   (\ref{interface}) is
\begin{equation}
\boldsymbol{q}=\boldsymbol{g}(\boldsymbol{x},\boldsymbol{u}),
\end{equation}
meaning that the keyframe only includes the position information.   For the  discrepancy  loss (\ref{loss_warped}), we use the squared  $l_2$ norm:
\begin{equation}\label{lossd}
L (\boldsymbol{\xi}_{\boldsymbol{\theta}},\mathcal{D})=\sum\nolimits_{i=1}^{N} \norm{\boldsymbol{q}^*(\tau_i)-
	\boldsymbol{g}\big(\boldsymbol{\xi}_{\boldsymbol{\theta}}(\tau_i)\big)}^2.
\end{equation}
In the following experiments, we evaluate different aspects of the method and provide analysis for each evaluation.

\subsection{Different Number of Keyframes}\label{experiment1.1}

\begin{figure}[h]
	\centering	\includegraphics[width=0.9\columnwidth]{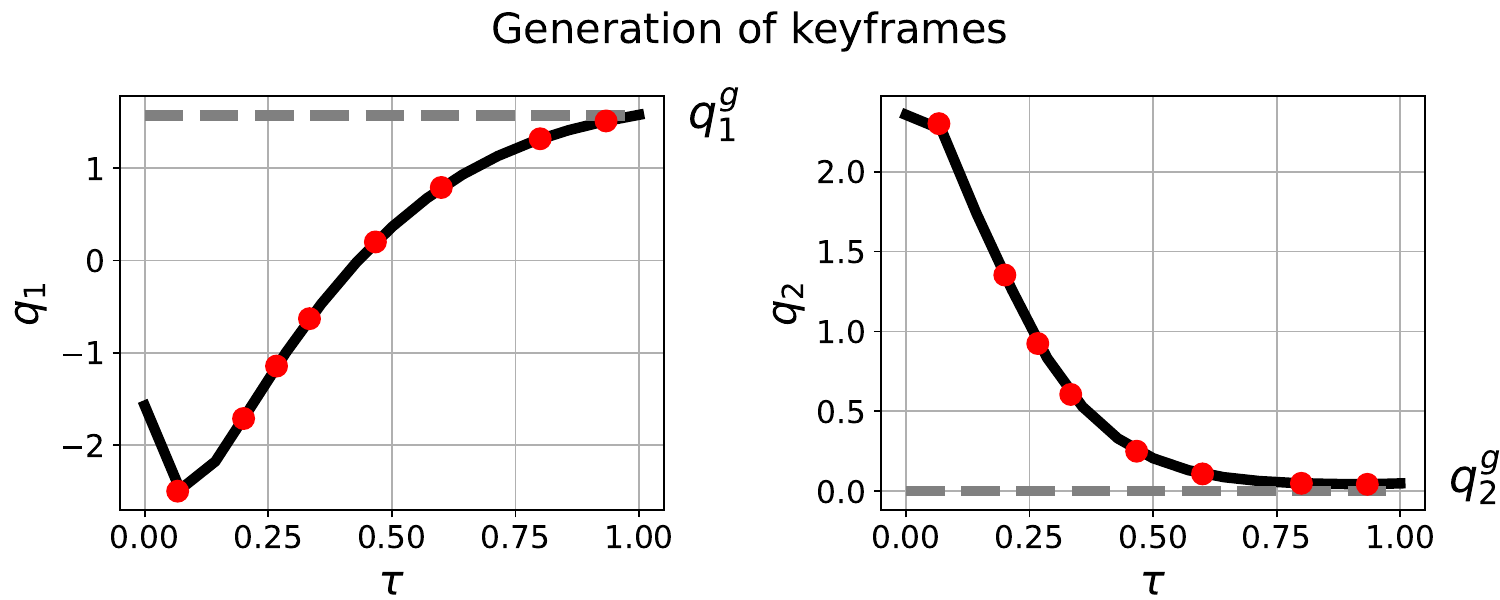}
	\caption{Generating  keyframes (marked as  red dots) from an optimal trajectory with $\boldsymbol{\theta}^{\text{true}}$. The gray dashed lines label the goal pose for each joint, i.e.,  $[q_1^{\text{g}}, q_2^{\text{g}}]\tran=[\pi/2, 0]\tran$.}
	\label{fig.robotarm.sparseselection}
\end{figure}

First, we evaluate the  performance of the proposed method for learning from  different numbers of  keyframes.      $\mathcal{D}$ is generated from  known/true cost  and time-warping functions. Given  
\begin{equation}\label{truetheta}
\boldsymbol{\theta}^{\text{true}}= [ 3, 3, 3, 3, 5]\tran,
\end{equation} the robot optimal trajectory is computed by solving the  optimal control problem   (\ref{ocsys}), shown in Fig. \ref{fig.robotarm.sparseselection}. Then, we select some  points (red dots) from Fig.  \ref{fig.robotarm.sparseselection} as our keyframes  $\mathcal{D}$,  listed  in Table \ref{table.robotarm.truesaprsedemo}.  We  evaluate the  performance of the proposed  method to  recover $\boldsymbol{\theta}^{\text{true}}$ given different numbers of the  keyframes. The learning rate is  $\eta=0.1$, and the  initial  $\boldsymbol{\theta}_0$ is randomly given. For each evaluation case, we have run the experiment for 10 trials with different random seeds  for  $\boldsymbol{\theta}_0$.

\begin{figure}[h]
	\centering	\includegraphics[width=1\columnwidth]{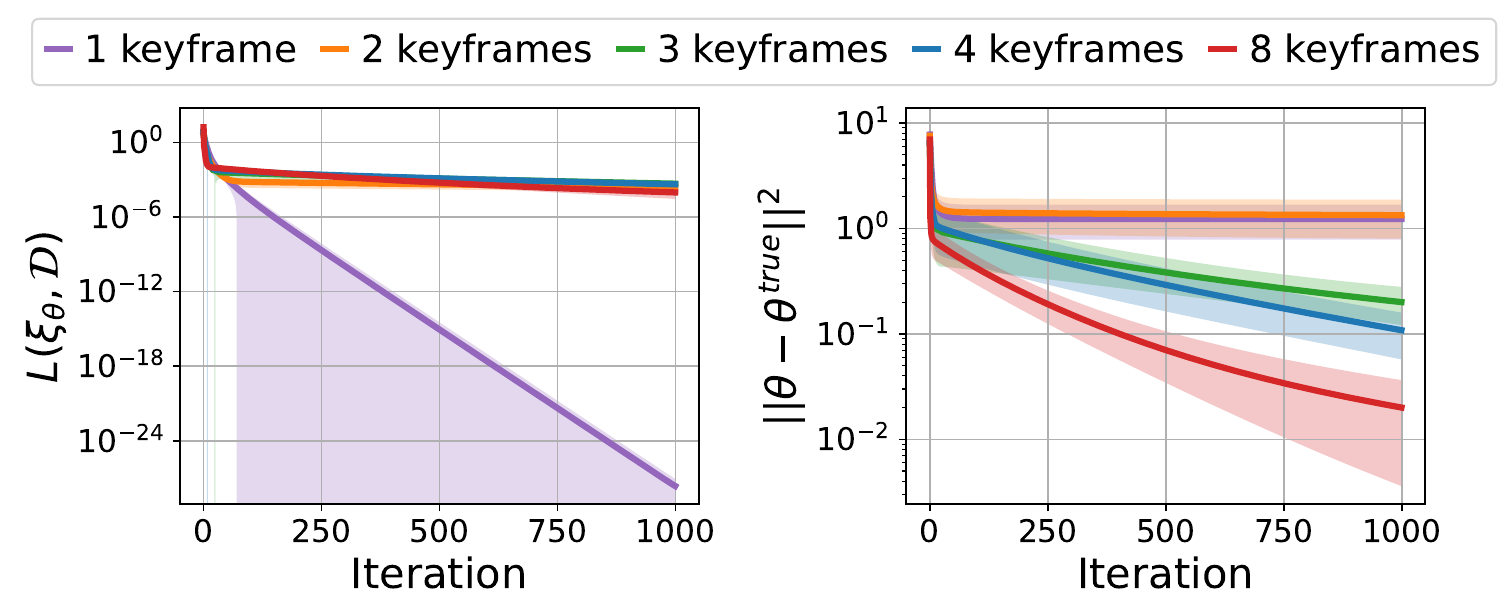}
	\caption{Learning from different numbers of keyframes. The left panel shows the loss (\ref{lossd}) versus iteration, the right shows the parameter error $\norm{\boldsymbol{{\theta}}_{k}-\boldsymbol{{\theta}}^{\text{true}}}^2$ versus iteration.  The solid line and shaded area denote the mean and standard derivation over all 10 trials.}
	\label{fig.robotarm.trueresults}
\end{figure}

\begin{table}[h]
	\centering
	\caption{Keyframes  $\mathcal{D}$ generated in Fig. \ref{fig.robotarm.sparseselection}}.
	\begin{tabular}{ccc}
		\toprule
		No. &Time stamp $\tau_i$ ($T=1$)  &  Keyframe $\boldsymbol{y}^*(\tau_{i})$ \\
		\midrule
		\#1 &	$\tau_1=0.067$s &   $\boldsymbol{q}^*({\tau_1})=[-2.497,  2.301]$  \\
		\#2 &	$\tau_2=0.2$s &  $\boldsymbol{q}^*({\tau_2})= [-1.71,  1.353]$ \\
		\#3 &	$\tau_3=0.267$s &   $\boldsymbol{q}^*({\tau_3})=[-1.142, 0.924]$  \\
		\#4 &	$\tau_4=0.333$s &  $\boldsymbol{q}^*({\tau_4})=[-0.629,  0.606]$ \\
		\#5 &	$\tau_5=0.467$s &  $\boldsymbol{q}^*({\tau_5})= [ 0.201,  0.25 ]$ \\
		\#6 &	$\tau_6=0.6$s &  $\boldsymbol{q}^*({\tau_6})= [ 0.791,  0.108]$ \\
		\#7 &	$\tau_7=0.8$s &  $\boldsymbol{q}^*({\tau_7})= [ 1.319,  0.049]$ \\
		\#8 &	$\tau_8=0.933$s &  $\boldsymbol{q}^*({\tau_8})= [ 1.512, 0.043]$ \\
		\bottomrule
	\end{tabular}
	\label{table.robotarm.truesaprsedemo}
\end{table}
We choose different numbers of  keyframes from Table \ref{table.robotarm.truesaprsedemo} to learn  the  time-warping and  cost functions, and the results are  in Fig. \ref{fig.robotarm.trueresults}. The left panel of Fig. \ref{fig.robotarm.trueresults} shows the loss (\ref{lossd}) versus iteration, and the right  shows the  parameter error $\norm{\boldsymbol{\theta}-\boldsymbol{\theta}^{\text{true}}}^2$
versus iteration. 

Fig. \ref{fig.robotarm.trueresults} shows that when the number of keyframes  $N\geq3$ (blue, green, and red lines), the loss $L (\boldsymbol{\xi}_{\boldsymbol{\theta}_k},\mathcal{D})$ and   parameter error $\norm{\boldsymbol{{\theta}}_{k}-\boldsymbol{{\theta}}^{\text{true}}}^2$  converge to zeros,  indicating that both the   cost  and time-warping functions  are successfully learned.   When $N\leq 2$, while the loss converges to zero, $\boldsymbol{\theta}_k$ does not converge to    $\boldsymbol{\theta}^{\text{true}}$  (orange and purple lines in the right panel). This indicates when $N\leq 2$, {there are multiple  cost  and time-warping functions, besides $\boldsymbol{\theta}^{\emph{true}}$,  that  lead to the given  keyframes}. In other words, with fewer keyframes, we cannot \emph{uniquely} determine the cost and time-warping functions, as they are over-parameterized relative to given keyframes. Intuitively, to uniquely determine $\boldsymbol{\theta}^{\text{true}}$, the number of constraints imposed by the given keyframes, $oN$ (recall $o$ is the dimension of $\boldsymbol{g}()$), should be no less than the number of all unknown parameters, $r{+}s$, that is, $N\geq \frac{r{+}s}{o}$.
Please refer to Section \ref{section_discussions.a} for  more analysis.

From the right panel of Fig. \ref{fig.robotarm.trueresults}, we also observe that different numbers of keyframes ($N\geq  3$) also influence the converge rate. For instance, the convergence rate with 8 keyframes (red line) is faster than that of 4 keyframes (blue line). Since the proposed method  updates the cost and time-warping functions by finding the deepest descent direction of loss, thus, the more keyframes are given, the better informed the gradient direction will be,  making the convergence to the true parameters faster.

\begin{figure}[h]
	\centering	\includegraphics[width=0.99\columnwidth]{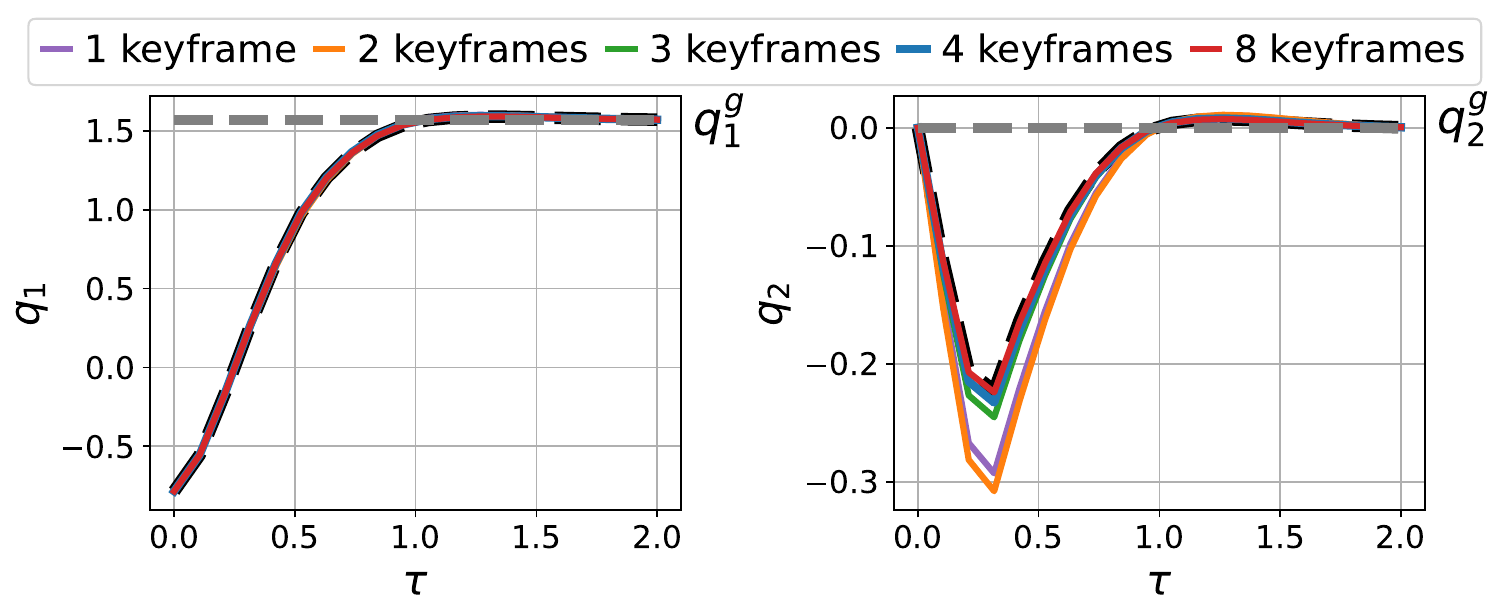}
	\caption{Generalization  of the learned cost function given new  initial condition $\boldsymbol{x}(0)$ and new   horizon $T$. The gray dashed lines mark the goal  for each joint $[q_1^{\text{g}}, q_2^{\text{g}}]\tran=[\pi/2, 0]\tran$.}
	\label{fig.robotarm.truegeneralize}
\end{figure}

Lastly, we test the generalization of the learned  cost  and time-warping functions, by setting the robot arm to new initial state $\boldsymbol{x}(0)=[-\frac{\pi}{4}, 0, 0,0]\tran$ and new  horizon  $T{=}2$ (both are very different from the ones  in  learning). The generated motion using the learned $\boldsymbol{\theta}$ (mean value over all trials) is shown in Fig. \ref{fig.robotarm.truegeneralize}, where  we have also plotted the  trajectory  of  $\boldsymbol{\theta}^{\text{true}}$ for reference. To compare  the generalization performance, we  compute the distance between the final state $\boldsymbol{x}(T)$ of the generalized motion and the goal state  $\boldsymbol{x}^{\text{g}}=[\frac{\pi}{2}, 0, 0, 0]\tran$, and  list the  results   in Table \ref{table.distancetogoal}.
Both Fig. \ref{fig.robotarm.truegeneralize}  and Table \ref{table.distancetogoal} show that the learned $\boldsymbol{\theta}$  enables  to generate new   motion in unseen conditions. Further, Table \ref{table.distancetogoal} shows that the increasing  keyframes could lead to better generalization. 
Notably, we  see that although the learned $\boldsymbol{\theta}$s  from  1 or 2 keyframes are different from $\boldsymbol{\theta}^{\text{true}}$, they can still obtain fair generalization.  This could be due to the formulation of the distance-to-goal features   (\ref{exp.robotarm.cost}). Although the learned  weight vector $\boldsymbol{\theta}$ is different from $\boldsymbol{\theta}^{\text{true}}$, the distance-to-goal features largely contributes to  a similar   performance. We will show later  in Section \ref{section.learningneural} that when (\ref{exp.robotarm.cost}) is replaced with a  neural cost function, fewer keyframes will lead to   poor generalization. Thus, for the same number of keyframes,  different cost function formulations could lead to different generalization abilities.   But as we will see in Section \ref{section.learningneural},  a common observation is that the more keyframes are given, the better the generalization will be for the learned cost function.

\begin{table}[h]
	\centering
	\caption	{Distance between the final state $\boldsymbol{x}(T)$ of  the generalized motion and the  goal  $\boldsymbol{x}^{\text{g}}$, i.e.,  $\norm{\boldsymbol{x}(T)-\boldsymbol{x}^{\text{g}}}$.}
	\begin{tabular}{ll}
		\toprule
		The learned $\boldsymbol{\theta}$ (mean value) from  &   $\norm{\boldsymbol{x}(T)-\boldsymbol{x}^{\text{g}}}$  \\
		\midrule
		1 keyframe &   $0.00581$  \\
		2 keyframes &  $0.00580$ \\
		3 keyframes &  $0.00392$ \\
		4 keyframes &  $0.00382$ \\
		8 keyframes &  $0.00358$ \\
		\textbf{True $\boldsymbol{\theta}^{\text{true}}$} &  $\boldsymbol{0.00346}$ \\
		\bottomrule
	\end{tabular}
	\label{table.distancetogoal}
\end{table}

\subsection{Non-optimal Keyframes  }\label{section.robotarm.nonoptimal}
\begin{figure}[h]
	\centering	\includegraphics[width=1\columnwidth]{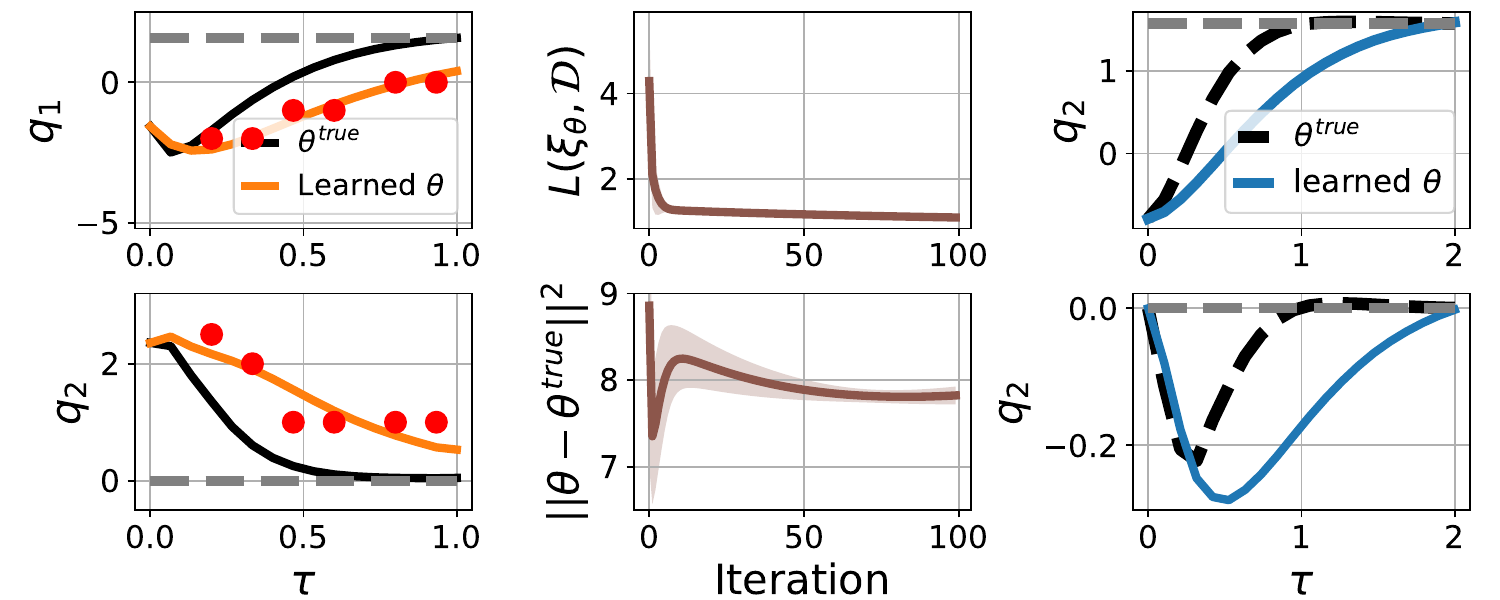}
	\caption{Learning from non-optimal keyframes. The first column shows the given  keyframes (red dots), which   deviate from the optimal trajectory (black lines), and the reproduced trajectory (orange lines) from the learned $\boldsymbol{\theta}$ (mean value over all 10 trials). The second column shows the  loss and parameter error versus iteration; the solid line and shaded area denote the mean and standard derivation over all 10 trials.  The third column shows the generalization of the learned $\boldsymbol{\theta}$ to new  initial condition $\boldsymbol{x}(0)$ and new time horizon $T$. The gray dashed lines in the first and third columns mark the goal  for each joint $[q_1^{\text{g}}, q_2^{\text{g}}]\tran=[\pi/2, 0]\tran$. As calculated in Table \ref{table.distancetogoal}, $\norm{\boldsymbol{x}(T)-\boldsymbol{x}^{\text{g}}}$ for the generalized motion in the third column is 0.107.}
	\label{fig.robotarm.randomresults}
\end{figure}

Next, we evaluate the performance of the proposed method given non-optimal keyframes. This emulates the situation where a demonstration could be polluted by biased sensing error, noise, hardware error, etc. We select keyframes $\mathcal{D}$ by corrupting each keyframe in Fig. \ref{fig.robotarm.trueresults} with a biased error, as shown in the first column (red dots) of Fig. \ref{fig.robotarm.randomresults}.  We evaluate the performance of the method given such biased keyframes. The other experiment settings follow the previous experiment. We have run each experiment for 10 trials with different random seeds for the initial $\boldsymbol{\theta}_0$.

In Fig. \ref{fig.robotarm.randomresults}, the  loss and parameter error versus iteration  are shown in the top and bottom panels of the second column, respectively. The solid line and shaded area denote the mean and standard derivation over all 10 trials.  We  use the learned  $\boldsymbol{\theta}$ to reproduce the optimal trajectory of the robot, which is shown  in the first column (orange lines). In the third column, we test the generalization of the learned $\boldsymbol{\theta}$ to the new  initial condition $\boldsymbol{x}(0)=[-\frac{\pi}{4}, 0, 0,0]\tran$ and new  horizon $T=2$. Here, we also compare with the  trajectory  of  $\boldsymbol{\theta}^{\text{true}}$ (dashed black lines). From  Fig. \ref{fig.robotarm.randomresults}, we have the following  comments.

Since the keyframes in the first column are non-optimal, there does not exist a  $\boldsymbol{\theta}$ that \emph{exactly} corresponds to those non-optimal keyframes. Thus, the loss in the second column does not converge to zero. Despite those,  the method still finds a $\boldsymbol{\theta}$ such that its produced trajectory is \emph{closest} to the keyframes, as shown by orange lines in the first column.
The second column shows that  the learned $\boldsymbol{\theta}$ is different from   $\boldsymbol{\theta}^{\text{true}}$.

The generalization in the third column shows that given the new initial condition and horizon, the generalized motion still approaches the goal, and the final distance of the generalized motion to the goal is $\norm{\boldsymbol{x}(T)-\boldsymbol{x}^{\text{g}}}=0.107$, which is larger compared to the one in Table \ref{table.distancetogoal}. 

\subsection{Different  Time-Warping Functions}\label{experiment1.2} 
In this set of experiments, we test the learning performance of using polynomial time-warping functions of different complexity. The keyframes $\mathcal{D}$ are the red dots in the first column of Fig. \ref{fig.robotarm.randomresults}. For each polynomial time-warping function, we have run the experiment for 10 trials with different random seeds for  initial $\boldsymbol{\theta}_0$. Other experiment settings follow the previous one. The results are summarized in Table \ref{table.timewarping}. Here, the first column shows the learned time-warping functions; the second column is the   final converged losses, and the statistics (mean+standard deviation) are over 10 trials. We have the following comments.

\begin{table}[h]
	\centering
	\caption	{Learning with different time-warping functions}
	\begin{tabular}{ll}
		\toprule
		Learned  time-warping function   &   $\min  L (\boldsymbol{\xi}_{\boldsymbol{\theta}},\mathcal{D})$ (mean$\pm$std) \\
		\midrule
		$t=2.55\tau$ &   $1.017\pm0.014$  \\
		$t=2.90\tau-0.55\tau^2$ &  $0.876\pm 0.008$ \\
		$t=2.94\tau+0.28\tau^2 -0.88\tau^3$ &  $0.831\pm 0.006$ \\
		$t=2.89\tau +0.53\tau^2-0.49\tau^3-0.60\tau^4$ &  $0.822\pm0.002$ \\
		\bottomrule
	\end{tabular}
	\label{table.timewarping}
\end{table}

Table \ref{table.timewarping}  shows that a higher order of polynomial time-warping function leads to the lower final loss.  This is because a higher degree polynomial introduces additional degrees of freedom, which enable to represent more complex time mapping and contribute to further decreasing the loss. 
Meanwhile,    Table \ref{table.timewarping}  shows 
that (i) the first-order terms in all learned time-warping polynomials are similar, (ii) the higher-order terms are relatively small compared to the first-order term, and (iii) adding higher-order terms to the time-warping polynomial only decreases a small amount of final loss. All those observations indicate that the first-order term dominates the final performance. 
We may conclude that in practice, it is preferable to  start with a simplified time-warping function. The subsequent experiments will use the first-order time-warping function for simplicity.

\subsection{Learning Neural  Cost Functions} \label{section.learningneural}
In this session, we test the ability of the proposed method to learn neural-network cost functions. This is useful if a weight-feature cost function formulation cannot be specified due to the lack of prior knowledge. We set the cost function  (\ref{exp.robotarm.cost}) with the following neural-network cost function,
\begin{equation}\label{exp.robotarm.neuralcost}
\begin{aligned}
c(\boldsymbol{x},\boldsymbol{u},{\boldsymbol{p}})&=\boldsymbol{\phi}_{\boldsymbol{p}}\tran(\boldsymbol{x})\boldsymbol{\phi}_{\boldsymbol{p}}(\boldsymbol{x})+0.05\norm{\boldsymbol{u}}^2,\\ h(\boldsymbol{x},{\boldsymbol{p}})&=\boldsymbol{\phi}_{\boldsymbol{p}}\tran(\boldsymbol{x})\boldsymbol{\phi}_{\boldsymbol{p}}(\boldsymbol{x}),
\end{aligned}
\end{equation}
where $\boldsymbol{\phi}_{\boldsymbol{p}}(\boldsymbol{x})$ is a \emph{4-8}  fully-connected neural network \cite{nielsen2015neural} (i.e., 4-neuron input layer and 8-neuron output layer), and  $\boldsymbol{p}\in\mathbb{R}^{40}$ is the parameter  of  the neural   network, i.e., all weight matrices and bias vectors. Note that (\ref{exp.robotarm.neuralcost}) uses dot product in the output layer of the neural network to guarantee the positiveness of the cost function. The time-warping polynomial has the degree of one. We use the  keyframes  in  Fig. \ref{fig.robotarm.sparseselection} (also in Table \ref{table.robotarm.truesaprsedemo}).  Other experiment settings are the same as the previous ones. In each evaluation case below,  we have run the experiment for 10 trials with different random seeds for the initial $\boldsymbol{\theta}_0$.

\begin{figure}[h]
	\begin{subfigure}{.159\textwidth}
		\centering
		\includegraphics[width=\linewidth]{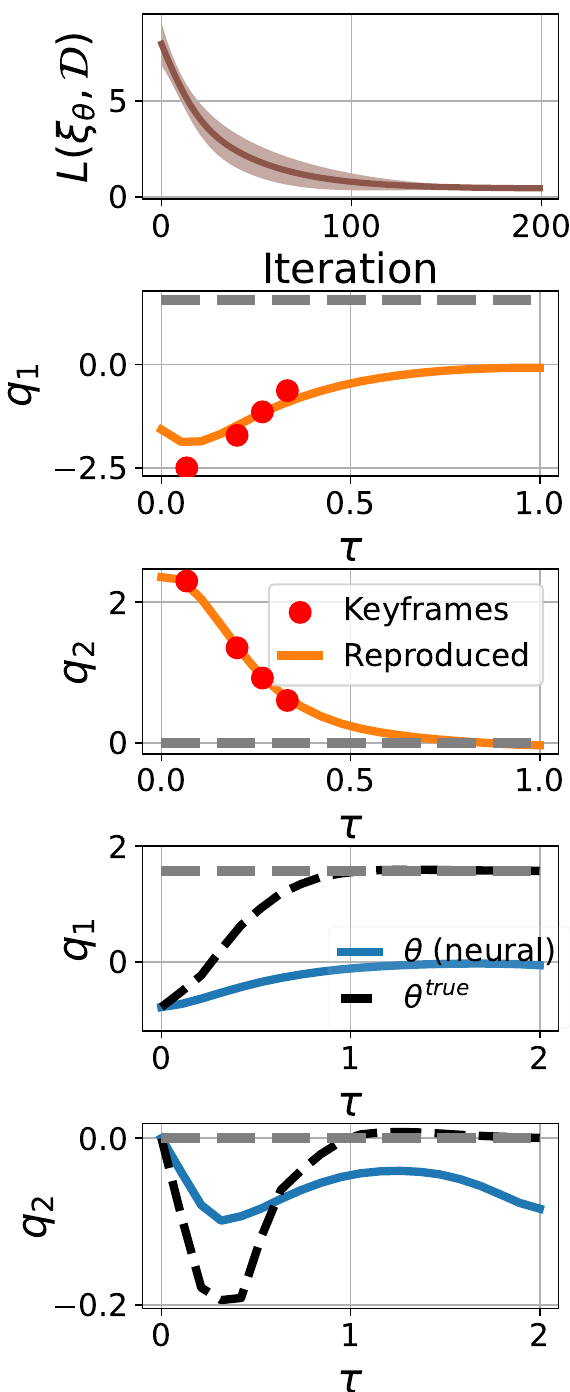}
		\caption{Case 1}
		\label{fig.robotarm.neural.4}
	\end{subfigure}
	\begin{subfigure}{.159\textwidth}
		\centering
		\includegraphics[width=\linewidth]{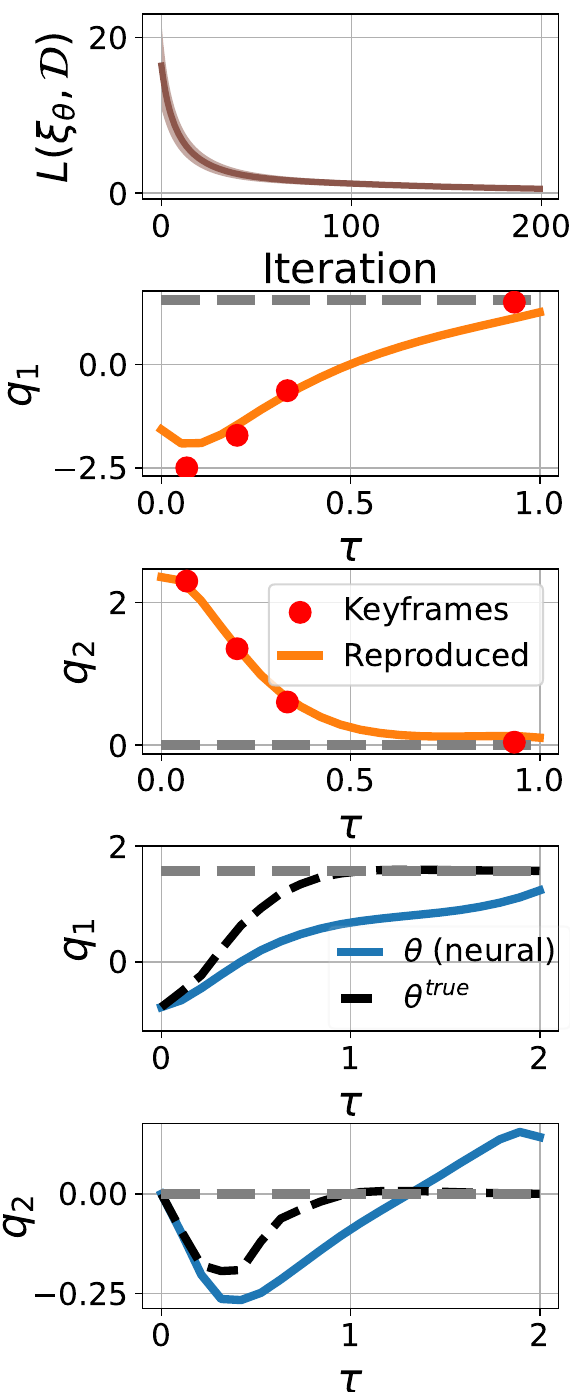}
		\caption{Case 2}
		\label{fig.robotarm.neural.5}
	\end{subfigure}
	\begin{subfigure}{.159\textwidth}
		\centering
		\includegraphics[width=\linewidth]{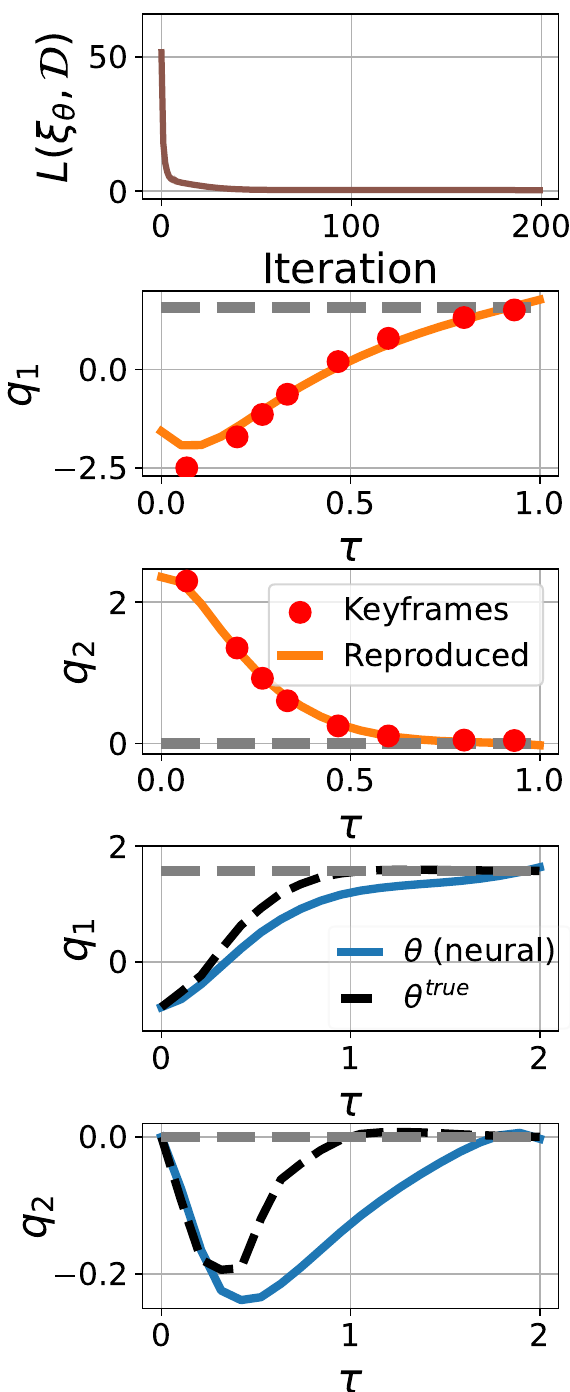}
		\caption{Case 3}
		\label{fig.robotarm.neural.9}
	\end{subfigure}
	\caption{Learning neural cost functions from  keyframes. Three cases of  keyframes are used,  shown in the red dots in each column. In each case, the first row shows   loss versus iteration (the solid line and shaded area denote the mean and standard derivation over all 10 trials). The second and third rows show the reproduced trajectories (orange lines) of the learned $\boldsymbol{\theta}$. The fourth and fifth rows show the generalization (blue lines) of the learned $\boldsymbol{\theta}$ to new  initial state and   horizon, and  the  motion (dashed black lines)  of  $\boldsymbol{\theta}^{\text{true}}$  is also plotted for reference.  {In second-fifth rows, the gray dashed lines mark the goal for each joint $[q_1^{\text{g}}, q_2^{\text{g}}]\tran=[\pi/2, 0]\tran$. } }
	\label{fig.robotarm.neural}
\end{figure}

We plot the learning and generalization results in  Fig. \ref{fig.robotarm.neural}. We test with three cases of the keyframes shown in red dots in the second row, and the corresponding results are shown in each  column. In each case,  the first row shows the loss versus iteration; and second and third rows show the reproduced trajectories (orange lines) by the learned  cost  and time-warping functions; and the fourth and fifth rows show  the generalization (blue lines) of the learned  cost function to new initial state $\boldsymbol{x}(0)=[-\frac{\pi}{4}, 0, 0,0]\tran$ and  new horizon $T=2$. The  motion (dashed black lines) of $\boldsymbol{\theta}^{\text{true}}$
is also plotted for reference. In Table \ref{table.distancetogoal2}, we compute the  distance between $\boldsymbol{x}(T)$ of the generalized motion  and the goal  $\boldsymbol{x}^{\text{g}}=[\frac{\pi}{2}, 0, 0, 0]\tran$ to measure the generalization performance. We have the following comments.

\begin{table}[h]
	\centering
	\caption	{  Distance of  $\boldsymbol{x}(T)$ of   the generalized motion (in  the fourth and fifth rows in Fig. \ref{fig.robotarm.neural}) to   goal  $\boldsymbol{x}^{\text{g}}=[\frac{\pi}{2}, 0, 0, 0]\tran$.}
	\begin{tabular}{ll}
		\toprule
		The learned $\boldsymbol{\theta}$ (mean value over all trials)  &   $\norm{\boldsymbol{x}(T)-\boldsymbol{x}^{\text{g}}}$  \\
		\midrule
		Case 1 &   $1.638$  \\
		Case 2 &  $0.717$ \\
		Case 3 &  $0.388$ \\
		\textbf{True $\boldsymbol{\theta}^{\text{true}}$} &  $\boldsymbol{0.00346}$ \\
		\bottomrule
	\end{tabular}
	\label{table.distancetogoal2}
\end{table}

First,  compared to the  distance-to-goal cost   (\ref{exp.robotarm.cost}), the  neural cost    (\ref{exp.robotarm.neuralcost}) is \emph{goal-blind}, meaning that the goal $\boldsymbol{q}^{\text{g}}=[\frac{\pi}{2}, 0]\tran$ is not encoded in the neural cost function before training. Thus, it is crucial for the robot to learn a goal-encoded neural  cost  for the success of the task. Case 1  and Case  2   use four keyframes to  learn a  cost function. The results in fourth and fifth rows of Fig. \ref{fig.robotarm.neural} and in Table \ref{table.distancetogoal2} indicate that  Case 2 has a better generalization  than  Case 1 does: Case 2 has a final distance $\norm{\boldsymbol{x}(T)-\boldsymbol{x}^{\text{g}}}{=}0.717$, while 
Case 1 has $\norm{\boldsymbol{x}(T)-\boldsymbol{x}^{\text{g}}}{=}1.638$.  This is because the keyframes in Case 1  are mainly clustered at the beginning of motion, and thus cannot provide sufficient information about the final goal. In contrast, Case 2 has a keyframe at the goal, and thus the learned neural cost function captures such goal information.

Second, we add more keyframes in Case 3.  It shows that more keyframes  lead to  better generalization of the learned neural cost function: the final distance is $\norm{\boldsymbol{x}(T)-\boldsymbol{x}^{\text{g}}}=0.388$,
which is better than those in Case 2 and Case 1. 

Lastly,  the learned neural cost function in Case 2 or Case 3, while controlling the robot  to approach the goal, has a trajectory that is different from the true one (black dashed lines). This manifests the  generalizability of learning cost functions. We also note that the neural cost function  (\ref{exp.robotarm.neuralcost}) is over-parameterized, relative to the fewer given keyframes.  Despite this, the learned neural cost still shows a fair generalization to new motion conditions, given a proper selection of keyframes.

\subsection{Comparison with Related Methods}\label{experiment1.3}
In this session, we compare the proposed method with the related work.
For all comparisons  below, the learning process uses the  keyframe data in Fig. \ref{fig.robotarm.sparseselection} (Table \ref{table.robotarm.truesaprsedemo}). The generalization  is tested by setting the robot  to a new initial condition $\boldsymbol{x}(0)=[-\frac{\pi}{4}, 0, 0,0]\tran$ and a new time horizon $T=2$. 
Other settings follow the previous experiments if not explicitly stated.

\subsubsection{Comparison with Kinematic  Learning  \cite{akgun2012keyframe}}\label{exp.compare.keyframes} 
\begin{figure}[h]
	\begin{subfigure}{.159\textwidth}
		\centering
		\includegraphics[width=\linewidth]{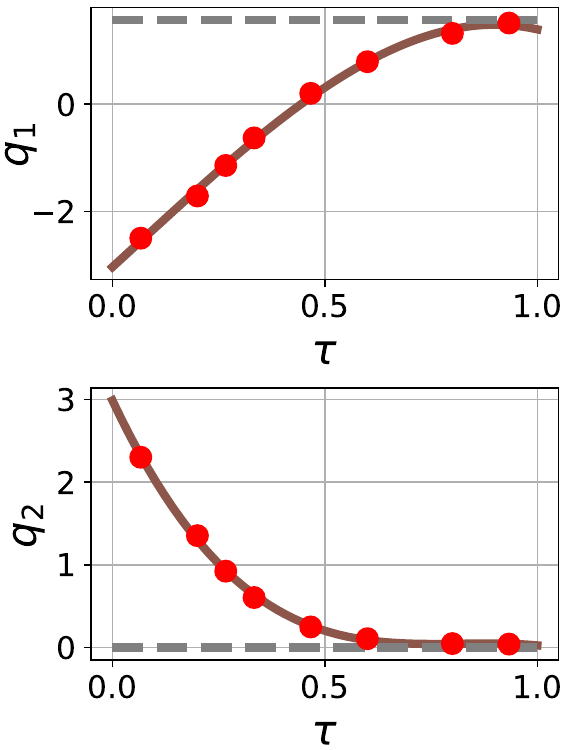}
		\caption{Spline fitting to the keyframes \cite{akgun2012keyframe}}
		\label{fig.compare.keyframe.1}
	\end{subfigure}
	\begin{subfigure}{.159\textwidth}
		\centering
		\includegraphics[width=\linewidth]{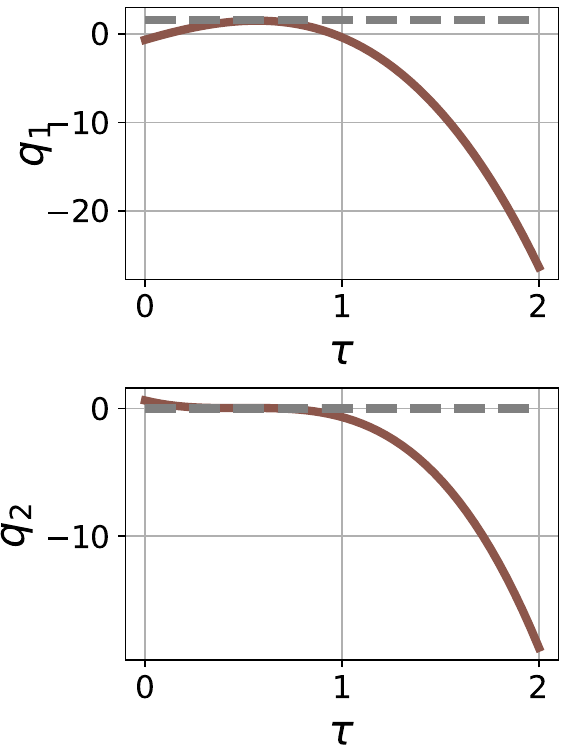}
		\caption{Generalization of the fitted spline  \cite{akgun2012keyframe}}
		\label{fig.compare.keyframe.2}
	\end{subfigure}
	\begin{subfigure}{.159\textwidth}
		\centering
		\includegraphics[width=\linewidth]{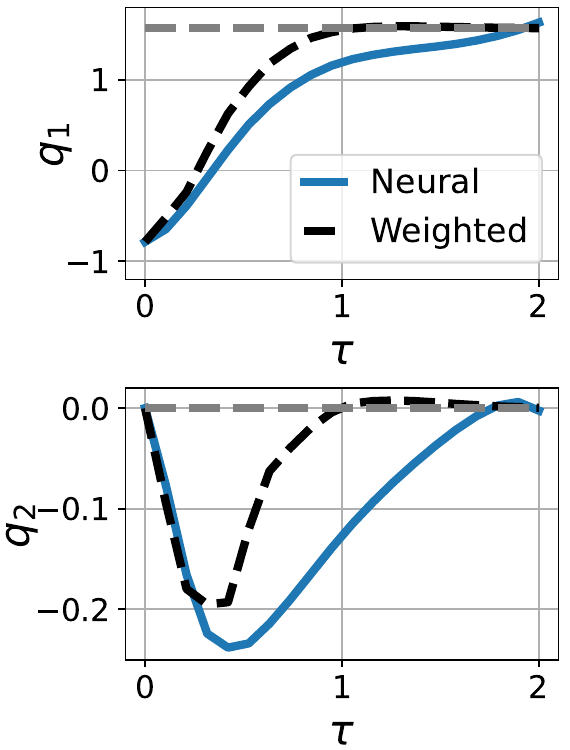}
		\caption{Generalization for the proposed method}
		\label{fig.compare.keyframe.3}
	\end{subfigure}
	\caption{Comparison between \cite{akgun2012keyframe} and the proposed method. (a) is the  spline model fitted to  keyframes; (b) is the generalization of the fitted spline in new motion condition (new $\boldsymbol{x}(0)$ and new $T$); and (c) is the generalization of the learned  weighted cost function (\ref{exp.robotarm.cost}) and learned neural cost function (\ref{exp.robotarm.neuralcost}) in the previous experiments. The gray dashed lines mark the goal  of each joint $[q_1^{\text{g}}, q_2^{\text{g}}]\tran=[\pi/2, 0]\tran$.  The final distance $\norm{\boldsymbol{x}(T)-\boldsymbol{x}^{\text{g}}}$ of the generalized motion is  33.670 for the fitted spline in (b),  0.388 for the learned neural cost function in (c), and 0.00358 for the learned  weighted cost function  in (c).
	} 
	\label{fig.compare.keyframe}
\end{figure}

Following \cite{akgun2012keyframe}, we  fit the    keyframes  in Table \ref{table.robotarm.truesaprsedemo} with   a fifth-order spline, as shown in the brown lines in Fig. \ref{fig.compare.keyframe.1}. The fitted spline  is then used to generalize the robot motion in the new  condition (i.e., a new initial condition and a new   horizon). To do this, following the idea of \cite{akgun2012keyframe},  we  compare which given keyframe  is closest to the new $\boldsymbol{x}(0)$, then from which  we perform  extrapolation based on the fitted spline to generate the new  trajectory over the new horizon $T=2$. The generated    trajectories are plotted in Fig. \ref{fig.compare.keyframe.2}. For comparison, we also plot the generalized motion of the previously learned weighted  cost  (\ref{exp.robotarm.cost}) and neural cost  (\ref{exp.robotarm.neuralcost}) in Fig. \ref{fig.compare.keyframe.3}. We have the following comments on the results.

First,  the spline function fits well to the keyframes (red dots in Fig. \ref{fig.compare.keyframe.1}). However, the generalization of the obtained spline model is poor: the generalized motion has a final distance of 33.670 to the goal. The poor performance is because the spline is only a \emph{local kinematic model}, and it cannot generalize motion that is far away from the keyframes. 

Second, given the same number of keyframes,   learning cost functions shows evident advantage in generalization. As in Fig. \ref{fig.compare.keyframe.3}, both the learned weighted cost function and neural cost function can successfully control the robot to reach the goal in new conditions. The reason why cost functions have superior performance is that a  cost function is a  compact representation of robot motion, and it represents a space of motion trajectories parameterized by different initial conditions and time horizons. Previous work  \cite{ravichandar2020recent} had the same conclusion.

\smallskip

\subsubsection{Comparison with Numerical Differentiation}
Recall that a key technique of the  Continuous PDP  is  Differential Pontryagin's Maximum Principle, which efficiently computes the \emph{analytic gradient} of the trajectory of a continuous-time optimal control system with respect to system parameters. An alternative  is numerical differentiation, that is,  one uses \emph{numerical differentiation} to obtain   $\frac{d L}{d \boldsymbol{\theta}}$. The experiments below compare those two options. Other experiment settings are the same as the previous ones.

\begin{figure}[h]
	\begin{subfigure}{.238\textwidth}
		\centering
		\includegraphics[width=\linewidth]{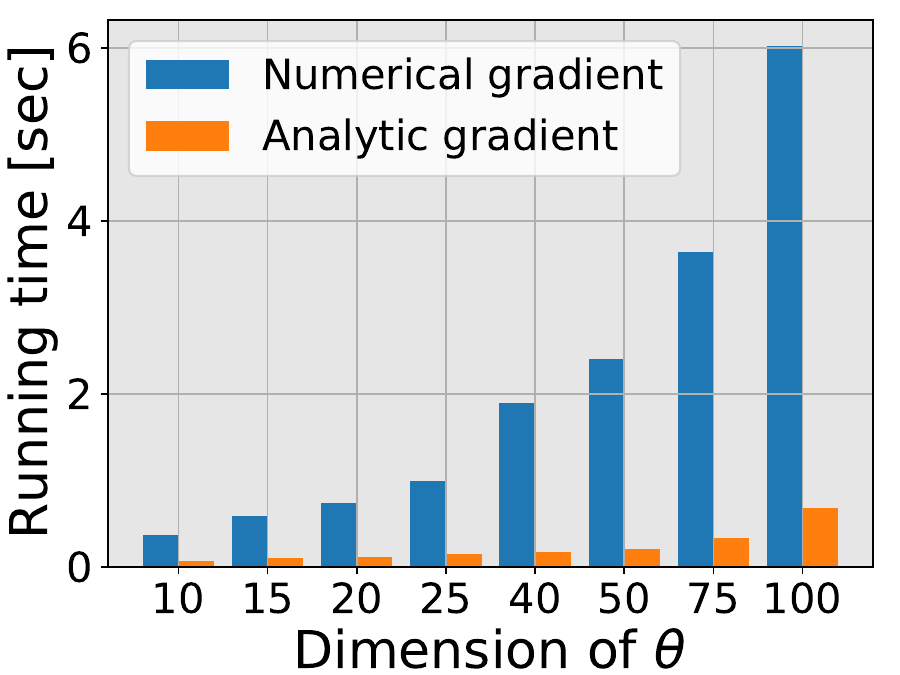}
		\caption{Varying parameter dimension}
		\label{fig.compare.gradient.1}
	\end{subfigure}
	\begin{subfigure}{.238\textwidth}
		\centering
		\includegraphics[width=\linewidth]{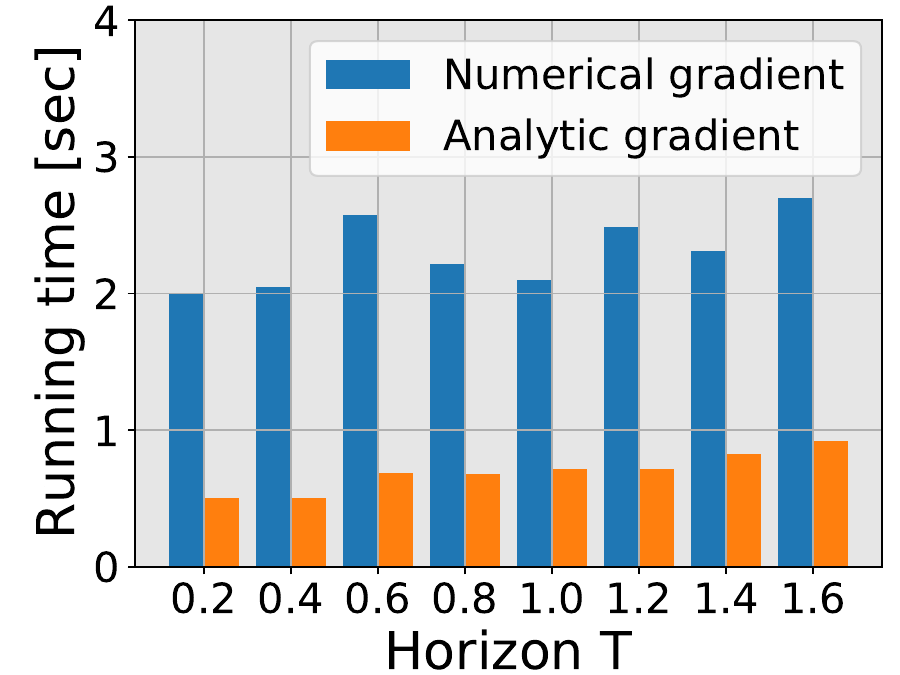}
		\caption{Varying system horizon}
		\label{fig.compare.gradient.2}
	\end{subfigure}
	\caption{Comparison of computation time between numerical differentiation and  Continuous PDP.
	} 
	\label{fig.compare.gradient}
\end{figure}

Consider  the neural cost function  in  (\ref{exp.robotarm.neuralcost}). We vary the size of the neural network, i.e., the dimension of  $\boldsymbol{p}$, and the system time horizon $T$. We compare the computation time needed to compute $\frac{d L}{d \boldsymbol{\theta}}$ by  Continuous PDP  and numerical differentiation.  The results are  in Fig. \ref{fig.compare.gradient}, based on which we have the following comments.

Fig. \ref{fig.compare.gradient.1} shows an exponential increase of computational time of  numerical differentiation when the number of  system parameters (dimension of $\boldsymbol{\theta}$) increases. This is because numerical differentiation requires evaluating the loss by perturbing the parameter vector in each dimension. Each perturbation and evaluation require solving an optimal control problem once, thus causing high-computational cost for high-dimensional  $\boldsymbol{\theta}$. 
In contrast, the  Continuous PDP solves analytical gradients by performing the Riccati-type iteration (Lemma \ref{theorem1}). Since there is no need to repetitively solve optimal control problems during the differentiation, the proposed method can handle the large-scale optimization problem, such as  $\boldsymbol{\theta}\in\mathbb{R}^{100}$  in Fig. \ref{fig.compare.gradient.1}.

Fig. \ref{fig.compare.gradient.2}  shows the comparison results given different system horizons. One observation is that the complexity of   Continuous PDP  is approximately linear to the system horizon $T$. This is because the numerical integration of the Riccati-type equations in Lemma \ref{theorem1} is linear to the horizon $T$.   

\smallskip

\subsubsection{Comparison with  \cite{hatz2012estimating}}
\label{comparetoplain} In this part, we compare the proposed method with  \cite{hatz2012estimating}. As discussed in the related work, \cite{hatz2012estimating} formulates a  problem similar to (\ref{problem}) which also minimizes a trajectory discrepancy loss (\ref{lossd}), but the authors solve it by replacing the inner optimal control problem with the Pontryagin's Maximum Principle conditions, thus turning a bi-level optimization into a  plain constrained optimization.  We compare their method with the  Continuous PDP in terms of convergence, sensitivity to different initialization, and generalization.

\begin{figure}[h]
	\begin{subfigure}{.238\textwidth}
		\centering
		\includegraphics[width=\linewidth]{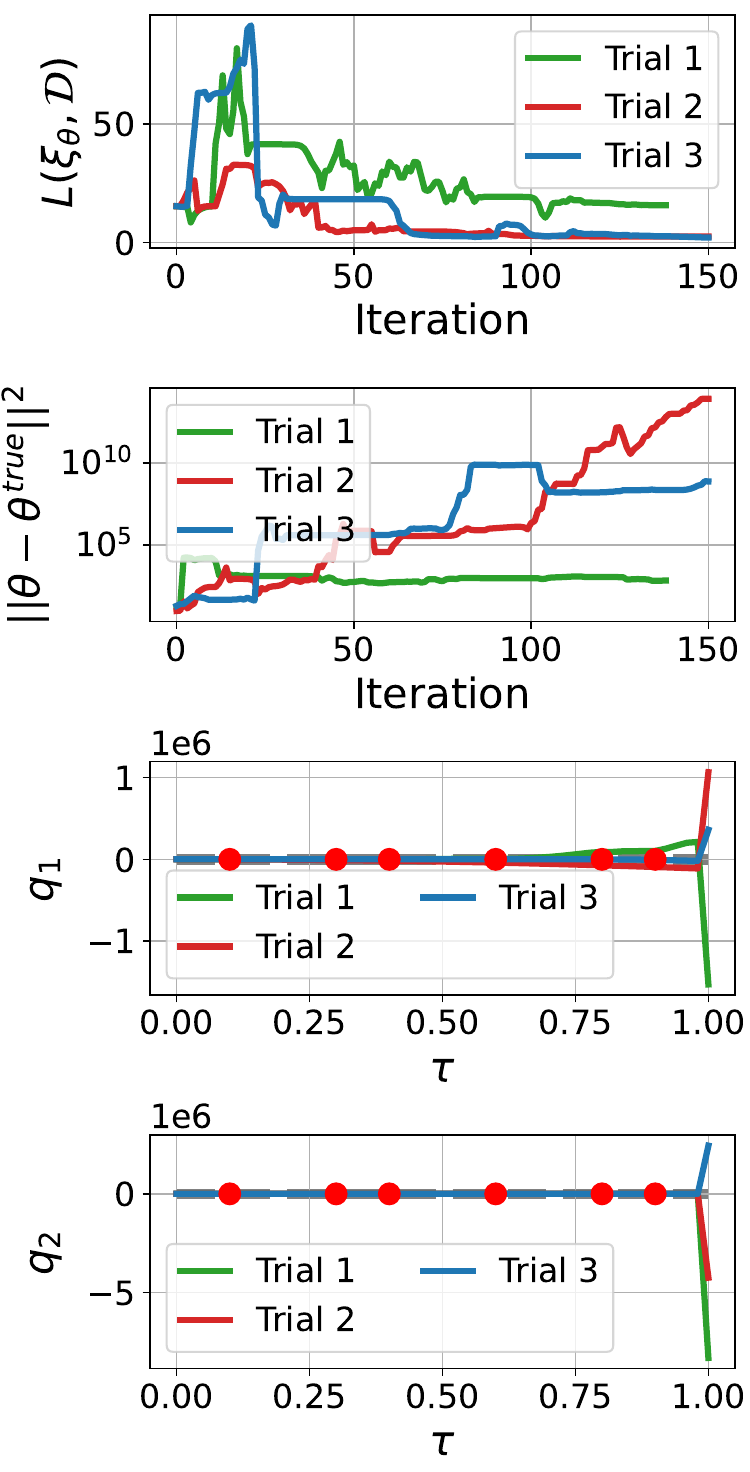}
		\caption{The method in \cite{hatz2012estimating}.}
		\label{fig.compare.plain.1}
	\end{subfigure}
	\begin{subfigure}{.238\textwidth}
		\centering
		\includegraphics[width=\linewidth]{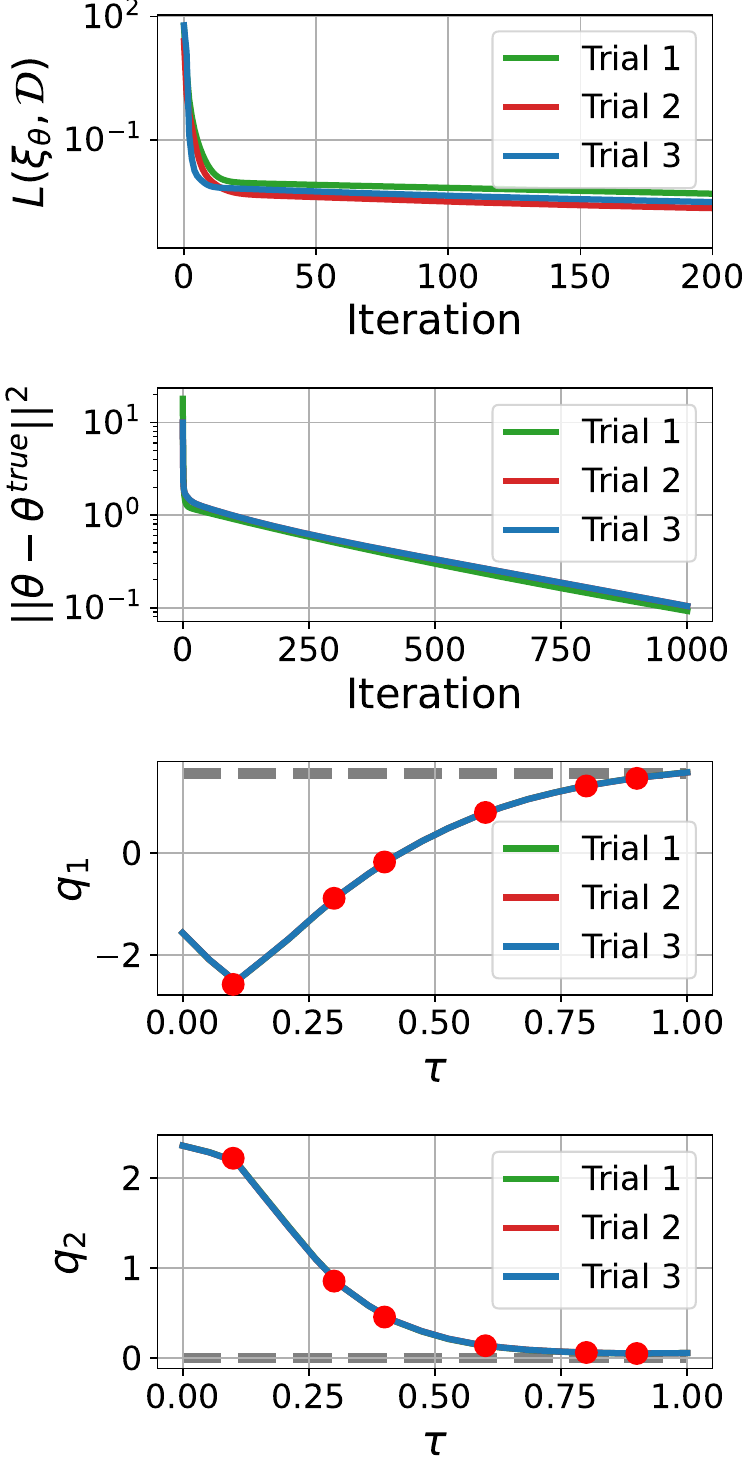}
		\caption{The proposed method.}
		\label{fig.compare.plain.2}
	\end{subfigure}
	\caption{Comparison between the method in \cite{hatz2012estimating} (a) and the proposed method (b). Each method has three trials using different initial  guesses $\boldsymbol{\theta}_0$, and at each trial, both methods start from the same  $\boldsymbol{\theta}_0$. Different trials are shown in different color. The first and second rows show the  loss and    parameter error versus iteration, respectively. The third and fourth rows show the reproduced trajectory of the learned $\boldsymbol{\theta}$. For   \cite{hatz2012estimating}, the converted plain optimization is solved by IPOPT \cite{wachter2006implementation}. Gray dashed lines mark the goal  of each joint $[q_1^{\text{g}}, q_2^{\text{g}}]\tran=[\pi/2, 0]\tran$. }
	\label{fig.compare.plain}
\end{figure}

Both methods use the same keyframes shown in the third and fourth rows in Fig. \ref{fig.compare.plain},  first-order polynomial time-warping function (\ref{beta}), and the cost function parameterization (\ref{exp.robotarm.cost}). Other experiment settings are the same as the previous experiments unless explicitly stated. Fig. \ref{fig.compare.plain} presents the results of  \cite{hatz2012estimating} (left) and  the proposed method (right). Here, each method has three trials from different initial guesses $\boldsymbol{\theta}_0$ (i.e., using different random seeds). Different trials are shown in different colors. The first and second rows plot the loss and parameter error versus iteration, respectively. The third and fourth rows show the reproduced trajectories with the learned $\boldsymbol{\theta}$. We have the following comments.

First, following \cite{hatz2012estimating}, the converted plain optimization  has 504 constraint equations and 509 decision variables. This is large-scale and non-convex optimization, and we used  IPOPT  \cite{wachter2006implementation} to solve it. But IPOPT is very likely to get stuck to local optima for this problem. This has been illustrated by  Fig. \ref{fig.compare.plain.1}: the loss has converged to a small value,  but the learned  $\boldsymbol{\theta}$ is  far away from   $\boldsymbol{\theta}^{\text{true}}$. Also,   in the third and fourth rows, although the produced trajectories are close to the keyframes (red dots), they are very different from the ground truth in Fig. \ref{fig.robotarm.sparseselection}.

The proneness of \cite{hatz2012estimating} to get stuck to bad solutions could be due to two main reasons. First,  since Pontryagin's Maximum Principle is just a \emph{necessary condition},  the solutions that satisfy this condition may include the saddle points, which  might not necessarily be the solution to the original optimal control problem. Thus, such a problem reformulation is not equivalent to the original bi-level problem in general. Second, the converted plain optimization can be large-scale and  highly nonlinear. If not properly initialized, it would easily get stuck into a local solution.

In contrast, the proposed method solves the problem by maintaining the bi-level structure. This bi-level treatment leads to more numerical tractability. The lower-level optimal control problem can be solved by many available trajectory optimization methods such as iLQR \cite{li2004iterative}, DDP \cite{jacobson1970differential}, and the upper level uses gradient-descent. Also, the bi-level treatment can lead to better performance in finding good (if not global) solutions. As empirically shown in Fig. \ref{fig.compare.plain.2}, with various random guesses  $\boldsymbol{\theta}_0$s, the proposed method  all converges to  the true  $\boldsymbol{\theta}^{\text{true}}$. 

Finally, we need to mention that in the Continuous PDP, Pontryagin's Maximum Principle is only used for \emph{differentiating the trajectory of the optimal control system}, not replacing the optimal control system. In other words, the trajectory has to be computed on the lower level \emph{before} its differentiation can be done. Therefore, the proposed method in this paper is fundamentally different from  \cite{hatz2012estimating}.

\section{Learning  from Keyframes for   planning in unknown environments}\label{section_application}
This section presents an application scenario of the proposed method:   a robot learns a motion planner from demonstrated keyframes to navigate through an unknown environment. A  user provides a few keyframes in the vicinity of obstacles in an environment, and a robot learns a   cost function from those keyframes such that its produced motion can avoid the obstacles. Experiments in this section are based on a 6-DoF  quadrotor. The code can be accessed at \url{https://github.com/wanxinjin/Learning-from-Sparse-Demonstrations}. A real-world demonstration is given at \url{https://youtu.be/BYAsqMxW5Z4}.

\subsection{6-DoF Quadrotor  Setup} \label{app.quadrotor}

The equation of motion of a quadrotor flying in {$SE(3)$} (full position and attitude) space is given by 
\begin{subequations}\label{uav_dynamics}
	\begin{align}
	\dot{\boldsymbol{r }}_{I}&{={\boldsymbol{v}}_{I}},\\
	m\dot{\boldsymbol{v}}_{I}&=m\boldsymbol{g}_{I}+\boldsymbol{f}_{I},\\
	\dot{\boldsymbol{q}}_{B/I}&=\frac{1}{2}\Omega(\boldsymbol{\omega}_{B}){\boldsymbol{q}}_{B/I},\label{quaterdiff}\\
	J_{B}\dot{\boldsymbol{\omega}}_{B}&=\boldsymbol{\tau}_{B}-\boldsymbol{\omega}_{{B}}\times J_{B}\boldsymbol{\omega}_{B}.
	\end{align}
\end{subequations}
Here, the subscripts $_{B}$ and $_{I}$ denote a quantity  expressed in the  body  and world coordinate frames, respectively;  $m$ is the quadrotor mass; $\boldsymbol{r}_I=[r_x,r_y,r_z]\tran\in\mathbb{R}^{3}$ and $\boldsymbol{v}_I\in\mathbb{R}^{3}$ are the quadrotor's position and velocity, respectively; $J_{B}\in\mathbb{R}^{3\times3}$ is the moment of inertia; $\boldsymbol{\omega}_{B}\in\mathbb{R}^{3}$ is the angular velocity; $\boldsymbol{q}_{B/I}\in\mathbb{R}^{4}$ is the unit quaternion \cite{kuipers1999quaternions} describing the attitude of the quadrotor  with respect to the world frame; (\ref{quaterdiff}) is the quaternion calculus, and $\Omega(\boldsymbol{\omega}_B)$  is the matrix  of $\boldsymbol{\omega}_{B}$  for quaternion multiplication \cite{kuipers1999quaternions};
$\boldsymbol{\tau}_{B}\in \mathbb{R}^{3}$ is the torque vector applied to the quadrotor; and $\boldsymbol{f}_{I}\in \mathbb{R}^{3}$ is the total force vector. The net force magnitude $ \norm{\boldsymbol{f}_{I}}={ {f} }\in\mathbb{R}$ (along the z-axis of the body frame) and torque $\boldsymbol{\tau}_{B}=[\tau_x, \tau_y, \tau_z]$ are generated by thrust $[T_1, T_2, T_3, T_4]$ of  four    propellers via

\begin{small}
	\begin{equation}
	\begin{bmatrix}
	{ {f}}\\
	\tau_x\\
	\tau_y\\
	\tau_z
	\end{bmatrix}=
	\begin{bmatrix}
	1&1 &1 &1\\
	0&-l_\text{w}/2& 0 & l_\text{w}/2 \\
	-l_\text{w}/2& 0 & l_\text{w}/2 & 0 \\
	\kappa&-\kappa&\kappa&-\kappa
	\end{bmatrix}
	\begin{bmatrix}
	T_1\\
	T_2\\
	T_3\\
	T_4
	\end{bmatrix},
	\end{equation}
\end{small}%
with $l_\text{w}$   the  wing length of the quadrotor  and $\kappa$ a fixed constant. In our experiment, the gravity constant is $10 \text{m/s}^2$ and  all other dynamics parameters  are units. The state vector is  
$
\boldsymbol{x}=[\boldsymbol{r}_I\tran , \boldsymbol{v}_I\tran  ,\boldsymbol{q}_{B/I}\tran, \boldsymbol{\omega}_B\tran]\tran \in\mathbb{R}^{13}
$
and  the control vector is  
$
\boldsymbol{u}=[T_1, T_2, T_3, T_4]\tran \in\mathbb{R}^{4}.
$

To achieve  ${SE(3)}$ maneuvering, we need to carefully design the attitude error.  Following \cite{lee2010geometric}, we define the attitude error between the  current attitude $\boldsymbol{q}$ and   goal  $\boldsymbol{q}^{\text{g}}$ as
\begin{equation}
e(\boldsymbol{q},\boldsymbol{q}_{\text{g}})=\frac{1}{2} \text{trace}(I-R\tran(\boldsymbol{q}^{\text{g}})R(\boldsymbol{q})),
\end{equation}
where $R(\boldsymbol{q})\in\mathbb{R}^{3\times3}$ is the   rotation matrix corresponding to  quaternion $\boldsymbol{q}$ (see \cite{kuipers1999quaternions} for more details).
For the   cost function formulation (\ref{costfun}), we  use a generic     polynomial function:
\begin{subequations}\label{uav_objective}
	\begin{align}
	c({\boldsymbol{x},\boldsymbol{u}},{\boldsymbol{p}}){= }&p_1r_{x}^2+ p_2r_{y}^2+p_3r_{z}^2+p_4r_x+p_5r_y+p_6r_z\nonumber\\
	&+p_7r_{x}r_y+p_8r_xr_z+p_9r_yr_z +0.1\norm{\boldsymbol{u}}^2,\label{uav_objective.1}\\
	h({\boldsymbol{x}}){=}&10\norm{\boldsymbol{r}_I{-}\boldsymbol{r}_I^{\text{g}}}^2+5\norm{\boldsymbol{v}_I}^2+\nonumber\\
	&100e(\boldsymbol{q}_{B/I},\boldsymbol{q}_{B/I}^{\text{g}})+5\norm{\boldsymbol{\omega}_B}^2, \label{uav_objective.2}
	\end{align}
\end{subequations}
where $\boldsymbol{r}_{I}=[r_x,r_y,r_z]\tran$ is the quadrotor's position, and we  have fixed  the final cost   $h(\boldsymbol{x})$  since  the quadrotor is always expected   to  land near  a goal position  $\boldsymbol{r}_{I}^\text{g}$ with   goal  attitude   $\boldsymbol{q}_{B/I}^{\text{g}}$; and the goal velocities here are zeros.  The cost function parameter  $\boldsymbol{p}=[p_1, p_2, p_3, p_4, p_5, p_6, p_7, p_8, p_9]\tran\in\mathbb{R}^9$  will   determine  how the quadrotor reaches the goal (i.e., the specific flying trajectory of the quadrotor).

\begin{figure}[h]
	\centering	\includegraphics[width=0.8\columnwidth]{uav_setup.pdf}
	\caption{Quadrotor flying in an environment with obstacles. The goal is to let the quadrotor to fly from the left,  go through the two gates (from left to right), and finally land near the goal position   on the right with  goal attitude. The plotted   trajectory is a planned motion   with a random   cost function, which fails to achieve the goal.}
	\label{uav_corrections}
\end{figure}

As shown in Fig. \ref{uav_corrections}, we aim  the quadrotor to fly from the left initial position $\boldsymbol{r}_I(0)$ with   $\boldsymbol{v}_I(0)$, $ \boldsymbol{q}_{B/I}(0)$, and $\boldsymbol{\omega}_B(0)$, sequentially pass through two gates (from left to right gates), and finally land near  the goal position $\boldsymbol{r}_I^\text{g}$ with   goal  attitude $\boldsymbol{q}_{B/I}^{\text{g}}$ on the right. With a   random  cost function,  the quadrotor  trajectory (blue line) does not meet the task requirement.  

\subsection{Learning from  Keyframes}\label{section.uav.learning}

\begin{table}[h]
	\centering
	\caption	{Keyframes $\mathcal{D}$ for the quadrotor.}
	\begin{tabular}{ccc}
		\toprule
		Keyframe \# &Time stamp $\tau_i$   & Keyframe $\boldsymbol{y}^*(\tau_{i})$ \\
		\midrule
		\#1	&$\tau_1=0.1$s &   \begin{tabular}{@{}c@{}}$\boldsymbol{r}_I({\tau_1})=[-4, -6, 3]$
		\end{tabular}  \\
		
		\#2	&$\tau_2=0.2$s &   \begin{tabular}{@{}c@{}}$\boldsymbol{r}_I({\tau_2})=[1, -6, 3]$	
		\end{tabular}  \\

		\#3	&$\tau_3=0.4$s &   \begin{tabular}{@{}c@{}}$\boldsymbol{r}_I({\tau_3})=[1, -1, 4]$
		\end{tabular}  \\

		\#4	&$\tau_4=0.6$s &   \begin{tabular}{@{}c@{}}$\boldsymbol{r}_I({\tau_4})=[-1, 1, 5]$
		\end{tabular}  \\

		\#5	&$\tau_5=0.8$s &   \begin{tabular}{@{}c@{}}$\boldsymbol{r}_I({\tau_5})=[2, 3, 4]$
		\end{tabular}  \\

		&horizon $T=1$s  &\\
		\bottomrule
	\end{tabular}
	\label{table_uav_corrections}
\end{table}

We arbitrarily choose five keyframes near the two gates,  listed in  Table \ref{table_uav_corrections}. Here,  `arbitrarily' means that we do not know whether these keyframes are realizable by an exact cost function. Without much deliberation, we assign a time stamp to each keyframe, such that they are (almost) evenly spaced in the time horizon $[0, T]$ (later we will also test the method given the random assignment of the time stamps). We also do not know whether $\tau_i$ and $T$  are achievable for the quadrotor dynamics. The keyframes  here contain only position information,  i.e.,  
\begin{equation}
\boldsymbol{r}_I=\boldsymbol{y}=\boldsymbol{g}(\boldsymbol{x},\boldsymbol{u})
\end{equation}
The  time-warping function is  the first-order polynomial function (\ref{beta}), and loss  $L(\boldsymbol{\xi}_{\boldsymbol{{\theta}}},\mathcal{D})$ is   in (\ref{lossd}).  The  learning rate  is  $\eta{=}10^{-2}$. The quadrotor's initial state  is $\boldsymbol{r}_I(0)=[{-}8, {-}8, 5]\tran$, $\boldsymbol{q}_{B/I}(0)=[1,0,0,0]\tran$, $\boldsymbol{v}_I(0)=[15, 5, {-}10]\tran$, and $\boldsymbol{\omega}_B(0)=\boldsymbol{0}$. The  goal position is  $\boldsymbol{r}_I^\text{g}=[8, 8, 0]\tran$ and  the goal attitude  $\boldsymbol{q}_{B/I}^{\text{g}}=[1,0,0,0]\tran$ (recall the goal velocities here are zeros).

\begin{figure}[h]
	\centering
	\begin{subfigure}{.158\textwidth}
		\centering
		\includegraphics[width=\linewidth]{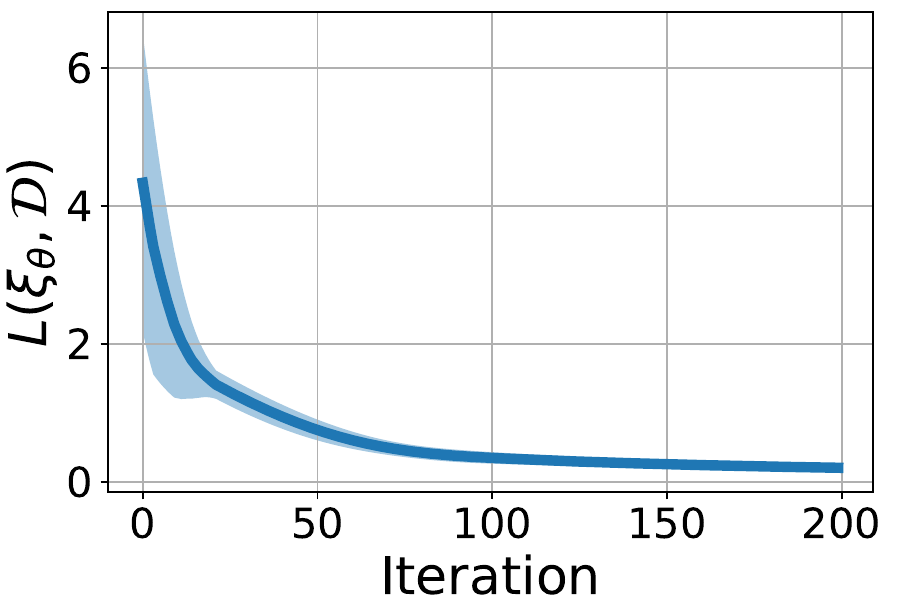}
		\caption{Loss for Fig. \ref{fig.uav.reproduce.1}}
		\label{fig.uav.loss.1}
	\end{subfigure}
	\begin{subfigure}{.158\textwidth}
		\centering
		\includegraphics[width=\linewidth]{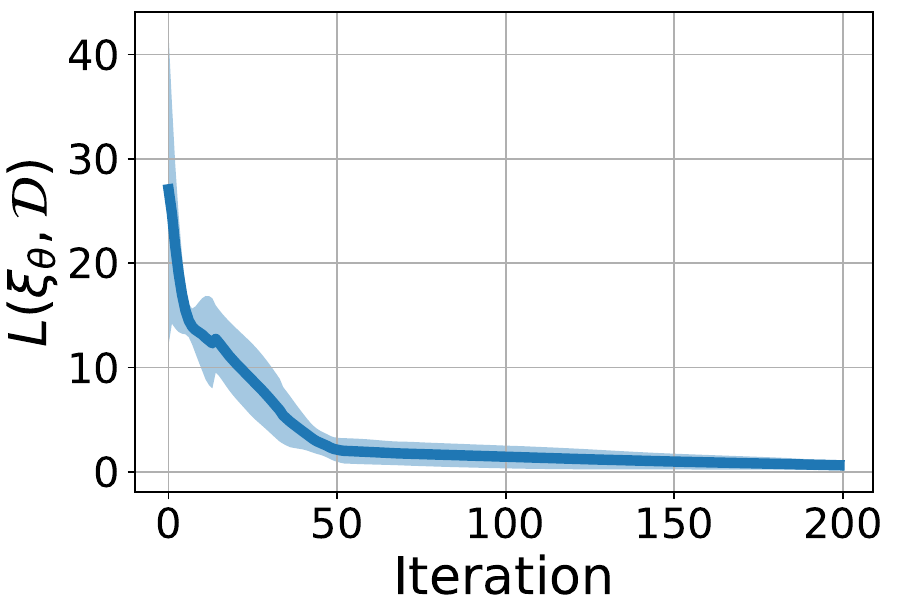}
		\caption{Loss for Fig. \ref{fig.uav.reproduce.2}}
		\label{fig.uav.loss.2}
	\end{subfigure}
	\begin{subfigure}{.158\textwidth}
		\centering
		\includegraphics[width=\linewidth]{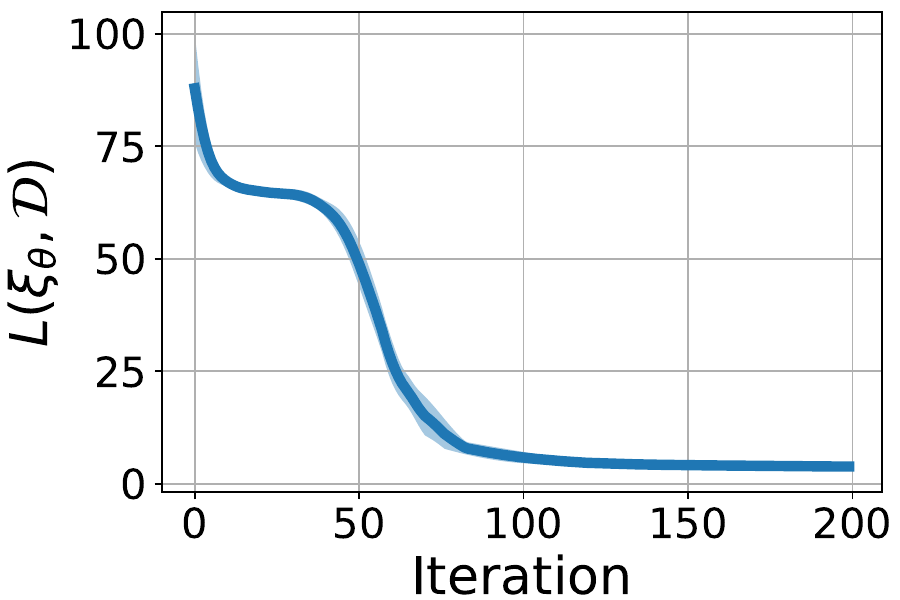}
		\caption{Loss for Fig. \ref{fig.uav.reproduce.3}}
		\label{fig.uav.loss.3}
	\end{subfigure}
	\begin{subfigure}{.158\textwidth}
		\centering
		\includegraphics[width=\linewidth]{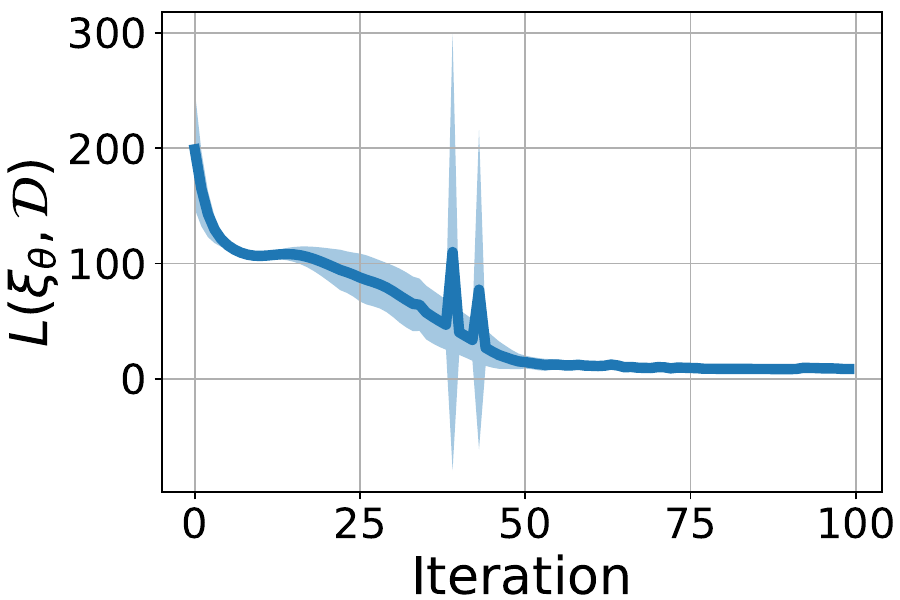}
		\caption{Loss for Fig. \ref{fig.uav.reproduce.4}}
		\label{fig.uav.loss.4}
	\end{subfigure}
	\begin{subfigure}{.158\textwidth}
		\centering
		\includegraphics[width=\linewidth]{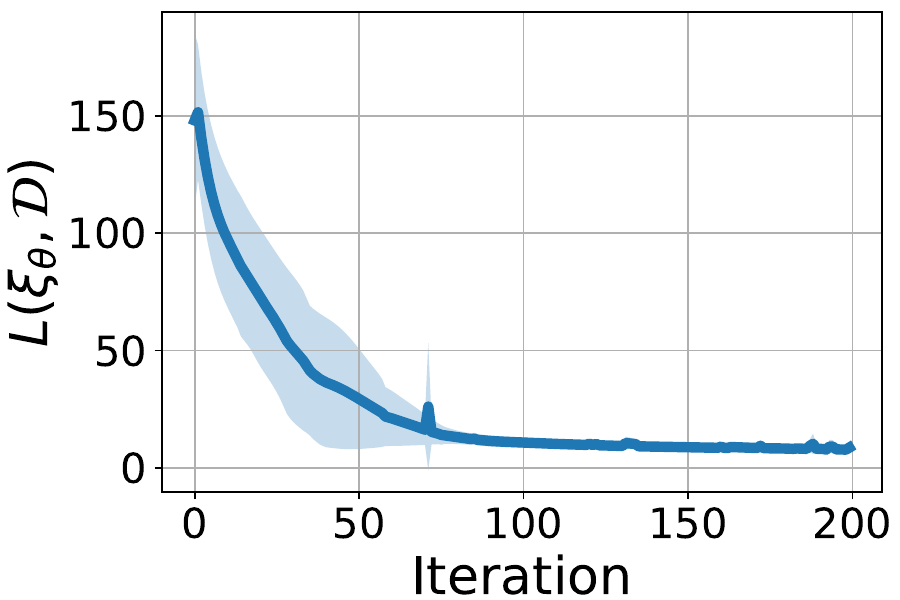}
		\caption{Loss for Fig. \ref{fig.uav.reproduce.5}}
		\label{fig.uav.loss.5}
	\end{subfigure}
	\begin{subfigure}{.158\textwidth}
		\centering
		\includegraphics[width=\linewidth]{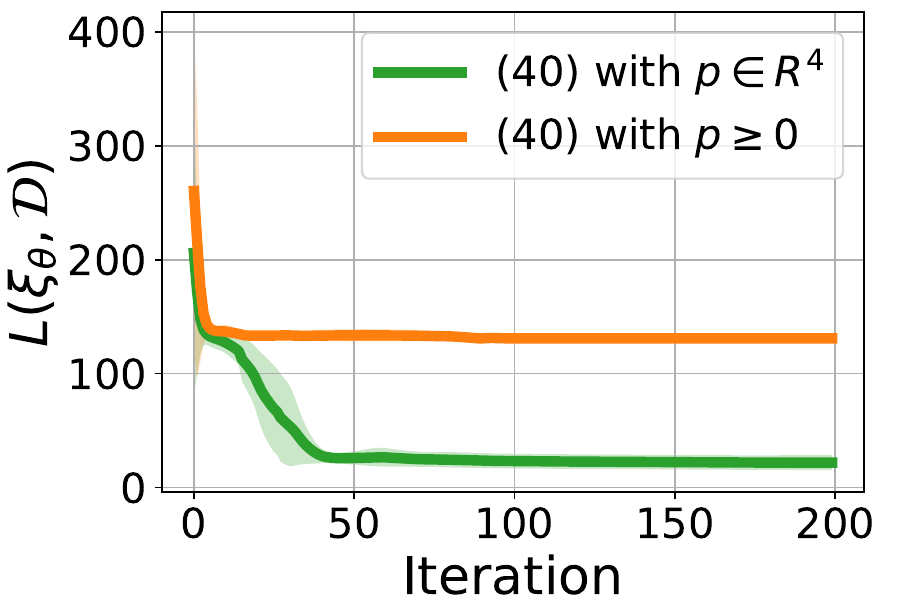}
		\caption{Loss for Fig. \ref{fig.uav.reproduce.6}}
		\label{fig.uav.loss.6}
	\end{subfigure}
	
	\caption{
		Loss  versus  iteration, corresponding to different  cases in Fig. \ref{fig.uav.reproduce}. In each case, the solid line and shaded area denote the mean and standard derivation over  10 trial of the experiment, respectively. The final loss (mean+std)   for each case is: $0.203\pm0.035$ in  (a), $0.625\pm0.470$ in (b), $ 3.819\pm0.805$ in (c), $8.548\pm0.880$ in (d), $8.647\pm2.022$ in (e), $21.777\pm6.608$ for $\boldsymbol{p}\in\mathbb{R}^4$ and $130.902\pm 0.006$ for  $\boldsymbol{p}\geq \boldsymbol{0}$  in (f).
	} 
	\label{fig.uav.loss}
\end{figure}

\begin{figure*}[h]
	\centering
	\begin{subfigure}{.32\textwidth}
		\centering
		\includegraphics[width=\linewidth]{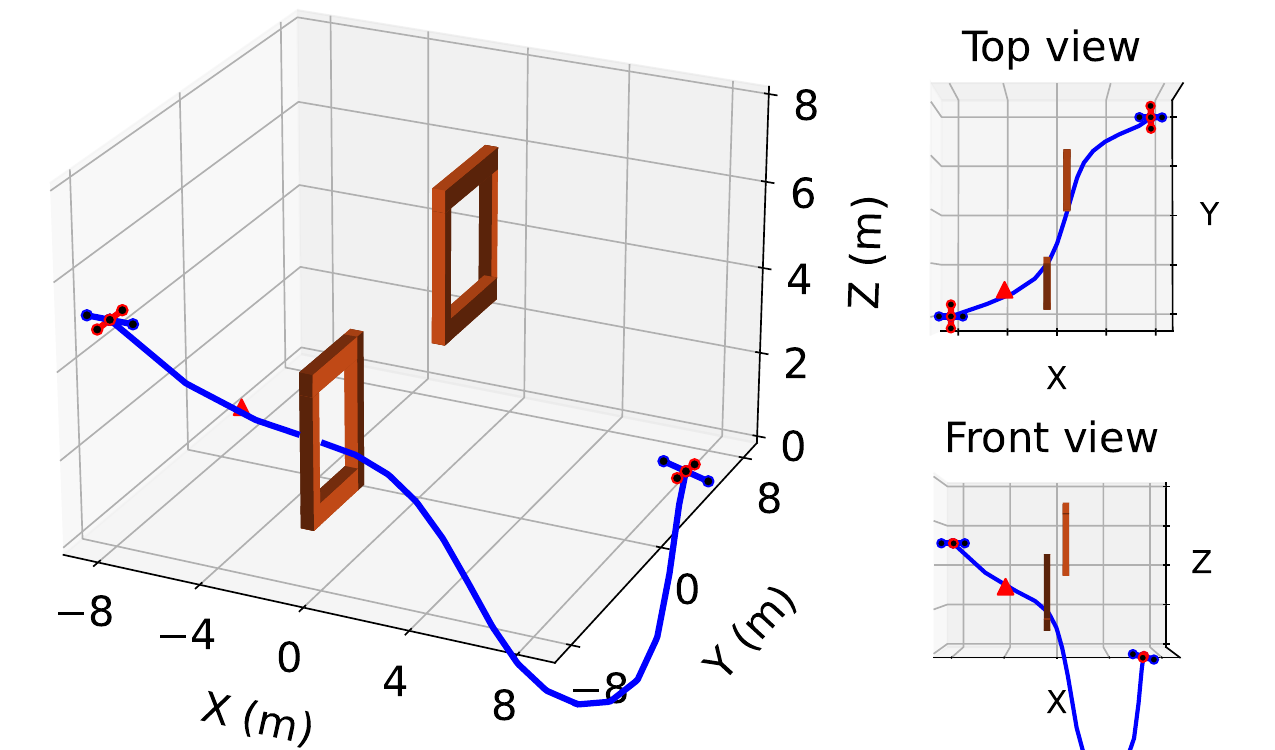}
		\caption{Learning from  keyframe \#1.}
		\label{fig.uav.reproduce.1}
	\end{subfigure}
	\begin{subfigure}{.32\textwidth}
		\centering
		\includegraphics[width=\linewidth]{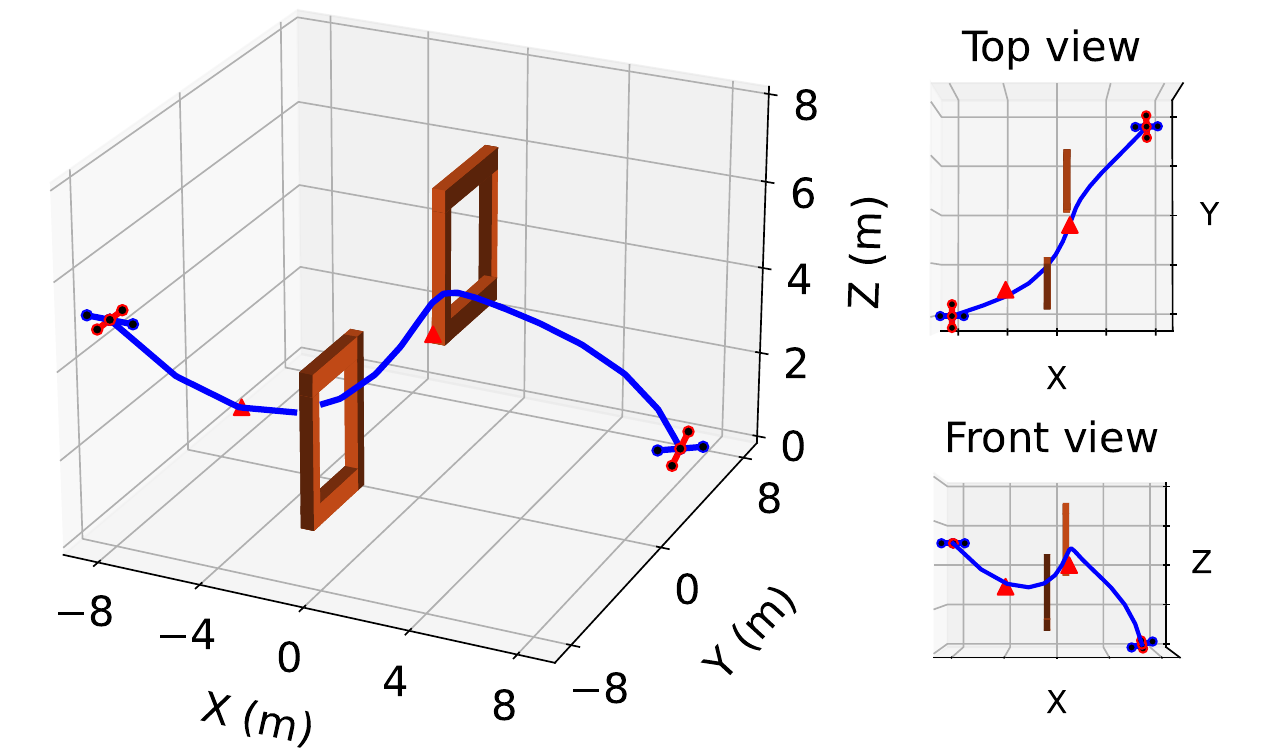}
		\caption{Learning from  keyframes \#1 and \#3.}
		\label{fig.uav.reproduce.2}
	\end{subfigure}
	\begin{subfigure}{.32\textwidth}
		\centering
		\includegraphics[width=\linewidth]{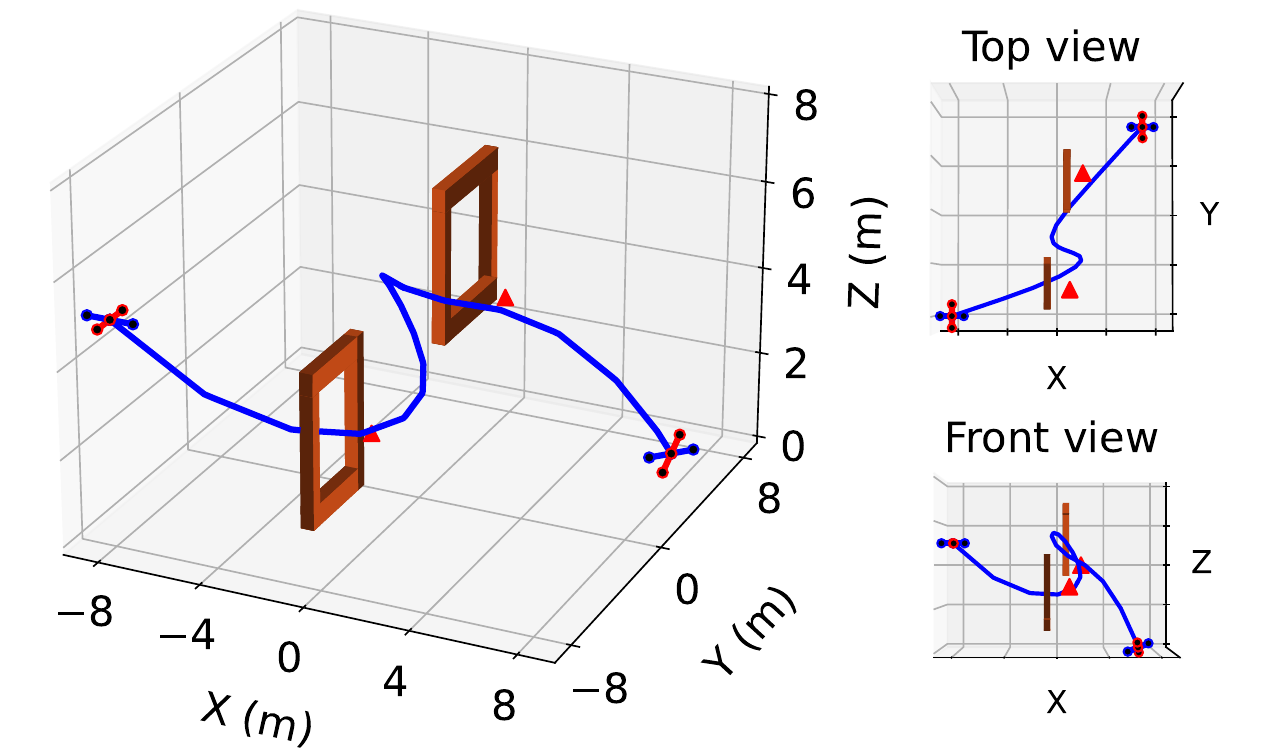}
		\caption{Learning from keyframes \#2 and \#5.}
		\label{fig.uav.reproduce.3}
	\end{subfigure}
	\begin{subfigure}{.32\textwidth}
		\centering
		\includegraphics[width=\linewidth]{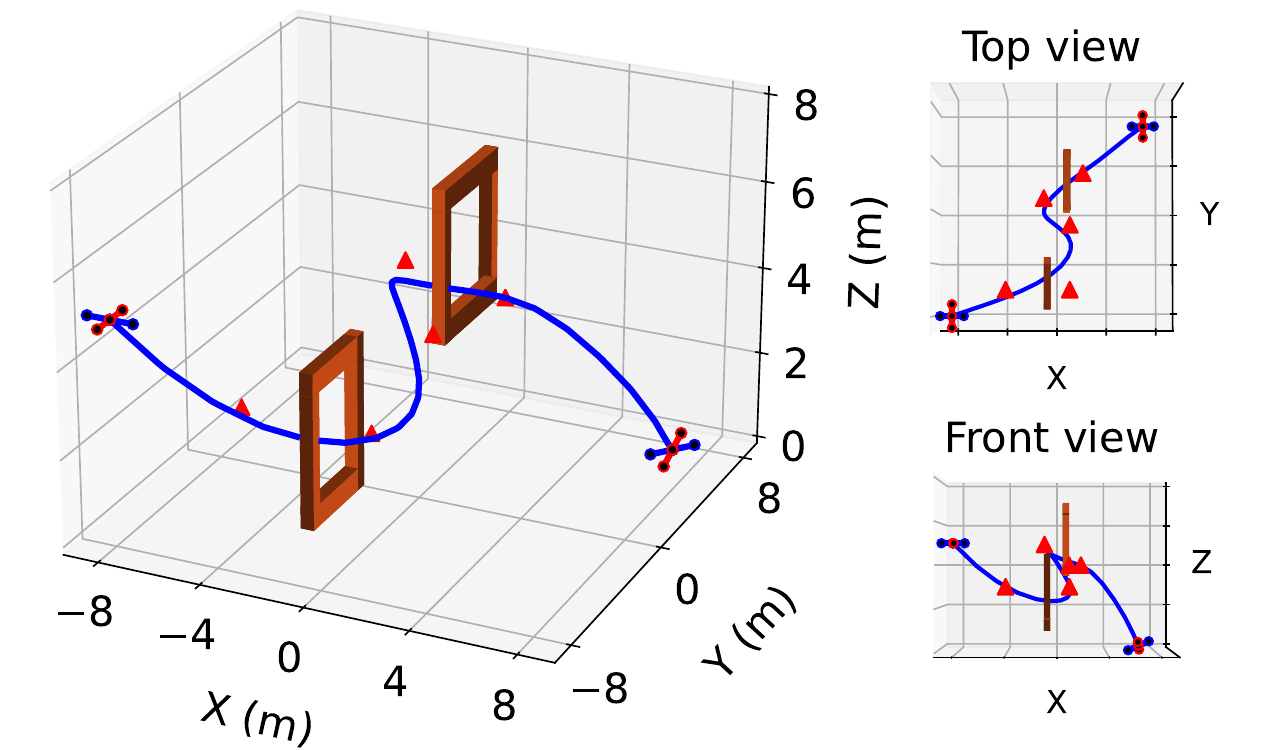}
		\caption{Learning from  keyframes \#1-\#5.}
		\label{fig.uav.reproduce.4}
	\end{subfigure}
	\begin{subfigure}{.32\textwidth}
		\centering
		\includegraphics[width=\linewidth]{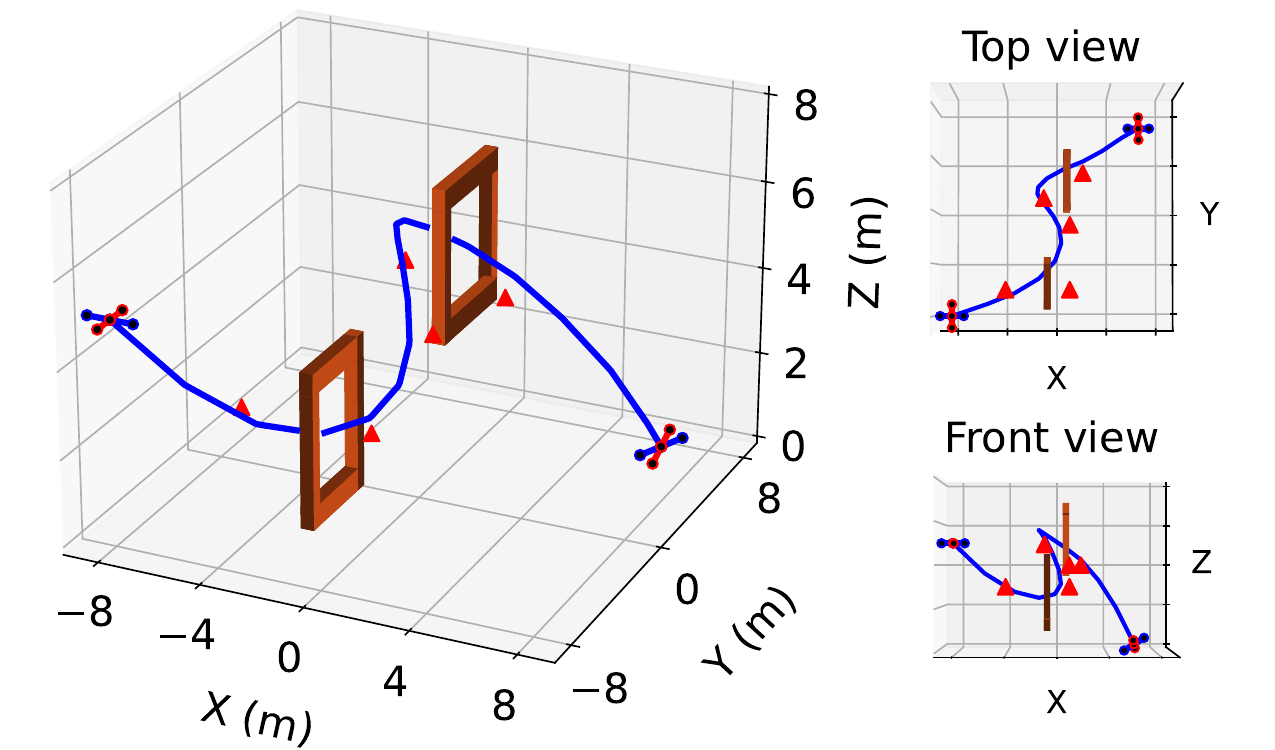}
		\caption{Keyframes \#1-\#5 with random stamps.}
		\label{fig.uav.reproduce.5}
	\end{subfigure}
	\begin{subfigure}{.32\textwidth}
		\centering
		\includegraphics[width=\linewidth]{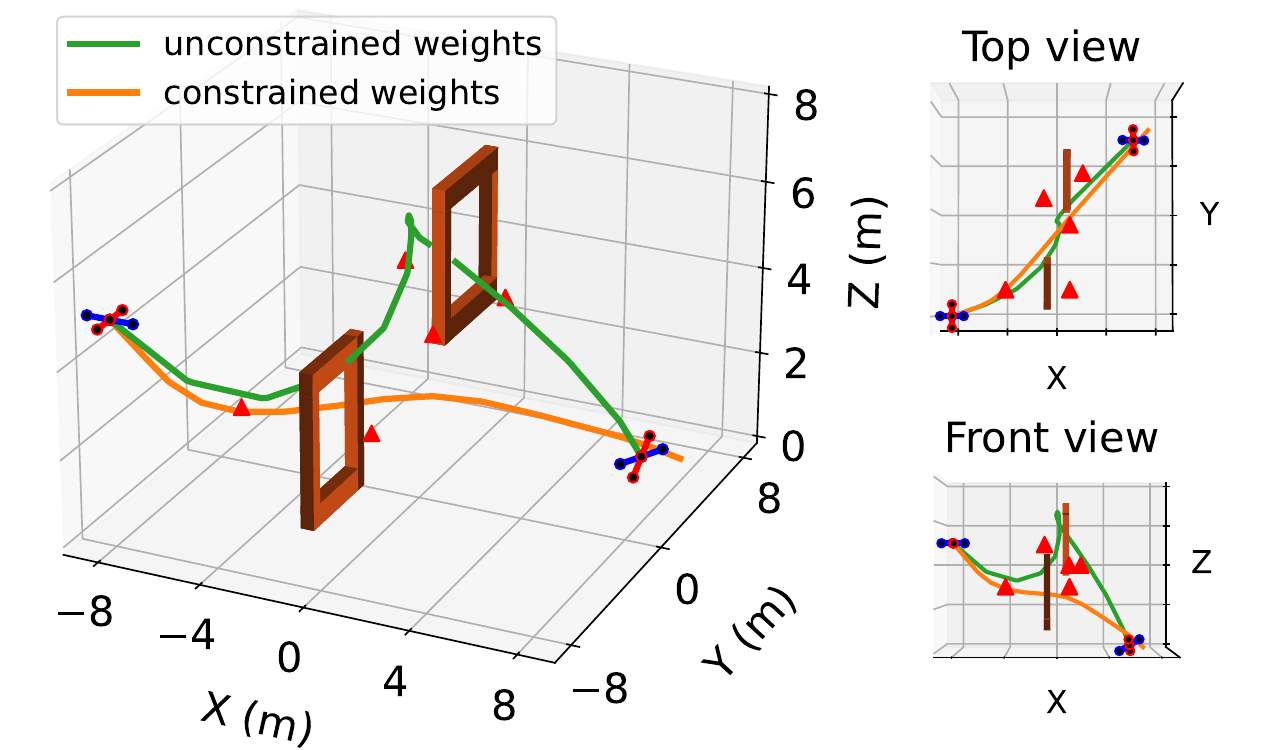}
		\caption{Learning (\ref{uav_objective.replace}) from keyframes \#1-\#5.}
		\label{fig.uav.reproduce.6}
	\end{subfigure}
	\caption{Learning from  keyframes given  in Table \ref{table_uav_corrections}. (\ref{fig.uav.reproduce.1})-(\ref{fig.uav.reproduce.4}) show the reproduced  trajectories by the  cost functions learned from different number of keyframes.  (\ref{fig.uav.reproduce.5}) shows the case where  we  randomize the time stamp  $\tau_i$ of each keyframe in Table \ref{table_uav_corrections}. (\ref{fig.uav.reproduce.6}) shows the reproduced motion of the learned    distance-to-obstacle cost function  (\ref{uav_objective.replace}); here the green line corresponds to the unconstrained weights subcase, i.e.,  $\boldsymbol{p}\in\mathbb{R}^4$, and the orange  to the   constrained weights subcase, i.e.,  $\boldsymbol{p}\geq\boldsymbol{0}$. Corresponding to each of the above-mentioned figures, the loss versus iteration is given in Fig. \ref{fig.uav.loss}.
	} 
	\label{fig.uav.reproduce}
\end{figure*}

 \subsubsection{Varying Number of Keyframes}
 
 Fig. \ref{fig.uav.loss.1}-\ref{fig.uav.loss.4}  and Fig.  \ref{fig.uav.reproduce.1}-\ref{fig.uav.reproduce.4} show different  cases where we  learn a cost function from  different numbers of keyframes. In  each case, we have run each experiment  case for 10 trials, with each trial using different random seeds for the initial $\boldsymbol{\theta}_0$. Fig. \ref{fig.uav.loss.1}-\ref{fig.uav.loss.4} plot the loss $L(\boldsymbol{\xi}_{\boldsymbol{{\theta}}},\mathcal{D})$ versus  iteration, and Fig. \ref{fig.uav.reproduce.1}-\ref{fig.uav.reproduce.4} show the reproduced trajectory using the learned  cost  and time-warping functions. We have the following comments on the results.

Fig. \ref{fig.uav.reproduce.1}-\ref{fig.uav.reproduce.4} show that given keyframes in different locations, the proposed method always finds a   cost function and a time-warping function such that the quadrotor's reproduced motion can get close to the keyframes.  Fig. \ref{fig.uav.reproduce.1}-\ref{fig.uav.reproduce.4} also show that by increasing the number of keyframes and putting the keyframes around the gates, the quadrotor can successfully learn a  cost function to fly through the two gates.  Since the  keyframes are arbitrarily placed and   exact cost and time-warping functions (in the  parameterization) may not exist, the final losses are not zeros as in Fig. \ref{fig.uav.loss.1}-\ref{fig.uav.loss.4}.   Recall that we only make the path cost tunable, while the final cost given and fixed. Different placement of keyframes leads to different learned path costs and thus different motion trajectories. This cost formulation can be useful for learning how to move instead of where to move. 

\smallskip

\subsubsection{Random Time Stamps}
In Fig. \ref{fig.uav.reproduce.5} and Fig. \ref{fig.uav.loss.5}, we  randomize the time stamp $\tau_i$ of each keyframe in Table \ref{table_uav_corrections} (drawn from a uniform distribution), and   the  cost function is  learned from the randomly-timed keyframes. The other  settings follow the previous experiment. Fig. \ref{fig.uav.loss.5} plots the loss versus  iteration, and Fig. \ref{fig.uav.reproduce.5} show the reproduced trajectory from the learned  cost  and time-warping functions.

Comparing Fig. \ref{fig.uav.reproduce.4} and Fig. \ref{fig.uav.reproduce.5}  with Fig.  \ref{fig.uav.loss.4} and Fig. \ref{fig.uav.loss.5}, respectively, we can see that  the choice of time stamps of keyframes does not affect  too much the learning: the final loss (mean+std) is {$8.548\pm 0.880$}  for Fig. \ref{fig.uav.reproduce.4} and {$8.647\pm 2.022$} for  Fig. \ref{fig.uav.reproduce.5}.  This result is understandable because whatever the keyframe time is, the proposed method always learns a  time-warping function, which maps demonstration time to the robot dynamics time;  thus performance of robot execution will not change significantly. The results show the importance of using a  time-warping function in general LfD problems. The ability to handle the time misalignment is one of the key features of the proposed method.

\smallskip

\subsubsection{Distance-to-Obstacle Cost Parameterization}\label{section.distace-feature-formulation}
In Fig. \ref{fig.uav.reproduce.6} and Fig. \ref{fig.uav.loss.6}, we replace the polynomial  cost  function (\ref{uav_objective.1})  with the following  distance-to-obstacle cost function:
\begin{equation}
\begin{aligned}
c({\boldsymbol{x},\boldsymbol{u}},{\boldsymbol{p}}){= }-\sum_{i=1}^{4}p_i\norm{\boldsymbol{r}_I-\boldsymbol{o}_i}^2 +0.1\norm{\boldsymbol{u}}^2,\label{uav_objective.replace}
\end{aligned}
\end{equation}
where the \emph{given} $\boldsymbol{o}_i$ is the obstacle $i$'s position, which  is the position of the left and right pillars of the two gates,  shown in Fig. \ref{uav_corrections}; and $\boldsymbol{p}\in\mathbb{R}^4$ are the weights for each to-obstacle distance $\norm{\boldsymbol{r}_{I}-\boldsymbol{o}_i}^2$. We learn ({\ref{uav_objective.replace}}) from the five keyframes  in Table \ref{table_uav_corrections}. 
Other experiment settings follow the previous session. We further divide the experiment into two subcases: In the first subcase (green line), we treat the weights  $\boldsymbol{p}$ as unconstrained variables (i.e., it could be $\boldsymbol{p}\leq\boldsymbol{0}$, the obstacles could have an  `attracting' effect on quadrotor motion); and in the second subcase (orange line), we force $\boldsymbol{p}\geq\boldsymbol{0}$. The loss for those two subcases are plotted in Fig. \ref{fig.uav.loss.6}, and the reproduced trajectories of the learned cost functions are  in Fig. \ref{fig.uav.reproduce.6}, where the green line corresponds to the  unconstrained weights subcase while the orange  to  the  constrained weights subcase.

In the  unconstrained weights   subcase in Fig. \ref{fig.uav.reproduce.6}, one  observation is that the  motion (green line) is  similar to the motion  in Fig. \ref{fig.uav.reproduce.4}  and Fig. \ref{fig.uav.reproduce.5}. In fact,  the learned weights are $\boldsymbol{p}{=}[-0.806,  -1.406,  -2.578, -2.246]\tran{\leq} \boldsymbol{0}$. This indicates that \emph{each  obstacle  has an attracting effect on  the quadrotor's motion}. Considering the distance-to-obstacle cost  in (\ref{uav_objective.replace}) is  a second-order polynomial function in $\boldsymbol{r}_I$, similar to  (\ref{uav_objective.1}), the  results  in Fig. \ref{fig.uav.reproduce.6} could    explain  those in Fig.~\ref{fig.uav.reproduce.4}  and Fig.~\ref{fig.uav.reproduce.5}. Specifically, one can intuitively think of  learning  a general polynomial cost function    (\ref{uav_objective.1}) as  a process of  finding some   `virtual attracting points' in the unknown environment, and both their locations and  attracting weights will be encoded in the learned polynomial coefficients. 

We also note that the reproduced motion (green line) in the unconstrained weights subcase in Fig. \ref{fig.uav.reproduce.6}  has a larger distance to the keyframes than the motion in \ref{fig.uav.reproduce.4}.  
This has been quantitatively shown by their final loss values in Fig. \ref{fig.uav.loss}: the former has a final  loss   {$21.777\pm6.608$} while the latter   { $8.548\pm0.880$}. This is  because  the  formulation   (\ref{uav_objective.1}) has $\boldsymbol{p}\in\mathbb{R}^9$, while     (\ref{uav_objective.replace})  only has  $\boldsymbol{p}\in\mathbb{R}^4$. Learning  (\ref{uav_objective.1}) allows us to optimize  both    weights and locations of the `virtual attracting points', while learning (\ref{uav_objective.replace})  allows us to only optimize the weights as the location of the obstacle $\boldsymbol{o}_i$ are given.

In the non-negative weights subcase (the orange line) in Fig. \ref{fig.uav.reproduce.6}, since we always force   $\boldsymbol{p}\geq \boldsymbol{0}$, the obstacles only have a `repelling' effect on  the quadrotor motion, and thus the final quadrotor motion avoids all obstacles, as in Fig. \ref{fig.uav.reproduce.6}. Also, Fig. \ref{fig.uav.loss.6} shows that the final loss of this subcase  is {$130.902\pm 0.006$}  is higher than   {$21.777\pm6.608$} of the subcase of unconstrained weights. This is  because the search space $\{\boldsymbol{p} \,|\, \boldsymbol{p}\geq \boldsymbol{0} \}$ in the former is only part of that $\{\boldsymbol{p} \,|\, \boldsymbol{p}\in\mathbb{R}^4 \}$ of the latter subcase.

\medskip

In summary of  all experiments in this subsection, we   conclude: 
(i) the proposed method can learn a   cost function (and a time-warping function) from sparse keyframes for motion planning in an unmodeled environment;   (ii) since the method jointly  learns a time-warping function, the time stamps of keyframes do not significantly influence the performance; and (iii)  learning a generic (e.g., polynomial or neural) cost function can be intuitively thought of finding some \emph{virtual attracting points} in the unknown environment, whose locations and weights will be encoded in the learned cost function.

\begin{figure*}[h]
	\centering
	\begin{subfigure}{.32\textwidth}
		\centering
		\includegraphics[width=\linewidth]{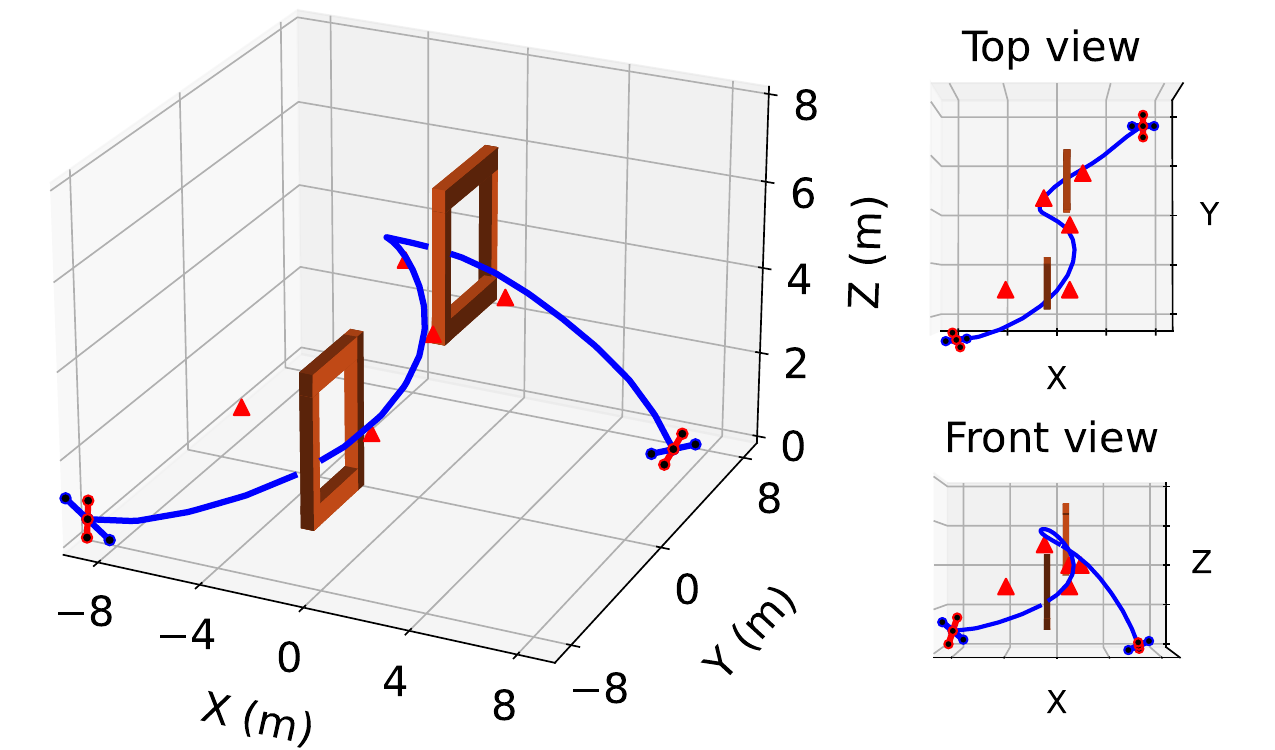}
		\caption{New initial condition 1. Generalization of        the polynomial cost learned in Fig. \ref{fig.uav.reproduce.5}.}
		\label{fig.uav.generalize.1}
	\end{subfigure}
	\hfill
	\begin{subfigure}{.32\textwidth}
		\centering
		\includegraphics[width=\linewidth]{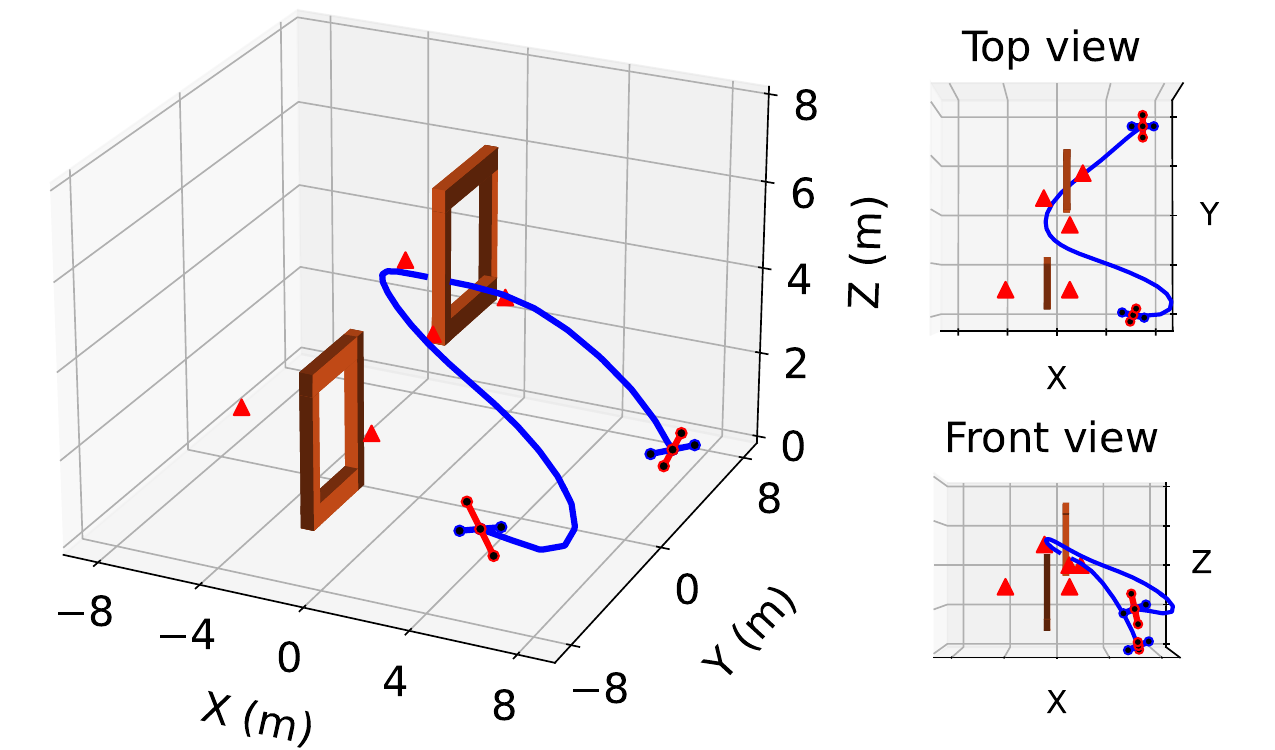}
		\caption{New initial condition 2. Generalization of the        polynomial cost learned in Fig. \ref{fig.uav.reproduce.5}.}
		\label{fig.uav.generalize.2}
	\end{subfigure}
	\hfill
	\begin{subfigure}{.32\textwidth}
		\centering
		\includegraphics[width=\linewidth]{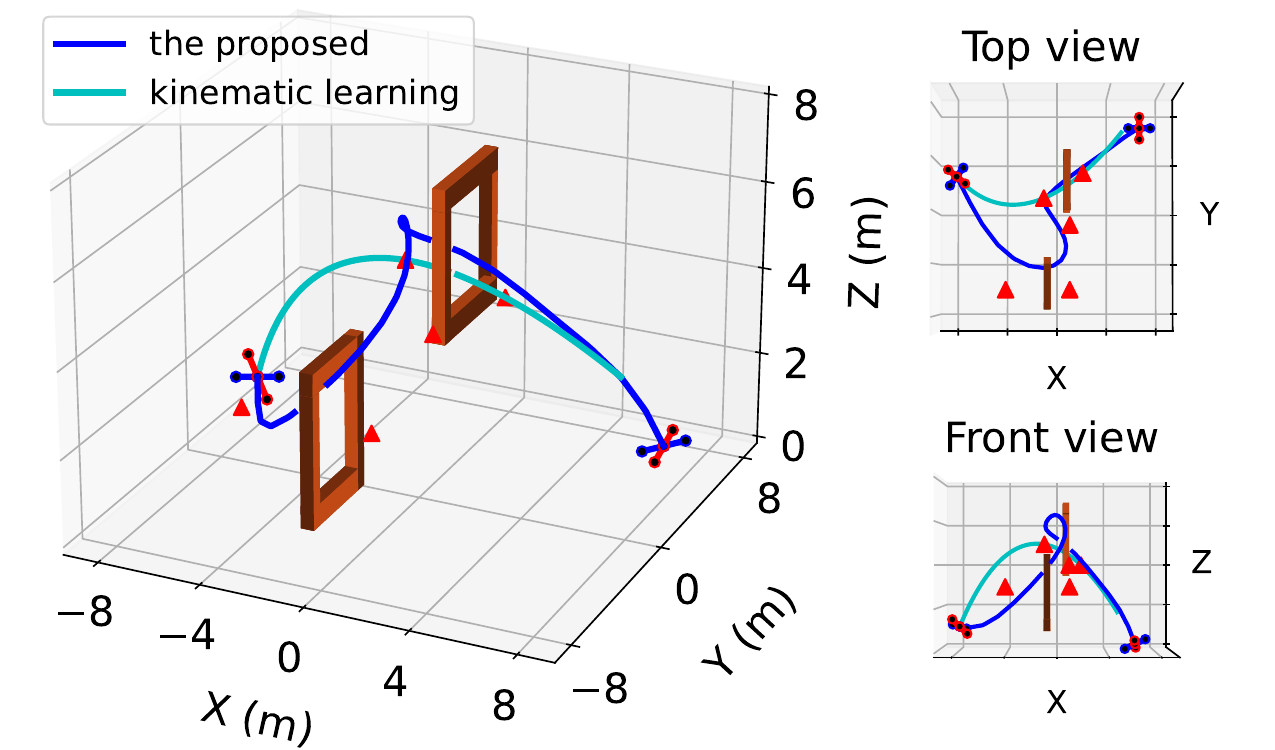}
		\caption{New initial condition 3. Comparing with generalization of kinematic learning \cite{akgun2012keyframe}}
		\label{fig.uav.generalize.3}
	\end{subfigure}
	\par\smallskip
	\begin{subfigure}{.32\textwidth}
		\centering
		\includegraphics[width=\linewidth]{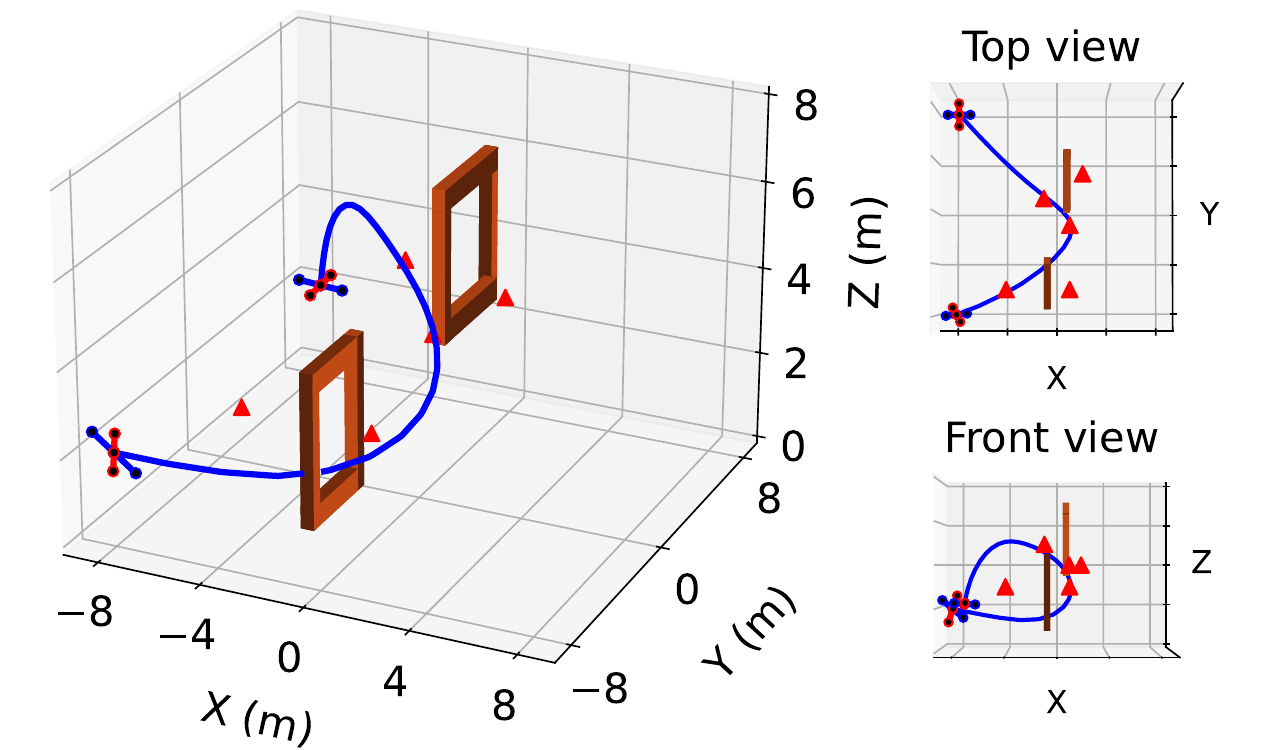}
		\caption{New landing goal.  Generalization of the        polynomial cost learned in Fig. \ref{fig.uav.reproduce.5}.}
		\label{fig.uav.generalize.4}
	\end{subfigure}
	\hfill
	\begin{subfigure}{.32\textwidth}
		\centering
		\includegraphics[width=\linewidth]{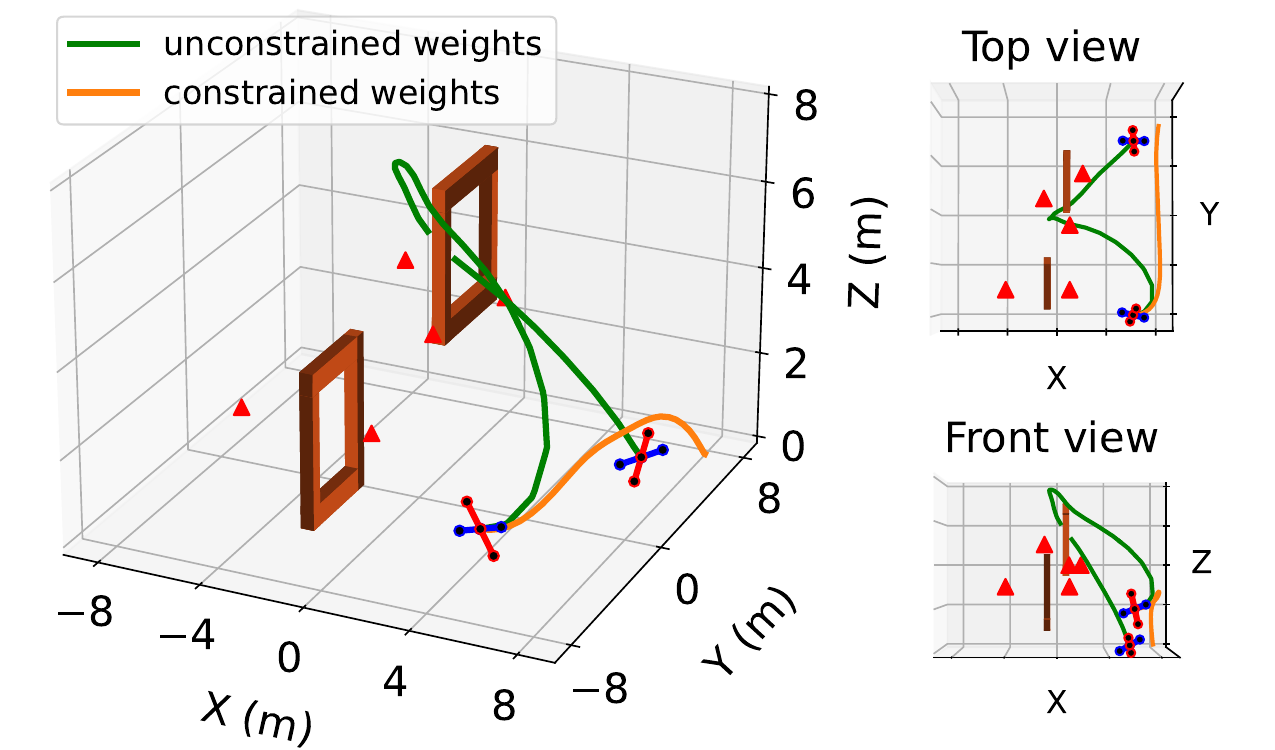}
		\caption{New initial condition 2. Generalization of        distance-to-obstacle  cost learned in Fig. \ref{fig.uav.reproduce.6}. }
		\label{fig.uav.generalize.5}
	\end{subfigure}
	\hfill
	\begin{subfigure}{.32\textwidth}
		\centering
		\includegraphics[width=\linewidth]{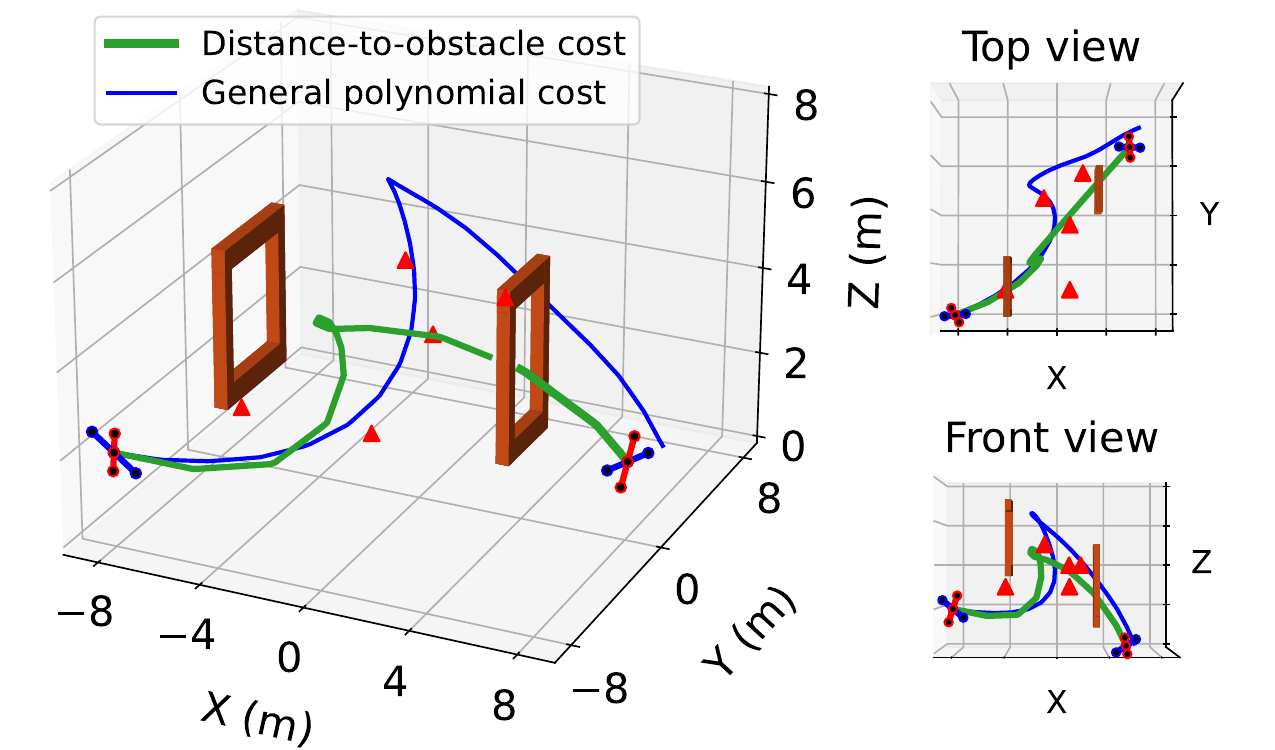}
		\caption{New placement of  obstacles. Generalization comparison between the polynomial   cost (\ref{uav_objective.1}) and distance-to-obstacle cost (\ref{uav_objective.replace}). }
		\label{fig.uav.generalize.6}
	\end{subfigure}
	\caption{Generalization test  of the    cost   functions  learned in Fig. \ref{fig.uav.reproduce}.  (a)-(c) are the generalized motion of the  polynomial cost function (learned in Fig. \ref{fig.uav.reproduce.5}) given different new initial conditions.  	
		In (c), we also compare with the generalized motion of the kinematic  learning method \cite{akgun2012keyframe}  (discussed in Section \ref{exp.compare.keyframes}). (d) shows the generalized motion of the  polynomial cost function (learned in Fig. \ref{fig.uav.reproduce.5}) given a new landing goal.  	 (e) is the  generalized motion   of  the  distance-to-obstacle cost function (learned in Fig. \ref{fig.uav.reproduce.6}) given new initial condition 2; here,  green and orange colors    correspond to the  unconstrained and constrained weights subcases, respectively. (f) is  the generalization  of the  polynomial cost function (learned in Fig. \ref{fig.uav.reproduce.5}) and  the   distance-to-obstacle cost function (learned in Fig. \ref{fig.uav.reproduce.6})  given new placement of obstacles. All quantitative measures  are  in Table  \ref{table_uav_generalized}.
	}
	
	\label{fig.uav.generalize}
\end{figure*}

\subsection{Generalization  of  Learned  Cost Functions}

In this session, we will test the generalization of the cost functions learned in the previous session. We will set the quadrotor with a  new initial condition,  a new landing goal, and  new placement of  obstacles. Given these new conditions, we use the learned cost and time-warping functions  to plan the motion of the quadrotor, respectively. We check if the  motion plan can successfully achieve the task goal: flying through the gates and landing near the goal position.   
Quantitatively, we evaluate the generalization performance by calculating the averaged distance between the generalized motion and  keyframes and the averaged distance between the generalized motion and the centers of obstacles (gates).

\subsubsection{New Initial Conditions} In Fig. \ref{fig.uav.generalize.1}-Fig. \ref{fig.uav.generalize.3}, we  test the generalization  of the   cost function learned in  Fig. \ref{fig.uav.reproduce.4} to new initial conditions (the landing position is the same as the one in the learning stage in the previous session). Here, we use the  following new initial conditions, as also visualized in Fig. \ref{fig.uav.generalize.1}-Fig. \ref{fig.uav.generalize.3}, respectively,

\emph{New initial condition 1:}   position $\boldsymbol{r}_I(0){=}[{-}8,{-}10,1]\tran$,   attitude  quaternion $\boldsymbol{q}_{B/I}(0){=}[0.88, -0.42, 0.19, 0.14]\tran$,    velocity  $\boldsymbol{v}_I(0){=}[15, 0, 0]\tran$, and angular velocity $\boldsymbol{\omega}_B(0){=}[0, 0, 0]\tran$.

\emph{New initial condition 2:}   position $\boldsymbol{r}_I(0){=}[6, -8, 2]\tran$,   attitude quaternion $\boldsymbol{q}_{B/I}(0){=}[0.88,-0.45,-0.05,-0.15]\tran$,    velocity  $\boldsymbol{v}_I(0){=}[10, 0, 0]\tran$, and angular velocity $\boldsymbol{\omega}_B(0){=}[0, 0, 0]\tran$.

\emph{New initial condition 3:}  position $\boldsymbol{r}_I(0){=}[-8, 5, 1]\tran$,   attitude quaternion $\boldsymbol{q}_{B/I}(0)=[0.88,0.14,0.14,0.43]\tran$,    velocity  $\boldsymbol{v}_I(0){=}[10, -20, 0]\tran$,  and angular velocity $\boldsymbol{\omega}_B(0){=}[0, 0, 0]\tran$.

Note that the above new initial conditions  are very different from the ones used in  learning stage (Fig. \ref{fig.uav.reproduce}). For each new initial condition, the generalized motion  is shown in Fig. \ref{fig.uav.generalize.1}-Fig. \ref{fig.uav.generalize.3}, respectively. Fig. \ref{fig.uav.generalize.3}   also plots the generalized motion (cyan color) of the  kinematic  learning  method \cite{akgun2012keyframe} as a comparison.
In Fig. \ref{fig.uav.generalize.5}, we  plot the  generalized motion from the   learned distance-to-obstacle  cost function (\ref{uav_objective.replace})    in Fig. \ref{fig.uav.reproduce.6}. Here, the green line corresponds to  the unconstrained weights  subcase and the orange to the  constrained weights subcase. The quantitative measures for all  generalized motion in Fig. \ref{fig.uav.generalize} are  in Table  \ref{table_uav_generalized}.

	\begin{table}[h]
	\centering
	\caption	{Measure of the  generalized motion. }
	\begin{tabular}{cccccc}
		\toprule
		Fig.  & Avg. distance to keyframes & Avg. distance to center of gate
		\\		\midrule
		\ref{fig.uav.generalize.1} &  1.460 & 0.863 \\
		\midrule
		\ref{fig.uav.generalize.2} & 1.014 & 1.187  \\
		\midrule
		\ref{fig.uav.generalize.3} & 		
		\begin{tabular}{@{}c@{}} 1.715 (the proposed)
			\\ 3.428 (kinematic learning [41])
		\end{tabular} &  \begin{tabular}{@{}c@{}} 1.020 (the proposed)
			\\ 3.660 (kinematic learning [41])
		\end{tabular}
		\\
		\midrule
		\ref{fig.uav.generalize.4}  &  1.841 & 1.492\\
		\midrule
		\ref{fig.uav.generalize.5} &	
		\begin{tabular}{@{}c@{}} 1.641	(unconstrained weights)
			\\ 7.580  (constrained weights)
		\end{tabular}&
		\begin{tabular}{@{}c@{}} 1.423	(unconstrained weights)
			\\ 7.950  (constrained weights)
		\end{tabular}
		\\
		\midrule
		\ref{fig.uav.generalize.6}& 	\begin{tabular}{@{}c@{}} 1.902 (dist.-to-obstacle cost)
			\\ 1.514 (polynomial cost)
		\end{tabular} &
		\begin{tabular}{@{}c@{}} 1.617 (dist.-to-obstacle cost)
			\\ 3.186 (polynomial cost)
		\end{tabular} 
		\\
		\bottomrule
	\end{tabular}
	\begin{tablenotes}\footnotesize
		\item[*] A keyframe's distance  to a trajectory is  the distance between this keyframe  and its nearest point on the trajectory, and we  average the distance over all keyframes.
	\end{tablenotes}
	\label{table_uav_generalized}
\end{table}

\subsubsection{New Landing Goal} Fig. \ref{fig.uav.generalize.4}   tests the generalization  of the    cost function  learned in  Fig. \ref{fig.uav.reproduce.4}   to a new landing goal. 
The initial condition is as follows:    position $\boldsymbol{r}_I(0){=}[-8, -8, 2]\tran$,   attitude quaternion  $\boldsymbol{q}_{B/I}(0){=}[0.88,-0.42,0.19,0.14]\tran$,   velocity  $\boldsymbol{v}_I(0){=}[15, 5, -2]\tran$, and  angular velocity $\boldsymbol{\omega}_B(0){=}[0, 0, 0]\tran$. We set the new landing goal  to  $\boldsymbol{r}_I^\text{g}=[-8, 8, 2]\tran$ and  $\boldsymbol{q}_{B/I}^{\text{g}}=[0.97,0,0,0.25]\tran$. The  quantitative measure for the  generalized motion is  in Table  \ref{table_uav_generalized}.

\subsubsection{New Placement of  Obstacles}\label{section.varyingobstacles}

Fig. \ref{fig.uav.generalize.6} tests  the generalization  of the  learned cost functions under new placement of  obstacles. We change the positions of  two gates and then use the learned cost functions to generate  new motion in the new environment. The initial condition is  the same as the one in Fig. \ref{fig.uav.generalize.4} and the landing goal as that in Fig. \ref{fig.uav.generalize.1}.  Fig. \ref{fig.uav.generalize.6} shows the generalized motion of the  distance-to-obstacle  cost function (\ref{uav_objective.replace}) (unconstrained weights, learned in Fig. \ref{fig.uav.reproduce.6}) and the   polynomial cost function (\ref{uav_objective.1}) (learned in Fig. \ref{fig.uav.reproduce.5}). Note that in the generalization of the distance-to-obstacle cost  (\ref{uav_objective.replace}),  $o_i$ is set as the obstacle's new location.  The quantitative measure for the generalized motion is in Table  \ref{table_uav_generalized}.

\smallskip

\subsubsection{Result Analysis}\label{section.analysis}
With the new initial conditions and new landing goal,  Fig. \ref{fig.uav.generalize.1}- \ref{fig.uav.generalize.4} show that the generalized motion can still follow the  keyframes,  pass through the gates, and land near the goal. Fig. \ref{fig.uav.generalize.3} also shows that  the generalization of the  kinematic learning   \cite{akgun2012keyframe} fails to fly through both gates.  As discussed in Section \ref{exp.compare.keyframes},  since \cite{akgun2012keyframe}   focuses on learning a low-level kinematic representation, it has limited generalizability particularly  when the new conditions are very different from ones in learning. In contrast,  a learned  cost function can be shared across different motion conditions. Thus, learning cost functions shows better generalizability. 
Fig. \ref{fig.uav.generalize.5}
also shows that the generalization of the distance-to-obstacle cost function (\ref{uav_objective.replace}) (unconstrained weights) is comparable to that  in  Fig. \ref{fig.uav.generalize.2}.

The special attention should be paid to Fig. \ref{fig.uav.generalize.2} and Fig. \ref{fig.uav.generalize.5} (unconstrained weights), where the quadrotor seems to have ignored the first two keyframes (and hence the left gate). This could be explained by Bellman's principle of optimality \cite{bellman1966dynamic}. Specifically, the  motion in Fig. \ref{fig.uav.generalize.2}  can be interpreted as the final segment of the `complete'  trajectory in Fig. \ref{fig.uav.reproduce.5}, i.e., it can be viewed as the solution to a sub-problem, for which the initial condition starts from a middle point of a `complete' trajectory and minimizes the  \emph{remaining} cost-to-go. In other words, if a complete trajectory from the initial start to a goal is optimal with respect to a cost function, the sub-trajectory of this complete trajectory from any middle point to the goal is also optimal with respect to the same cost function. Thus, the quadrotor motion in  Fig. \ref{fig.uav.generalize.2} and Fig. \ref{fig.uav.generalize.5} is continuing to finish the rest optimal motion instead of flying back to pass through the first gate.

Fig. \ref{fig.uav.generalize.6} shows the  generalization of the learned distance-to-obstacle cost function (\ref{uav_objective.replace})  versus that of the learned generic polynomial cost function (\ref{uav_objective.1}) in a varying environment. With Fig. \ref{fig.uav.generalize.6} and  Table \ref{table_uav_generalized}, one can conclude that  the  polynomial cost function generalizes poorly to  new placement of  obstacles, compared to the distance-to-obstacle cost function. Specifically, the generalized motion of the learned polynomial cost function still tries to follow the original keyframes instead of going through the new gates: its distance to the keyframes is $1.514$ versus  the distance-to-obstacle cost function's $1.902$, while its distance to the centers of gates is $3.186$ versus  the distance-to-obstacle cost's $1.617$. This is understandable because the learned polynomial cost function only `remembers' the representation of the  keyframes in the original environment and is unaware of the obstacle changes in the new environment.  On the contrary,  the distance-to-obstacle cost function (\ref{uav_objective.replace}) is  defined on  the locations of obstacles, and can be updated with the new locations of obstacles. Hence, in the new environment in Fig. \ref{fig.uav.generalize.6}, the generalized motion of the distance-to-obstacle cost tries to go through the new gates. As indicated in Table \ref{table_uav_generalized}, the generalization of the distance-to-obstacle cost function has a smaller distance ($1.617$) to the new gates than the polynomial cost does ($3.186$). This can also be visualized in Fig. \ref{fig.uav.generalize.6}, where the motion of the distance-to-obstacle cost is attempting to reach the left gate (although it is not successfully passing through it). The above results suggest that an environment-dependent formulation of cost functions, such as a cost function that is defined on both robot state and environment features, could generalize better in a varying environment. But one also needs to note that such a formulation additionally requires the knowledge/model of the environment features. More discussion of the cost formulations is given in  Section \ref{section_choice}.

In summary of all the above experiments and analyses, we   conclude that (i) the proposed method can learn a  cost function from a small number of keyframes;  (ii) 
the learned cost function shows good generalization to unseen motion conditions; and (iii) to generalize to varying environments, an environment-informed formulation of cost functions would be needed, such as the cost function formulation which depends on both robot's state and environment features.

\section{Discussion}\label{section_discussions}

This section further provides discussion on some aspects of the proposed method.

\subsection{Why Do  Keyframes Suffice?}\label{section_discussions.a}

We  provide one explanation for why sparse  keyframes can suffice to recover a cost function. Consider  problem   (\ref{problem}). For   trajectory $\boldsymbol{\xi}_{\boldsymbol{\theta}}$ produced by  optimal control system  (\ref{ocsys}), since we are only interested in the trajectory points $\boldsymbol{\xi}_{\boldsymbol{\theta}}{(\tau_i)}$ at the   time stamps $\tau_i$ ($1\leq i\leq N$), we discretize the optimal control system at these time steps, yielding  \cite{patterson2014gpops}
\begin{subequations}\label{discussion_discrete}
	\begin{align}
	\text{dynamics:}& \,\,\, \boldsymbol{x}_{i+1}={\boldsymbol{\bar f}}(\boldsymbol{x}_i,\bar{\boldsymbol{u}}_i, \boldsymbol{\theta}), \quad \boldsymbol{x}_0=\boldsymbol{x}(0),\\
	\text{objective:}& \,\,\, J(\boldsymbol{\theta})=\sum_{i=0}^{N-1}\bar{c}(\boldsymbol{x}_i,\bar{\boldsymbol{u}}_i,\boldsymbol{{\theta}})+ \bar{h}(\boldsymbol{x}_N,\bar{\boldsymbol{u}}_N, \boldsymbol{\theta}),
	\end{align}
\end{subequations}
where  we denote $\boldsymbol{x}_i=\boldsymbol{x}(\tau_{i})$, and discrete-time ${\boldsymbol{ \bar f}}$ satisfies
\begin{align}
\boldsymbol{x}_{i+1}&={\boldsymbol{\bar f}}(\boldsymbol{x}_i,\bar{\boldsymbol{u}}_i, \boldsymbol{\theta})
=\boldsymbol{x}_i+\int_{\tau_i}^{\tau_{i+1}}v_{\boldsymbol{\beta}}(\tau)\boldsymbol{f}(\boldsymbol{x}(\tau),\boldsymbol{u}(\tau)) d\tau,\nonumber
\end{align}
and the discrete version of the  cost function satisfies
\begin{align}
\bar{c}(\boldsymbol{x}_i, \bar{\boldsymbol{u}}_i, \boldsymbol{\theta})&=\int_{\tau_i}^{\tau_{i+1}}v_{\boldsymbol{\beta}}(\tau)c_{\boldsymbol{p}}(\boldsymbol{x}(\tau),\boldsymbol{u}(\tau))d\tau,\nonumber\\
\bar{h}(\boldsymbol{x}_N, \bar{\boldsymbol{u}}_N, \boldsymbol{\theta})&=\int_{\tau_N}^{T}v_{\boldsymbol{\beta}}(\tau)c_{\boldsymbol{p}}(\boldsymbol{x}(\tau),\boldsymbol{u}(\tau))d\tau + h_{\boldsymbol{p}}(\boldsymbol{x}(T)).\nonumber
\end{align}
Here, the new  input $\bar{\boldsymbol{u}}_i\in\mathbb{R}^{d}$ in $\bar{\boldsymbol{f}}$ may not necessarily have the same dimension as  $\boldsymbol{u}(\tau)\in \mathbb{R}^{n}$ in the original   $\boldsymbol{f}$, e.g., $\bar{\boldsymbol{u}}_i$ contains all   possible controls over  time range $[\tau_{i}, \tau_{i+1}]$ \cite{patterson2014gpops}. The  solution $\{{\boldsymbol{x}_{0:N}}, {\bar{\boldsymbol{u}}_{0:N}}\}$ to the discrete-time optimal control system (\ref{discussion_discrete})  satisfies the KKT conditions:
\begin{equation}\label{discussion_kkt}
\begin{aligned}
\boldsymbol{x}_{i+1}&=\bar{\boldsymbol{f}}(\boldsymbol{x}_i,\bar{\boldsymbol{u}}_i, \boldsymbol{\theta}), \quad& i=0,\cdots N-1,\\
\boldsymbol{\lambda}_{i}&=\frac{\partial\bar{c}}{\partial\boldsymbol{x}_i}+\frac{\partial \bar{\boldsymbol{f}}\tran}{\partial\boldsymbol{x}_i}\boldsymbol{\lambda}_{i+1}, &\quad i=1,\cdots N-1,\\
\boldsymbol{0}&=\frac{\partial\bar{c}}{\partial\bar{\boldsymbol{u}}_i}+\frac{\partial \bar{\boldsymbol{f}}\tran}{\partial\bar{\boldsymbol{u}}_i}\boldsymbol{\lambda}_{i+1}, &\quad i=0,\cdots N-1,\\
\boldsymbol{\lambda}_{N}&=\frac{\partial\bar{h}}{\partial\boldsymbol{x}_N},\quad \frac{\partial\bar{h}}{\partial\bar{\boldsymbol{u}}_N}=\boldsymbol{0} &  i=N.
\end{aligned}
\end{equation}
The output of the discrete-time system (\ref{discussion_discrete}) can be overloaded by $\boldsymbol{y}(\tau_i)=\boldsymbol{g}(\boldsymbol{x}_i,\bar{\boldsymbol{u}}_i)$. To simplify analysis, we  assume that  keyframes $\mathcal{D}$  are realizable by a $\boldsymbol{\theta}$. Then,
\begin{equation}\label{discussion_outpug}
\boldsymbol{y}^*({\tau}_i)=\boldsymbol{g}(\boldsymbol{x}_i,\bar{\boldsymbol{u}}_i).
\end{equation}

Given the keyframes $\mathcal{D}$ in (\ref{corrections}),  recovering a cost function can be viewed as  a problem of solving a set of non-linear equations in (\ref{discussion_kkt}) and (\ref{discussion_outpug}), where  unknowns are    $\{\boldsymbol{x}_{1:N}, \bar{\boldsymbol{u}}_{0:N}, \boldsymbol{\lambda}_{1:N}, \boldsymbol{\theta}\}\in\mathbb{R}^{2Nn+(N+1)d+(r+s)}$, and  the total number of constraints  (equations)  are $2Nn+(N+1)d+No$. Here, $(r+s)$ is the dimension of $\boldsymbol{\theta}$ and $o$ is the dimension of $\boldsymbol{y}$. A necessary condition to  uniquely determine   $\{\boldsymbol{x}_{1:N}, \bar{\boldsymbol{u}}_{0:N}, \boldsymbol{\lambda}_{1:N}, \boldsymbol{\theta}\}$  requires the number of constraints to be no less than the number of unknowns, yielding
\begin{equation}\label{discussion_condition}
N\geq \frac{r+s}{o}.
\end{equation}
On the other hand, if   (\ref{discussion_condition}) is not fulfilled or given   $\mathcal{D}$ is less informative,     the unknowns then cannot be uniquely determined, which means that there might exist multiple $\boldsymbol{\theta}$s such that all resulting trajectories pass the same sparse keyframes. This case has been shown  in Section \ref{experiment1.1} (Fig. \ref{fig.robotarm.trueresults}).

Note that the above discussion uses a perspective different from the development of this paper. It should be noted that the above explanation fails to explain the case where the given keyframes are not realizable: $\min_{\boldsymbol{\theta}} L(\boldsymbol{\xi}_{\boldsymbol{\theta}},\mathcal{D})>0$, 
e.g.,   sub-optimal data as   in Section \ref{section.robotarm.nonoptimal}. We leave its further exploration as one future direction of this work.

\subsection{Cost Function Formulation}\label{section_choice}
In general, there are two types of cost function formulations, as discussed below.
\subsubsection{Cost  Depending Purely on Robot States}
	The first type of cost function formulation can be written as $c_{\boldsymbol{p}}(\boldsymbol{x},\boldsymbol{u})$, which only depends on robot state and input $(\boldsymbol{x}, \boldsymbol{u})$. The  polynomial cost function (\ref{uav_objective.1}) in Section \ref{section_application} belongs to this type. This formulation type can generalize well to different  motion conditions, e.g., new initial condition and new goals, as shown in Fig. \ref{fig.uav.generalize.1} - Fig. \ref{fig.uav.generalize.5}. However, it cannot generalize to  varying environments as  in Fig. \ref{fig.uav.generalize.6}, as the environment information is not explicitly captured in this  cost formulation. 

\subsubsection{Cost   Depending on  Robot States and Environment Features}
The second type of cost formulations can be written as $c_{\boldsymbol{p}}(\boldsymbol{x},\boldsymbol{u}, \boldsymbol{o})$, which depends on both the robot state-input $(\boldsymbol{x},\boldsymbol{u})$ and the environment features $\boldsymbol{o}$. Here, $\boldsymbol{o}$ should be  given for the environment where the robot is trained. Demonstrations from different environments can also be used as the training data. 	The cost functions  in (\ref{exp.robotarm.cost}) and (\ref{uav_objective.replace}) belong to this type. One advantage of this formulation is that it has the ability to generalize to a  new environment  given its  environment features $\boldsymbol{o}$.
Section \ref{section.varyingobstacles} has shown such an advantage by comparing with  the first type of cost formulation. At the same time, one should note that the second type of cost formulation requires the knowledge of environment features $\boldsymbol{o}$, which may need additional modeling effort.

\subsubsection{Running Cost and Final Cost} The cost function in (\ref{costfun}) includes two terms: a running cost term  $c(\cdot)$ and  a final cost term  $h(\cdot)$. If no knowledge about the task goal is available,  one can use a (deep) neural network to represent both costs,  as shown in Section \ref{section.learningneural}. Since a neural cost function is usually goal-blind, the training data needs to include a keyframe at the goal.  If the task goal is known, such as in motion planning  in Section \ref{section_application}, the final cost  $h(\cdot)$ can be set to the distance-to-goal cost, and the running cost    $c(\cdot)$ is to be tuned. Tuning a running cost will determine how the robot moves to the goal. This has been shown in Section \ref{section_application}.

\subsubsection{Limitation}
We should note that whatever a cost formulation is, the proposed method requires all functions to be differentiable. This can be a limitation of the proposed method, compared to some existing feature-based IRL methods such as max-entropy IRL \cite{ziebart2008maximum}, which permits non-differentiable features. How to extend the proposed method to non-differentiable systems is a topic for our future work.

\subsection{Convergence and Numerical Integration Error}

\subsubsection{Algorithm Convergence}
The proposed Continuous PDP  solves a bi-level optimization problem  (\ref{problem}) using gradient descent. It treats the  trajectory $\boldsymbol{\xi}_{\boldsymbol{\theta}}$ of the  inner-level optimal control system simply as an `implicit' differentiable  function of the  system  parameter $\boldsymbol{\theta}$.  Generally, bi-level optimization is known to be strongly NP-hard \cite{hansen1992new,sinha2017review}.  Under certain assumptions, one can prove that the gradient-descent method can converge to a   stationary point \cite{ji2021bilevel}. With further assumptions on the outer-level and inner-level problems,   such as convexity and smoothness,  \cite{ghadimi2018approximation} shows that the gradient-descent method could converge to the global solution. However, in our case,   the requirement of convexity is too restricted to optimal control systems (\ref{ocsys}). As a future direction of this  work, 
we will try to explore the milder conditions for its convergence.

\subsubsection{Numerical  Integration Error} Another issue that might arise is the numerical integration error in solving the gradient of the inner-level trajectory using Lemma \ref{theorem1}, as it requires integrating several ODEs in both backward and forward passes. However, our previous experimental experience shows that due to the side effect of a time-warping function,  numerical integration error/stability can be potentially mitigated. A similar process has also been successfully used in some optimal control software such as  \cite{patterson2014gpops}.

A side effect of using a time-warping function $t=w(\tau)$ is that one can scale a long-horizon integration into a smaller horizon problem  by time-warping transformation,  then re-scale the solution back after integration (some refinement can be done afterwards).  For example,  $\int_{0}^{t_f}c(\boldsymbol{x}(t),\boldsymbol{u}(t))dt$ in (\ref{costfun})  over $[0, t_f]$ can be transformed to  $\int_{0}^{T}\frac{dw(\tau)}{d\tau}c(\boldsymbol{x}(\tau),\boldsymbol{u}(\tau))d\tau$  in (\ref{costfun_wrap2}) over the new horizon $[0, T]$, using the time-warping function  $t_f=w(T)$. In our problem of interest, since $T$ is given, one can manually pick a relatively small horizon $T<t_f$ and a small integration step size to mitigate the error of  numerical integration. In our previous experiments,  we set the keyframe horizon  as  $T=1$ for good numerical integration accuracy.  This time-warping trick has been successfully adopted by some optimal control software such as \cite{patterson2014gpops} for numerical stability. One  important caveat is that by using the time-warping transformation  $t=w(\tau)$, as shown from (\ref{costfun}) to (\ref{costfun_wrap2}), we have changed the original integrand  $c(\boldsymbol{x}(t),\boldsymbol{u}(t))$ to the new $\frac{dw(\tau)}{d\tau}c(\boldsymbol{x}(\tau),\boldsymbol{u}(\tau))$. Hence, if one wants to significantly decrease the horizon, i.e., $T\ll t_f$,  $\frac{dw(\tau)}{d\tau}$ would be very large, which may increase the stiffness of $\frac{dw(\tau)}{d\tau}c(\boldsymbol{x}(\tau),\boldsymbol{u}(\tau))$, causing numerical instability. Although we have rarely encountered such numerical issues in our previous experiments with $T=1$s, one might be cautious when  handling stiff ODEs/systems.

\subsection{Model-free versus Model-based}
The  formulation in this paper assumes robot dynamics to be known. We would point out that the proposed \emph{Continuous PDP is also able to solve model-free IOC/IRL, i.e.,  jointly learning a dynamics model and a cost function from keyframes}. To do that, one needs to replace the known dynamics (\ref{dyn}) with a  \emph{parameterized dynamics model}, which should be  differentiable. The Continuous PDP can update all parameters (including both dynamics and objective parameters) using gradient descent. We refer the reviewer to our previous work Pontryagin Differentiable Programming (PDP) \cite{jin2020pontryagin,jin2021safe} (discrete-time) for the model-free IOC/IRL formulation and experiments.

In fact,  after the problem reformulation in Section III.B, as shown in (\ref{dyn_wrap2}), the parameter of the time-warping function has been absorbed into the dynamics model and becomes the unknown parameter in the dynamics. Thus,  the Continuous PDP has already shown its ability to jointly update the parameters in both the dynamics model and cost function.

\section{Conclusions}
This paper proposes the method of Continuous Pontryagin Differentiable Programming (Continuous PDP)  to enable a robot to learn an objective function from a small set of demonstrated keyframes. As the given time stamps of the keyframes may not be achievable in the robot's actual execution, the Continuous PDP  jointly finds an objective function and a time-warping function such that the robot's final motion attains the minimal discrepancy loss to the keyframes. The Continuous PDP minimizes the discrepancy loss using projected gradient descent, by efficiently computing the gradient of the optimal trajectory with respect to the tunable function parameters in the system.    The efficacy and capability of the Continuous PDP  are demonstrated in robot arm and 6-DoF quadrotor planning tasks.

\appendix[Proof of Lemma \ref{theorem1}]
We consider the equation of  Differential Pontryagin's Maximum Principle in (\ref{dpmp}). Suppose that ${H_{uu}(\tau)}$  in (\ref{matHu}) is invertible for all $0\leq \tau\leq T$. We can solve  $\frac{\partial \boldsymbol{u}_{\boldsymbol{\theta}}}{\partial\boldsymbol{\theta}}$ from (\ref{dpmp.3}):
\begin{equation}\label{diffu}
\frac{\partial \boldsymbol{u}_{\boldsymbol{\theta}}}{\partial\boldsymbol{\theta}}={-}H_{uu}^{-1}(\tau)\Big(
H_{ux}(\tau)\frac{\partial \boldsymbol{x}_{\boldsymbol{\theta}}}{\partial\boldsymbol{\theta}}{+}{G}(\tau)\tran \frac{\partial \boldsymbol{\lambda}_{\boldsymbol{\theta}}}{\partial\boldsymbol{\theta}}{+}H_{ue}(\tau)
\Big).
\end{equation}
Substituting (\ref{diffu}) into both (\ref{dpmp.1}) and (\ref{dpmp.2}) and combining the definition of matrices in (\ref{raccaticoefficient}), we have
\begin{subequations}\label{appendixequ1}
	\begin{align}
	\frac{d}{d\tau}(\frac{\partial \boldsymbol{x}_{\boldsymbol{\theta}}}{\partial\boldsymbol{\theta}})&=A(\tau)\frac{\partial \boldsymbol{x}_{\boldsymbol{\theta}}}{\partial\boldsymbol{\theta}}-R(\tau)\frac{\partial \boldsymbol{\lambda}_{\boldsymbol{\theta}}}{\partial\boldsymbol{\theta}}+M(\tau), \label{appendixequ1.1}\\
	-\frac{d}{d\tau}(\frac{\partial \boldsymbol{\lambda}_{\boldsymbol{\theta}}}{\partial\boldsymbol{\theta}})&=Q(\tau)\frac{\partial \boldsymbol{x}_{\boldsymbol{\theta}}}{\partial\boldsymbol{\theta}}+A(\tau)^{\prime}\frac{\partial \boldsymbol{\lambda}_{\boldsymbol{\theta}}}{\partial\boldsymbol{\theta}}+N(\tau) \label{appendixequ1.2}.
	\end{align}
\end{subequations}
Motivated by (\ref{dpmp.4}), we assume 
\begin{equation}\label{appendixassumptionequ}
\frac{\partial \boldsymbol{\lambda}_{\boldsymbol{\theta}}}{\partial\boldsymbol{\theta}}=P(\tau)\frac{\partial \boldsymbol{x}_{\boldsymbol{\theta}}}{\partial\boldsymbol{\theta}}+W(\tau),
\end{equation}
with  $P(\tau)\in\mathbb{R}^{n\times n}$ and $W(\tau)\in\mathbb{R}^{n\times (s+r)}$, $0\leq \tau \leq T$, are two time-varying matrices. Of course, the above (\ref{appendixassumptionequ}) holds for $\tau=T$ because of (\ref{dpmp.4}), if 
\begin{equation}
P(\tau)= H_{xx}(T) \quad\text{and}\quad W(\tau)= H_{xe}(T).
\end{equation}

Substituting (\ref{appendixassumptionequ}) to (\ref{appendixequ1.1}) and (\ref{appendixequ1.2}), respectively, to eliminate $\frac{\partial \boldsymbol{x}_{\boldsymbol{\theta}}}{\partial\boldsymbol{\theta}}$,  we obtain the following
\begin{subequations}
	\begin{align}
	\frac{d}{d\tau}(\frac{\partial \boldsymbol{x}_{\boldsymbol{\theta}}}{\partial\boldsymbol{\theta}})&{=}(A-RP)\frac{\partial \boldsymbol{x}_{\boldsymbol{\theta}}}{\partial\boldsymbol{\theta}}+(-RW+M),\label{appendixequ2.1}\\
	{-}\dot{P}\frac{d}{d\tau}(\frac{\partial \boldsymbol{x}_{\boldsymbol{\theta}}}{\partial\boldsymbol{\theta}})&{=}(Q{+}\dot{P}{+}A\tran P)\frac{\partial \boldsymbol{x}_{\boldsymbol{\theta}}}{\partial\boldsymbol{\theta}}{+}(A^{\prime}W{+}N{+}\dot{W}),\label{appendixequ2.2}
	\end{align}
\end{subequations}
where $\dot{P}=\frac{d P(\tau)}{d\tau}$, $\dot{W}=\frac{d W(\tau)}{d\tau}$, and we here have suppressed the dependence of  $\tau$ for all time-varying matrices. By multiplying $(-\dot{P})$ on both sides of (\ref{appendixequ2.1}), and equaling the left sides of (\ref{appendixequ2.1}) and (\ref{appendixequ2.2}), we have
\begin{align}
&(-PA+PRP)\frac{\partial \boldsymbol{x}_{\boldsymbol{\theta}}}{\partial\boldsymbol{\theta}}+(PRW-PM)\nonumber\\
=&(Q+\dot{P}+A\tran P)\frac{\partial \boldsymbol{x}_{\boldsymbol{\theta}}}{\partial\boldsymbol{\theta}}+(A\tran W+N+\dot{W}).
\end{align}
The above equation holds if 
\begin{subequations}
	\begin{align}
	-PA+PRP&=Q+\dot{P}+A\tran P,\\
	PRW-PM&=A\tran W+N+\dot{W},
	\end{align}
\end{subequations}
which  directly are (\ref{cricc}). Substituting (\ref{appendixassumptionequ}) into (\ref{diffu}) yields (\ref{citer.1}), and (\ref{citer.2}) directly results from (\ref{dpmp.1}). This completes the proof. \qed

\bibliographystyle{IEEEtran}
\bibliography{trobib}

\begin{thebibliography}{10}
\providecommand{\url}[1]{#1}
\csname url@samestyle\endcsname
\providecommand{\newblock}{\relax}
\providecommand{\bibinfo}[2]{#2}
\providecommand{\BIBentrySTDinterwordspacing}{\spaceskip=0pt\relax}
\providecommand{\BIBentryALTinterwordstretchfactor}{4}
\providecommand{\BIBentryALTinterwordspacing}{\spaceskip=\fontdimen2\font plus
\BIBentryALTinterwordstretchfactor\fontdimen3\font minus
  \fontdimen4\font\relax}
\providecommand{\BIBforeignlanguage}[2]{{%
\expandafter\ifx\csname l@#1\endcsname\relax
\typeout{** WARNING: IEEEtran.bst: No hyphenation pattern has been}%
\typeout{** loaded for the language `#1'. Using the pattern for}%
\typeout{** the default language instead.}%
\else
\language=\csname l@#1\endcsname
\fi
#2}}
\providecommand{\BIBdecl}{\relax}
\BIBdecl

\bibitem{ravichandar2020recent}
H.~Ravichandar, A.~S. Polydoros, S.~Chernova, and A.~Billard, ``Recent advances
  in robot learning from demonstration,'' \emph{Annual Review of Control,
  Robotics, and Autonomous Systems}, vol.~3, 2020.

\bibitem{denivsa2015learning}
M.~Deni{\v{s}}a, A.~Gams, A.~Ude, and T.~Petri{\v{c}}, ``Learning compliant
  movement primitives through demonstration and statistical generalization,''
  \emph{IEEE/ASME transactions on mechatronics}, vol.~21, no.~5, pp.
  2581--2594, 2015.

\bibitem{moro2018learning}
C.~Moro, G.~Nejat, and A.~Mihailidis, ``Learning and personalizing socially
  assistive robot behaviors to aid with activities of daily living,'' \emph{ACM
  Transactions on Human-Robot Interaction}, vol.~7, no.~2, pp. 1--25, 2018.

\bibitem{kuderer2015learning}
M.~Kuderer, S.~Gulati, and W.~Burgard, ``Learning driving styles for autonomous
  vehicles from demonstration,'' in \emph{IEEE International Conference on
  Robotics and Automation}, 2015, pp. 2641--2646.

\bibitem{pomerleau1991efficient}
D.~A. Pomerleau, ``Efficient training of artificial neural networks for
  autonomous navigation,'' \emph{Neural computation}, vol.~3, no.~1, pp.
  88--97, 1991.

\bibitem{englert2013probabilistic}
P.~Englert, A.~Paraschos, M.~P. Deisenroth, and J.~Peters, ``Probabilistic
  model-based imitation learning,'' \emph{Adaptive Behavior}, vol.~21, no.~5,
  pp. 388--403, 2013.

\bibitem{calinon2007learning}
S.~Calinon, F.~Guenter, and A.~Billard, ``On learning, representing, and
  generalizing a task in a humanoid robot,'' \emph{IEEE Transactions on
  Systems, Man, and Cybernetics}, vol.~37, no.~2, pp. 286--298, 2007.

\bibitem{rahmatizadeh2018vision}
R.~Rahmatizadeh, P.~Abolghasemi, L.~B{\"o}l{\"o}ni, and S.~Levine,
  ``Vision-based multi-task manipulation for inexpensive robots using
  end-to-end learning from demonstration,'' in \emph{IEEE International
  Conference on Robotics and Automation}, 2018, pp. 3758--3765.

\bibitem{torabi2018behavioral}
F.~Torabi, G.~Warnell, and P.~Stone, ``Behavioral cloning from observation,''
  in \emph{International Joint Conference on Artificial Intelligence}, 2018,
  pp. 4950--4957.

\bibitem{abbeel2004apprenticeship}
P.~Abbeel and A.~Y. Ng, ``Apprenticeship learning via inverse reinforcement
  learning,'' in \emph{International Conference on Machine Learning}, 2004, pp.
  1--8.

\bibitem{ng2000algorithms}
A.~Y. Ng, S.~J. Russell \emph{et~al.}, ``Algorithms for inverse reinforcement
  learning.'' in \emph{International Conference of Machine Learning}, vol.~1,
  2000, p.~2.

\bibitem{moylan1973nonlinear}
P.~Moylan and B.~Anderson, ``Nonlinear regulator theory and an inverse optimal
  control problem,'' \emph{IEEE Transactions on Automatic Control}, vol.~18,
  no.~5, pp. 460--465, 1973.

\bibitem{macglashan2015between}
J.~MacGlashan and M.~L. Littman, ``Between imitation and intention learning,''
  in \emph{International Joint Conference on Artificial Intelligence}, 2015.

\bibitem{ratliff2006maximum}
N.~D. Ratliff, J.~A. Bagnell, and M.~A. Zinkevich, ``Maximum margin planning,''
  in \emph{International Conference on Machine Learning}, 2006, pp. 729--736.

\bibitem{ziebart2008maximum}
B.~D. Ziebart, A.~L. Maas, J.~A. Bagnell, and A.~K. Dey, ``Maximum entropy
  inverse reinforcement learning.'' in \emph{Association for the Advancement of
  Artificial Intelligence}, vol.~8, 2008, pp. 1433--1438.

\bibitem{mombaur2010human}
K.~Mombaur, A.~Truong, and J.-P. Laumond, ``From human to humanoid
  locomotion—an inverse optimal control approach,'' \emph{Autonomous Robots},
  vol.~28, no.~3, pp. 369--383, 2010.

\bibitem{keshavarz2011imputing}
A.~Keshavarz, Y.~Wang, and S.~Boyd, ``Imputing a convex objective function,''
  in \emph{IEEE International Symposium on Intelligent Control}.\hskip 1em plus
  0.5em minus 0.4em\relax IEEE, 2011, pp. 613--619.

\bibitem{puydupin2012convex}
A.-S. Puydupin-Jamin, M.~Johnson, and T.~Bretl, ``A convex approach to inverse
  optimal control and its application to modeling human locomotion,'' in
  \emph{International Conference on Robotics and Automation}, 2012, pp.
  531--536.

\bibitem{englert2017inverse}
P.~Englert, N.~A. Vien, and M.~Toussaint, ``Inverse kkt: Learning cost
  functions of manipulation tasks from demonstrations,'' \emph{International
  Journal of Robotics Research}, vol.~36, no. 13-14, pp. 1474--1488, 2017.

\bibitem{kingston2011time}
P.~Kingston and M.~Egerstedt, ``Time and output warping of control systems:
  Comparing and imitating motions,'' \emph{Automatica}, vol.~47, no.~8, pp.
  1580--1588.

\bibitem{osa2018algorithmic}
T.~Osa, J.~Pajarinen, G.~Neumann, J.~A. Bagnell, P.~Abbeel, J.~Peters
  \emph{et~al.}, ``An algorithmic perspective on imitation learning,''
  \emph{Foundations and Trends in Robotics}, vol.~7, no. 1-2, pp. 1--179, 2018.

\bibitem{doerr2015direct}
A.~Doerr, N.~D. Ratliff, J.~Bohg, M.~Toussaint, and S.~Schaal, ``Direct loss
  minimization inverse optimal control.'' in \emph{Robotics: Science and
  Systems}, 2015.

\bibitem{jin2018inverse}
W.~Jin, D.~Kuli{\'c}, S.~Mou, and S.~Hirche, ``Inverse optimal control from
  incomplete trajectory observations,'' \emph{International Journal of Robotics
  Research}, vol.~40, no. 6-7, pp. 848--865, 2021.

\bibitem{jin2019inverse}
W.~Jin, D.~Kuli{\'c}, J.~F.-S. Lin, S.~Mou, and S.~Hirche, ``Inverse optimal
  control for multiphase cost functions,'' \emph{IEEE Transactions on
  Robotics}, vol.~35, no.~6, pp. 1387--1398, 2019.

\bibitem{kuhn2014nonlinear}
H.~W. Kuhn and A.~W. Tucker, ``Nonlinear programming,'' in \emph{Traces and
  Emergence of Nonlinear Programming}.\hskip 1em plus 0.5em minus 0.4em\relax
  Springer, 2014, pp. 247--258.

\bibitem{pontryagin1962mathematical}
L.~S. Pontryagin, V.~G. Boltyanskiy, R.~V. Gamkrelidze, and E.~F. Mishchenko,
  \emph{The Mathematical Theory of Optimal Processes}.\hskip 1em plus 0.5em
  minus 0.4em\relax John Wiley \& Sons, Inc., 1962.

\bibitem{jin2021distributed}
W.~Jin and S.~Mou, ``Distributed inverse optimal control,'' \emph{Automatica},
  vol. 129, p. 109658, 2021.

\bibitem{mombaur2013forward}
K.~Mombaur, A.-H. Olivier, and A.~Cr{\'e}tual, ``Forward and inverse optimal
  control of bipedal running,'' in \emph{Modeling, simulation and optimization
  of bipedal walking}.\hskip 1em plus 0.5em minus 0.4em\relax Springer, 2013,
  pp. 165--179.

\bibitem{powell2009bobyqa}
M.~J. Powell, ``The bobyqa algorithm for bound constrained optimization without
  derivatives,'' \emph{Cambridge NA Report, University of Cambridge,
  Cambridge}, pp. 26--46, 2009.

\bibitem{rios2013derivative}
L.~M. Rios and N.~V. Sahinidis, ``Derivative-free optimization: a review of
  algorithms and comparison of software implementations,'' \emph{Journal of
  Global Optimization}, vol.~56, no.~3, pp. 1247--1293, 2013.

\bibitem{hatz2012estimating}
K.~Hatz, J.~P. Schloder, and H.~G. Bock, ``Estimating parameters in optimal
  control problems,'' \emph{SIAM Journal on Scientific Computing}, vol.~34,
  no.~3, pp. A1707--A1728, 2012.

\bibitem{das2020model}
N.~Das, S.~Bechtle, T.~Davchev, D.~Jayaraman, A.~Rai, and F.~Meier,
  ``Model-based inverse reinforcement learning from visual demonstrations,'' in
  \emph{Conference on Robotic Learning}, pp. 1930--1942.

\bibitem{abadi2016tensorflow}
M.~Abadi, P.~Barham, J.~Chen, Z.~Chen, A.~Davis, J.~Dean, M.~Devin,
  S.~Ghemawat, G.~Irving, M.~Isard, M.~Kudlur, J.~Levenberg, R.~Monga,
  S.~Moore, D.~G. Murray, B.~Steiner, P.~Tucker, V.~Vasudevan, P.~Warden,
  M.~Wicke, Y.~Yu, and X.~Zheng, ``{TensorFlow}: A system for {Large-Scale}
  machine learning,'' in \emph{12th USENIX Symposium on Operating Systems
  Design and Implementation (OSDI 16)}.\hskip 1em plus 0.5em minus 0.4em\relax
  USENIX Association, 2016, pp. 265--283.

\bibitem{jin2020pontryagin}
W.~Jin, Z.~Wang, Z.~Yang, and S.~Mou, ``Pontryagin differentiable programming:
  An end-to-end learning and control framework,'' in \emph{Advances in Neural
  Information Processing Systems}, 2020.

\bibitem{amos2018differentiable}
B.~Amos, I.~Jimenez, J.~Sacks, B.~Boots, and J.~Z. Kolter, ``Differentiable mpc
  for end-to-end planning and control,'' in \emph{Advances in Neural
  Information Processing Systems}, 2018, pp. 8299–--8310.

\bibitem{sakoe1978dynamic}
H.~Sakoe and S.~Chiba, ``Dynamic programming algorithm optimization for spoken
  word recognition,'' \emph{IEEE Transactions on Acoustics, Speech, and Signal
  Processing}, vol.~26, no.~1, pp. 43--49, 1978.

\bibitem{chang2019d3tw}
C.-Y. Chang, D.-A. Huang, Y.~Sui, L.~Fei-Fei, and J.~C. Niebles, ``D3tw:
  Discriminative differentiable dynamic time warping for weakly supervised
  action alignment and segmentation,'' in \emph{IEEE Conference on Computer
  Vision and Pattern Recognition}, 2019, pp. 3546--3555.

\bibitem{vakanski2012trajectory}
A.~Vakanski, I.~Mantegh, A.~Irish, and F.~Janabi-Sharifi, ``Trajectory learning
  for robot programming by demonstration using hidden markov model and dynamic
  time warping,'' \emph{IEEE Transactions on Systems, Man, and Cybernetics},
  vol.~42, no.~4, pp. 1039--1052, 2012.

\bibitem{vukovic2015trajectory}
N.~Vukovi{\'c}, M.~Miti{\'c}, and Z.~Miljkovi{\'c}, ``Trajectory learning and
  reproduction for differential drive mobile robots based on gmm/hmm and
  dynamic time warping using learning from demonstration framework,''
  \emph{Engineering Applications of Artificial Intelligence}, vol.~45, pp.
  388--404, 2015.

\bibitem{jin2020inverse}
Z.~Liang, W.~Jin, and S.~Mou, ``An iterative method for inverse optimal
  control,'' in \emph{Asian Control Conference}, 2022, pp. 959--964.

\bibitem{akgun2012keyframe}
B.~Akgun, M.~Cakmak, K.~Jiang, and A.~L. Thomaz, ``Keyframe-based learning from
  demonstration,'' \emph{International Journal of Social Robotics}, vol.~4,
  no.~4, pp. 343--355, 2012.

\bibitem{akgun2012trajectories}
B.~Akgun, M.~Cakmak, J.~W. Yoo, and A.~L. Thomaz, ``Trajectories and keyframes
  for kinesthetic teaching: A human-robot interaction perspective,'' in
  \emph{ACM/IEEE international conference on Human-Robot Interaction}, 2012,
  pp. 391--398.

\bibitem{fiacco1976sensitivity}
A.~V. Fiacco, ``Sensitivity analysis for nonlinear programming using penalty
  methods,'' \emph{Mathematical programming}, vol.~10, no.~1, pp. 287--311,
  1976.

\bibitem{levysensitivity}
A.~Levy and R.~Rockafellar, ``Sensitivity of solutions in nonlinear programs
  with nonunique multiplier,'' \emph{Recent Advances in Nonsmooth Optimzation},
  pp. 215--223.

\bibitem{krantz2012implicit}
S.~G. Krantz and H.~R. Parks, \emph{The implicit function theorem: history,
  theory, and applications}.\hskip 1em plus 0.5em minus 0.4em\relax Springer
  Science \& Business Media, 2012.

\bibitem{jin2021safe}
W.~Jin, S.~Mou, and G.~J. Pappas, ``Safe pontryagin differentiable
  programming,'' in \emph{Advances in Neural Information Processing Systems},
  2021.

\bibitem{li2004iterative}
W.~Li and E.~Todorov, ``Iterative linear quadratic regulator design for
  nonlinear biological movement systems,'' in \emph{International Conference on
  Informatics in Control, Automation and Robotics}, vol.~2, 2004, pp. 222--229.

\bibitem{jacobson1970differential}
D.~H. Jacobson and D.~Q. Mayne, \emph{Differential dynamic programming}.\hskip
  1em plus 0.5em minus 0.4em\relax Elsevier Publishing Company, 1970, no.~24.

\bibitem{Andersson2019}
J.~A.~E. Andersson, J.~Gillis, G.~Horn, J.~B. Rawlings, and M.~Diehl,
  ``{CasADi} -- {A} software framework for nonlinear optimization and optimal
  control,'' \emph{Mathematical Programming Computation}, vol.~11, no.~1, pp.
  1--36, 2019.

\bibitem{patterson2014gpops}
M.~A. Patterson and A.~V. Rao, ``Gpops-ii: A matlab software for solving
  multiple-phase optimal control problems using hp-adaptive gaussian quadrature
  collocation methods and sparse nonlinear programming,'' \emph{ACM
  Transactions on Mathematical Software}, vol.~41, no.~1, pp. 1--37, 2014.

\bibitem{kolstad1990derivative}
C.~D. Kolstad and L.~S. Lasdon, ``Derivative evaluation and computational
  experience with large bilevel mathematical programs,'' \emph{Journal of
  optimization theory and applications}, vol.~65, no.~3, pp. 485--499, 1990.

\bibitem{lewis2012optimal}
F.~L. Lewis, D.~Vrabie, and V.~L. Syrmos, \emph{Optimal control}.\hskip 1em
  plus 0.5em minus 0.4em\relax John Wiley \& Sons, 2012.

\bibitem{spong2008robot}
M.~W. Spong and M.~Vidyasagar, \emph{Robot dynamics and control}.\hskip 1em
  plus 0.5em minus 0.4em\relax John Wiley \& Sons, 2008.

\bibitem{nielsen2015neural}
C.~C. Aggarwal \emph{et~al.}, ``Neural networks and deep learning,''
  \emph{Springer}, vol.~10, pp. 978--3, 2018.

\bibitem{wachter2006implementation}
A.~W{\"a}chter and L.~T. Biegler, ``On the implementation of an interior-point
  filter line-search algorithm for large-scale nonlinear programming,''
  \emph{Mathematical programming}, vol. 106, no.~1, pp. 25--57, 2006.

\bibitem{kuipers1999quaternions}
J.~B. Kuipers, \emph{Quaternions and rotation sequences}.\hskip 1em plus 0.5em
  minus 0.4em\relax Princeton University Press, 1999, vol.~66.

\bibitem{lee2010geometric}
T.~Lee, M.~Leok, and N.~H. McClamroch, ``Geometric tracking control of a
  quadrotor uav on se(3),'' in \emph{IEEE Conference on Decision and Control},
  2010, pp. 5420--5425.

\bibitem{bellman1966dynamic}
R.~Bellman, ``Dynamic programming,'' \emph{Science}, vol. 153, no. 3731, pp.
  34--37, 1966.

\bibitem{hansen1992new}
P.~Hansen, B.~Jaumard, and G.~Savard, ``New branch-and-bound rules for linear
  bilevel programming,'' \emph{SIAM Journal on scientific and Statistical
  Computing}, vol.~13, no.~5, pp. 1194--1217, 1992.

\bibitem{sinha2017review}
A.~Sinha, P.~Malo, and K.~Deb, ``A review on bilevel optimization: from
  classical to evolutionary approaches and applications,'' \emph{IEEE
  Transactions on Evolutionary Computation}, vol.~22, no.~2, pp. 276--295,
  2017.

\bibitem{ji2021bilevel}
K.~Ji, J.~Yang, and Y.~Liang, ``Bilevel optimization: Convergence analysis and
  enhanced design,'' in \emph{International Conference on Machine Learning},
  2021, pp. 4882--4892.

\bibitem{ghadimi2018approximation}
S.~Ghadimi and M.~Wang, ``Approximation methods for bilevel programming,''
  \emph{arXiv preprint arXiv:1802.02246}, 2018.

\end{thebibliography}

\begin{IEEEbiography}[{\includegraphics[width=1.2in,height=1.2in,keepaspectratio]{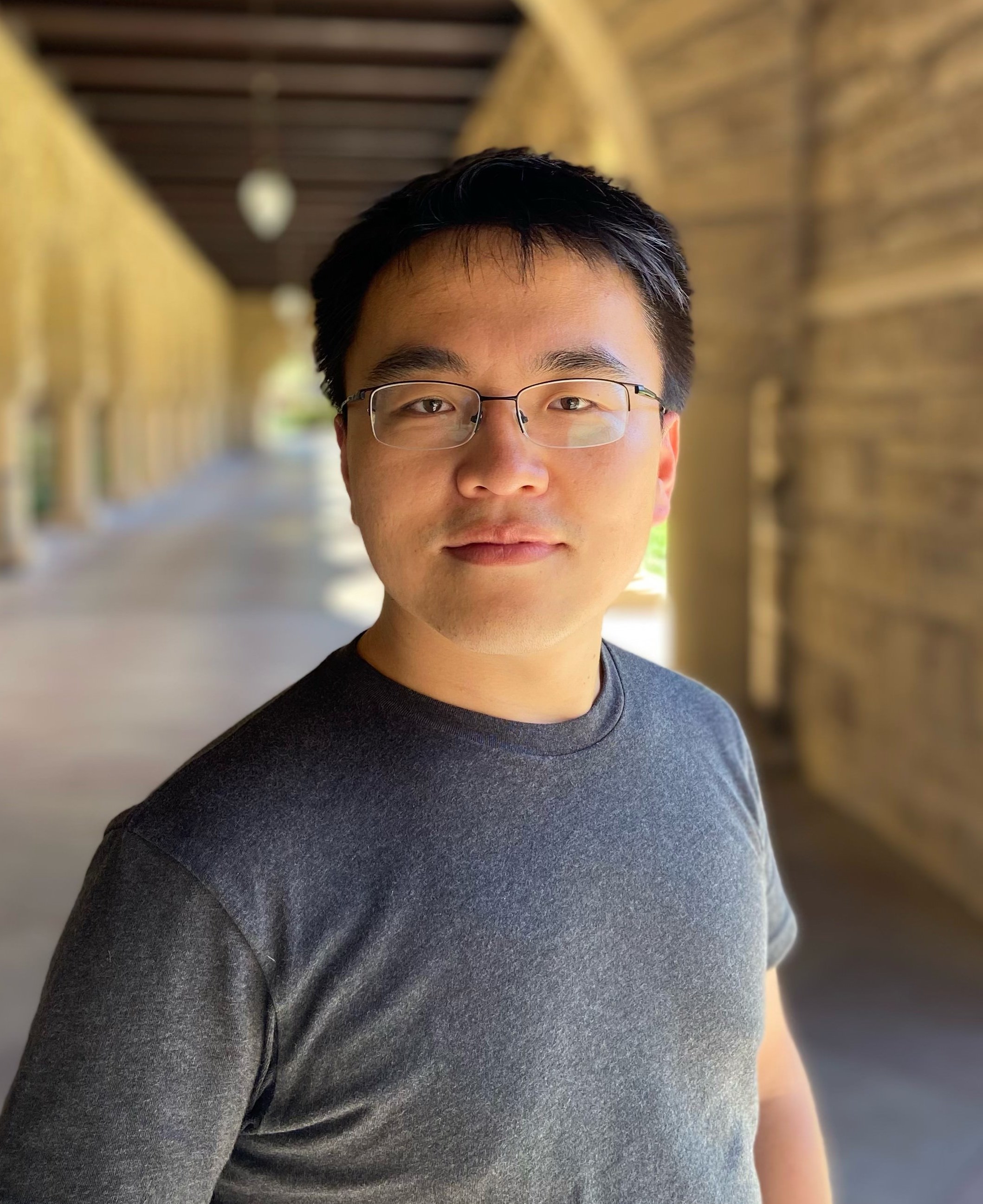}}]
{Wanxin Jin}  is a postdoctoral researcher in the GRASP Laboratory at the University of Pennsylvania. He received the Ph.D. degree in Autonomy and Control at  Purdue University in 2021. From 2016 to 2017, he was a Research Assistant at Technical University Munich, Germany.  Wanxin's research interests include robotics, control, machine learning, and optimization, with emphasis on learning,  planning, and control of robots as they interact with the world and humans.
\end{IEEEbiography}

\vspace{8pt}

\begin{IEEEbiography}[{\includegraphics[width=1.2in,height=1.25in,clip,keepaspectratio]{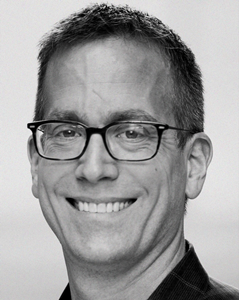}}]
	{Todd D. Murphey}
received his B.S. degree in mathematics from the University of Arizona and the Ph.D. degree in Control and Dynamical Systems from the California Institute of Technology. He is a Professor of Mechanical Engineering at Northwestern University. His laboratory is part of the Center for Robotics and Biosystems, and his research interests include robotics, control, machine learning in physical systems, and computational neuroscience.
\end{IEEEbiography}

\vspace{8pt}

\begin{IEEEbiography}[{\includegraphics[width=1.0in,height=1.25in,keepaspectratio]{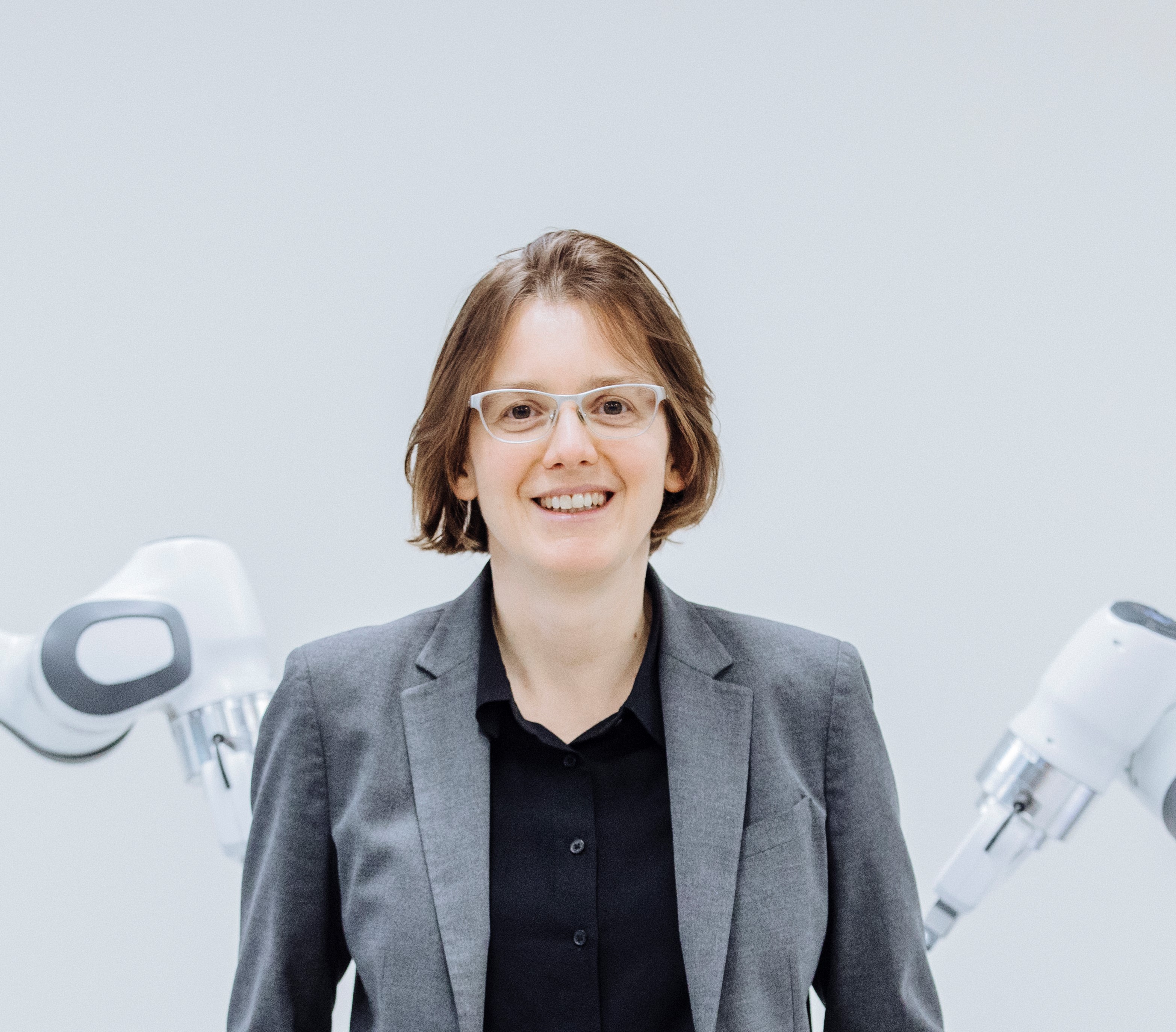}}]
{Dana Kuli\'{c}}  conducts research in robotics and human-robot interaction (HRI), and develops autonomous systems that can operate in concert with humans, using natural and intuitive interaction strategies while learning from user feedback to improve and individualize operation over long-term use. Dana Kuli\'{c} received the combined B. A. Sc. and M. Eng. degree in electro-mechanical engineering, and the Ph. D. degree in mechanical engineering from the University of British Columbia, Canada, in 1998 and 2005, respectively. From 2006 to 2009, Dr. Kuli\'{c} was a JSPS Post-doctoral Fellow and a Project Assistant Professor at the Nakamura-Yamane Laboratory at the University of Tokyo, Japan. In 2009, Dr. Kuli\'{c} established the Adaptive System Laboratory at the University of Waterloo, Canada, conducting research in human robot interaction, human motion analysis for rehabilitation and humanoid robotics.  Since 2019, Dr. Kuli\'{c} is a professor and director of Monash Robotics at Monash University, Australia. In 2020, Dr. Kuli\'{c} was awarded the ARC Future Fellowship.  Her research interests include robot learning, humanoid robots, human-robot interaction and mechatronics.
\end{IEEEbiography}

\vspace{8pt}

\begin{IEEEbiography}[{\includegraphics[width=1.0in,height=1.25in,clip,keepaspectratio]{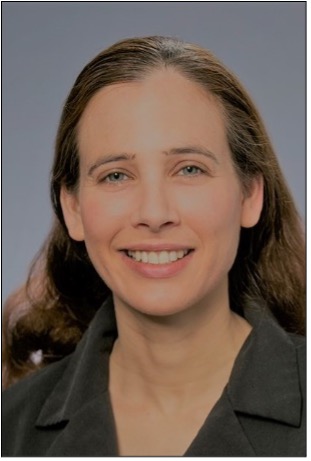}}]
	{Neta Ezer}
is the Northrop Grumman Corporate Director of Strategic Planning. Dr. Ezer previously served as Technical Fellow and Chief Technologist for Human-Machine Teaming in Northrop Grumman Mission Systems and as an Artificial Intelligence (AI) Architect for the Northrop Grumman AI Campaign.
	Prior to Northrop Grumman, Dr. Ezer was a Senior Human Engineering Researcher at the Futron Corporation, working on NASA Orion, International Space Station and human-robot interaction research. She served as Assistant Professor of Industrial Design at the Georgia Institute of Technology. Dr. Ezer has over 15 years of experience in human factors, user experience and AI, with over 40 published papers and proceedings in these areas. Dr. Ezer holds a B.S. in Industrial Design and M.S. and PhD degrees in Engineering Psychology from the Georgia Institute of Technology.
\end{IEEEbiography}

\vspace{8pt}

\begin{IEEEbiography}[{\includegraphics[width=1.0in,height=1.25in,clip,keepaspectratio]{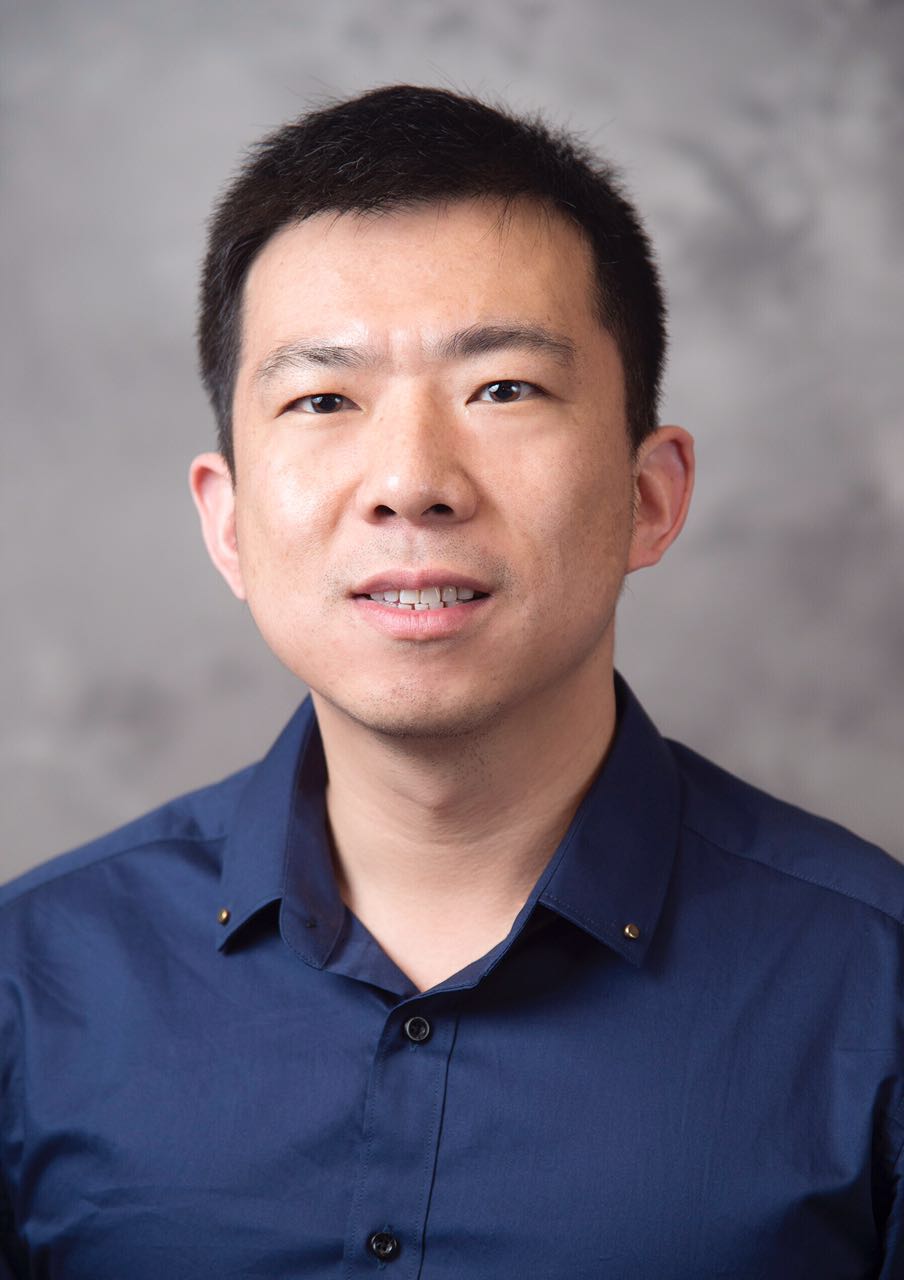}}]
	{Shaoshuai Mou}
is an Associate Professor in the School of Aeronautics and Astronautics at Purdue University. Before joining Purdue, he received a Ph.D. in Electrical Engineering at Yale University in 2014 and worked as a postdoc researcher at MIT for a year after that. His research interests include multi-agent system, control and learning, robotics control, human-robot teaming, resilient autonomy, and also experimental research involving autonomous air and ground vehicles. Dr. Mou co-directs Purdue University’s Center for Innovation in Control, Optimization and Networks (ICON), which aims to integrate classical theories in control/optimization/networks with recent advances
in machine learning/AI/data science to address fundamental challenges in autonomous and connected systems.  
\end{IEEEbiography}

\vfill

\end{document}